\documentclass[pdflatex,sn-mathphys-num]{sn-jnl}% Math and Physical Sciences Numbered Reference Style
\usepackage[minimal]{chemmacros}
\usepackage{graphicx}%
\usepackage{multirow}%
\usepackage{amsmath,amssymb,amsfonts}%
\usepackage{amsthm}%
\usepackage{mathrsfs}%
\usepackage[title]{appendix}%
\usepackage{xcolor}%
\usepackage{xspace}%
\usepackage{textcomp}%
\usepackage{manyfoot}%
\usepackage{booktabs}%
\usepackage{algorithm}%
\usepackage{algorithmicx}%
\usepackage{algpseudocode}%
\usepackage{listings}%
\usepackage{soul}%
\usepackage{threeparttable}
\usepackage{subcaption}
\usepackage{rotating}
\usepackage{longtable}
\usepackage{array}
\usepackage{mdframed}%additional
\usepackage{fancyvrb}%additional
\usepackage{spverbatim}%additional
\usepackage{fvextra}
\usepackage{soul}
\usepackage{subcaption}
\usepackage[T1]{fontenc}
\theoremstyle{thmstyleone}%
\theoremstyle{thmstyletwo}%

\theoremstyle{thmstylethree}%

\newcommand{\main}{\mbox{$\mathop{\textsc{MADDIE}}\limits$}\xspace}
\newcommand{\orchestrator}{\mbox{$\mathop{\normalfont\textsc{Orchestrator}}\limits$}\xspace}
\newcommand{\hitidentifier}{\mbox{$\mathop{\normalfont\textsc{HitIdentifier}}\limits$}\xspace}
\newcommand{\leadoptimizer}{\mbox{$\mathop{\normalfont\textsc{LeadOptimizer}}\limits$}\xspace}
\newcommand{\synthesizer}{\mbox{$\mathop{\normalfont\textsc{Synthesizer}}\limits$}\xspace}
\newcommand{\evaluator}{\mbox{$\mathop{\normalfont\textsc{Evaluator}}\limits$}\xspace}
\newcommand{\workspace}{workspace\xspace}

\newcommand{\delegate}{\mbox{$\mathop{delegate}\limits$}\xspace}
\newcommand{\process}{\mbox{$\mathop{process}\limits$}\xspace}

\newcommand{\vina}{\mbox{$\mathop{\mathtt{Vina}}\limits$}\xspace}
\newcommand{\qedd}{\mbox{$\mathop{\mathtt{QED}}\limits$}\xspace}
\newcommand{\sascore}{\mbox{$\mathop{\mathtt{SAS}}\limits$}\xspace}
\newcommand{\mut}{\mbox{$\mathop{\mathtt{Mut}}\limits$}\xspace}
\newcommand{\clintox}{\mbox{$\mathop{\mathtt{CTx}}\limits$}\xspace}
\newcommand{\herg}{\mbox{$\mathop{\mathtt{hERG}}\limits$}\xspace}
\newcommand{\lilly}{\mbox{$\mathop{\mathtt{Lly}}\limits$}\xspace}
\newcommand{\syn}{\mbox{$\mathop{\mathtt{Syn}}\limits$}\xspace}
\newcommand{\diversity}{\mbox{$\mathop{\mathtt{Div}}\limits$}\xspace}

\newcommand{\out}{\mbox{$\mathop{\textbf{$N_{out}$}}\limits$}\xspace}
\newcommand{\sat}{\mbox{$\mathop{\textbf{$N_{sat}$}}\limits$}\xspace}
\newcommand{\pass}{\mbox{$\mathop{\texttt{Pass}}\limits$}\xspace}
\newcommand{\tgtout}{\mbox{$\mathop{\textbf{$N_{tout}$}}\limits$}\xspace}

\newcommand{\interm}{\mbox{$\mathop{\textbf{$N_{itm}$}}\limits$}}

\newcommand{\liddia}{\mbox{$\mathop{\textsc{LIDDIA}}\limits$}\xspace}

\newcommand{\prompttopill}{Prompt\hbox{-}to\hbox{-}Pill\xspace}
\newmdenv[
  backgroundcolor=gray!5,
  linecolor=gray!50,
  skipabove=6pt,
  skipbelow=6pt,
  innertopmargin=6pt,
  innerbottommargin=6pt,
  innerleftmargin=6pt,
  innerrightmargin=6pt,
  font=\small
]{promptbox}

\newcommand{\molecule}{\mbox{$\mathop{m}\limits$}\xspace}

\AtBeginDocument{\hypersetup{linkcolor=black,filecolor=black}}

\begin{document}

\title[Article Title]{\main: Multi-Agent Drug Discovery Engine}


\author[1]{Ye Liu}\email{liu.12989@osu.edu}
\equalcont{These authors contributed equally to this work.}

\author[2]{Reza Averly}\email{averly.1@osu.edu}
\equalcont{These authors contributed equally to this work.}

\author[1]{Avery Meyer}\email{meyer.1938@osu.edu}
\author[2]{Frazier N. Baker}\email{baker.3239@osu.edu}
\author[2]{Ruoxi Gao}\email{gao.2396@osu.edu}
\author[3]{Daniel Adu-Ampratwum}\email{adu-ampratwum.1@osu.edu}
\author*[1,2,3,4]{Xia Ning}\email{ning.104@osu.edu}

\affil[1]{Biomedical Informatics, The Ohio State University}
\affil[2]{Computer Science and Engineering, The Ohio State University}
\affil[3]{Medicinal Chemistry and Pharmacognosy, The Ohio State University}
\affil[4]{Translational Data Analytics Institute, The Ohio State University}

% \affil[2]{\orgdiv{Department}, \orgname{Organization}, \orgaddress{\street{Street}, \city{City}, \postcode{10587}, \state{State}, \country{Country}}}

% \affil[3]{\orgdiv{Department}, \orgname{Organization}, \orgaddress{\street{Street}, \city{City}, \postcode{610101}, \state{State}, \country{Country}}}


\abstract{    
Drug discovery remains a resource-intensive and time-consuming process, involving billions of dollars and a decade of development to advance a drug to market.
While many \textit{in silico} tools have accelerated this process, 
operating them still demands substantial time, effort, and expertise from medicinal chemists.
Autonomous systems built on Large Language Models (LLMs) have emerged as a promising solution, 
yet current designs limit their performance.
Existing single-agent systems scale poorly to long-horizon, multi-step, and realistic tasks, 
whereas multi-agent systems typically execute predefined workflows that cannot capture the iterative, non-linear nature of a real drug discovery campaign.
%
% but single-agent systems scale poorly to complex tasks such as drug discovery,
% \yecomment{``such as" is not appropriate, as we only discuss drug discovery.}
% while multi-agent systems typically follow predefined workflows that do not reflect \yecomment{``reflect" is relatively weak. How about address or other verbs?} the non-linearity of a real drug-discovery campaign.
%
Here, we present \main, a multi-agent drug discovery engine.
It comprises a central agent, \orchestrator, and four specialized worker agents: \hitidentifier, \leadoptimizer, \synthesizer, and \evaluator, each representing a critical step in a drug discovery campaign.
% \yecomment{Do we need specify the name of these agents?}
%
Under \orchestrator's orchestration and coordination, 
these worker agents are composed into flexible workflows, enabling \main to adapt its strategy and navigate the drug discovery pipeline end-to-end.
Across 50 protein targets, \main produces molecules satisfying key property requirements and their synthesis routes, outperforming existing agentic systems, generative models, and general-purpose LLMs by a wide margin.
Our analysis reveals how these agents collaborate and that 
\main can employ different strategies across targets to achieve the campaign's goal, 
while still being efficient in time and tokens.
Together, they demonstrate \main as a low-cost, effective, and efficient autonomous engine for accelerating drug discovery. 
Code and dataset is available via \mbox{\url{https://github.com/ninglab/MADDIE}}. 

% \yecomment{I feel we can also highlight that \main gives the synthesizable routes for the output molecues, which xxx.}
% 
% \yecomment{The abstract above seems not finished. I drafted one and you can revise it. (@Reza)}
% \ye{
% Drug discovery requires searching an expansive chemical space against many competing requirements—
% % 
% binding affinity, drug-likeness, safety and synthesizability—
% % 
% and although in silico tools now address each of these, deploying them across a campaign still falls to medicinal chemists. 
% % 
% Autonomous systems powered by large language models have begun to take on this role, 
% % 
% but single-agent systems scale poorly to realistic campaigns, 
% % 
% while multi-agent systems typically follow predefined workflows, 
% % 
% at odds with campaigns that are non-linear and revise their course as evidence accumulates. 
% % 
% In this paper, we present \main, a Multi-Agent Drug Discovery Engine. 
% % 
% A central \orchestrator agent composes four specialized worker agents—\hitidentifier, \leadoptimizer, \synthesizer and \evaluator—into workflows it determines as a campaign unfolds, 
% % 
% sustained by a memory module and an evolving repository of candidate molecules. 
% % 
% Across 50 real protein targets, 
% % 
% \main is the only method to deliver the requested molecules jointly satisfying every requirement on every target, outperforming agentic, generative and general-purpose language model baselines by a wide margin.
% % 
% }
}

\maketitle

% \rz{IMPORTANT: Each person will be assigned with only one color in this paper. Feel free to pick your own color and modify the macros in macros.tex. \rz{Reza: Cyan}. \xia{Xia: Red}. \fb{Frazier: Olive (greenish gold color)}.}
% . \rx{Ruoxi: Green}. \ye{Ye: Teal}. \am{Avery: Violet}.

% \begin{bibunit}[sn-mathphys-num]
%%%%%%%%%%%%%%%%%%%%%%%%%%%%%%%%%%%
\section{Introduction}\label{section:introduction}
%%%%%%%%%%%%%%%%%%%%%%%%%%%%%%%%%%%

% \rz{
% \begin{itemize}
%     \item Name of methods:
%     \item (*) MADDIE: Multi Agent Drug Discovery Intelligent Engine
%     \item (1) MIDAS: \textbf{M}ulti-agent \textbf{I}n Silico \textbf{D}rug Discovery System
%     \item (*) MALADY: Multi-agent Language-based Drug Discovery Agent ("malady" is a disease, physical illness, or serious condition that causes suffering)
% \end{itemize}
% }

% \rz{
% \begin{itemize}
%     \item Drug discovery is a very long and expensive process
%     \item Research has sought to expedite the process
%     \item Recent trend has been on agent -- LLM powered with computational tools
%     \item More specifically on multi-agent, but they often use a predefined sequence, which does not reflect the non-linear process of real-world DD, or tackle only a subtask of drug discovery
%     \item We propose LIDDIA2, with 5 components blablabla. Add novelty here.
%     \item Contribution paragraph
%     \item Contributions: 
%     \item (1) LIDDIA2 as a multi-agent system for in silico drug discovery
%     \item (2) Provide rigorous benchmarking against many baselines, many metrics and meaningful targets
%     \item (3) Provide insights as to why LIDDIA2 can achieve such results, emphasizing on the collaboration of each agent and how they adapt to a campaign, resulting in different trajectories
%     \item (4) Provide a case study on an Alzheimer target, against reference ligand and approved drug, and also expert-evaluated
% \end{itemize}
% }

Drug discovery is a complex, capital-intensive, and time-consuming process: bringing one approved drug to market typically costs more than one billion dollars
% ~\citep{10.1001/jama.2020.1166-drug-discovery-time,sertkaya2024costs,10.1001/jamanetworkopen.2024.53275-drug-discovery-billions} 
and takes more than a decade~\citep{10.1001/jama.2020.1166-drug-discovery-time,sertkaya2024costs,10.1001/jamanetworkopen.2024.53275-drug-discovery-billions,drug-discovery-years}. 
% ~\citep{10.1001/jama.2020.1166-drug-discovery-time,drug-discovery-years}. 
%
To expedite this process, many \textit{in silico} tools have been developed~\citep{https://doi.org/10.1002/jcc.21334-vina,10.1021/acs.jctc.2c01145-unidock,swanson2024admet-admetai,chen2025generating-diffsmol,Dey2025,gao2026generatingdevelopable3dmolecules-conditar}, each designed for a specific task in the pipeline (e.g., predicting binding affinity).
% ~\citep{10.1021/acs.jctc.2c01145-unidock}). 
%
Using them, however, has traditionally fallen to medicinal chemists. 
Autonomous systems capable of intelligently using these tools have gained considerable attention recently~\citep{averly-etal-2025-liddia,liu2025drugagentautomatingaiaideddrug,pan2025frogent,gao2025pharmagents,wang2025txgemma,cao2026mozi,Kantorow2026.06.20.733481,10.1093/bioadv/vbaf323-prompt-to-pill,10.1021/acs.jcim.5c02454-agentd},
substantially reducing the time and effort demanded of medicinal chemists while showing promising results.
These systems are typically built upon Large Language Models (LLMs), reasoning engines capable of intelligent decision-making and planning over complex tasks~\citep{singh2025openai-gpt5}.
LIDDIA~\citep{averly-etal-2025-liddia} is one of the first such systems: a \emph{single}-agent capable of autonomously navigating drug discovery \emph{in silico}.
Following LIDDIA, several systems have been introduced, with more attention towards \emph{multi}-agent systems~\citep{10.1021/acs.jcim.5c02454-agentd,liu2025drugagentautomatingaiaideddrug,wang2025txgemma,gao2025pharmagents,10.1093/bioadv/vbaf323-prompt-to-pill}. 
Some of them
% , such as DrugAgent~\citep{liu2025drugagentautomatingaiaideddrug} and TxGemma~\citep{wang2025txgemma}, 
are designed for a specific task under the drug discovery pipeline,
% ~\citep{liu2025drugagentautomatingaiaideddrug,wang2025txgemma}, 
while others for end-to-end drug discovery~\citep{gao2025pharmagents,10.1093/bioadv/vbaf323-prompt-to-pill}. 
% \yecomment{In the first paragraph, there are several reference papers used multiple times, such as~\citep{liu2025drugagentautomatingaiaideddrug,wang2025txgemma}. Please check them and avoid repeated use, particularly in one paragraph.}

While promising, these agentic systems leave room for improvements.
Single-agent systems such as \mbox{LIDDIA}~\citep{averly-etal-2025-liddia} and 
% \amcomment{LIDDIA or \textsc{LIDDiA}? Whichever form, make consistent across all LIDDIA mentions}
\mbox{AgentD}~\citep{10.1021/acs.jcim.5c02454-agentd} demonstrate how LLMs can intelligently coordinate different \textit{in silico} tools for drug discovery, yet their performance does not scale well to more realistic and complex tasks (Section~\ref{sec:results:main}).
Multi-agent systems such as \mbox{PharmAgents}~\citep{gao2025pharmagents}, \prompttopill~\citep{10.1093/bioadv/vbaf323-prompt-to-pill} and \mbox{F.A.D.E.}~\citep{Kantorow2026.06.20.733481} are designed for complex tasks, but they typically assume a predefined workflow. 
This assumption does not reflect real-world drug discovery, which is a highly non-linear process governed by many requirements. 
For instance, successful drugs must bind well to the target, while maintaining favorable pharmacokinetic and physicochemical property profiles~\citep{DSegall2012}.
However, identifying such molecules is highly difficult, requiring numerous iterations of searching the expansive chemical space~\citep{bohacek1996art}.
Other multi-agent systems have also been proposed that enable more flexible workflows but are still limited in adaptability~\citep{pan2025frogent,cao2026mozi}. 

%
% Other multi-agent systems have also been proposed, though they are typically designed for a specific task under the drug discovery pipeline~\citep{liu2025drugagentautomatingaiaideddrug,wang2025txgemma}.

% Multi-agent systems such as Prompt-to-Pill~\citep{10.1093/bioadv/vbaf323-prompt-to-pill} and AgentD~\citep{10.1021/acs.jcim.5c02454-agentd} are designed for complex tasks, but they typically assume a predefined workflow.
% %
% Multi-agent systems such as Prompt-to-Pill~\citep{10.1093/bioadv/vbaf323-prompt-to-pill} and AgentD~\citep{10.1021/acs.jcim.5c02454-agentd} are designed for complex tasks, but they typically assume a predefined workflow. 
% %
% This assumption does not reflect real-world drug discovery, which is a highly non-linear process governed by many requirements. 
% %
% For instance, successful drugs must bind well to the target while maintaining favorable pharmacokinetic and physicochemical property profiles~\citep{X}.
% %
% However, identifying such molecules is incredibly challenging, requiring numerous iterations of searching the chemical space and evaluation~\citep{X}.

\begin{figure}[t]
    \centering
    \includegraphics[width=0.8\linewidth]{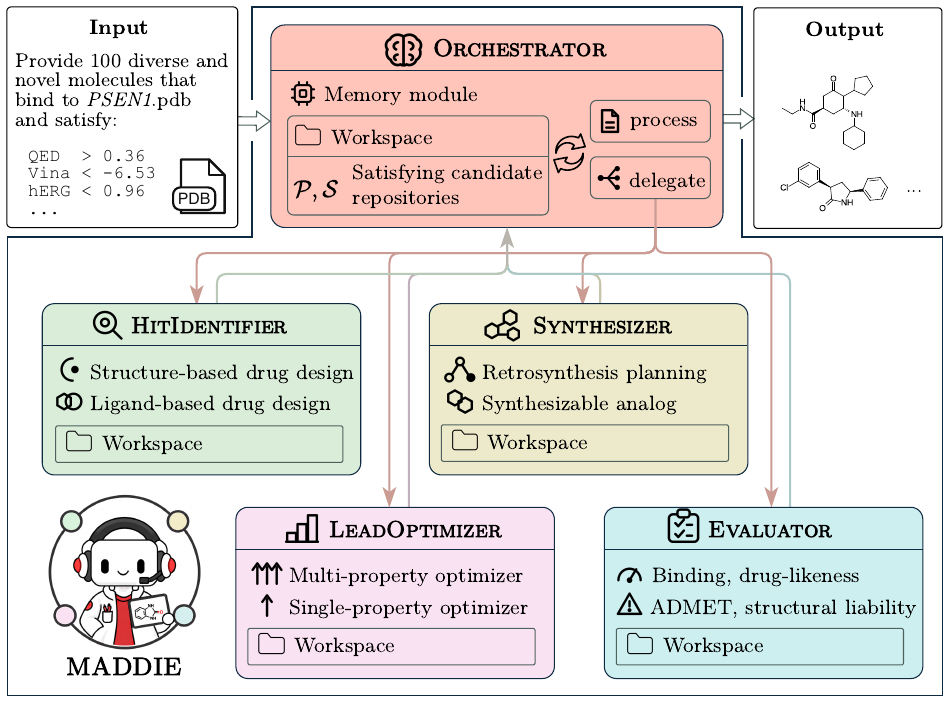}
    \vspace{3pt}
    \caption{\textbf{\main} architecture, consisting of \orchestrator, workspace \ and four worker agents: \hitidentifier, \leadoptimizer, \synthesizer and \evaluator. 
    %\yecomment{The blocks for four worker agents are taller than \orchestrator, still leaving much space. Can you update this, @Reza?}
    }
    \label{fig:method-main-candidate}
\end{figure}

Building upon these insights, we propose \main, a \textbf{M}ulti-\textbf{A}gent \textbf{D}rug \textbf{DI}scovery \textbf{E}ngine. 
As illustrated in Figure~\ref{fig:method-main-candidate}, the system is composed of five agents, a central agent called \textbf{(1)} \orchestrator, and four specialized worker agents: \textbf{(2)} \hitidentifier; \textbf{(3)} \leadoptimizer; \textbf{(4)} \synthesizer; and \textbf{(5)} \evaluator. 
\orchestrator serves as an intelligent orchestration component that governs the whole campaign. 
Unlike static workflows, \orchestrator dynamically determines the workflow through its actions, 
% \yecomment{was wondering should we unify the terms, ``step" and ``action"?}
sustains a long-horizon campaign using its memory module and monitors the progress by maintaining its evolving repository of candidate molecules.
% (Section~\ref{section:method:orchestrator}). 
%
Under its orchestration, \main's worker agents are composed into flexible workflows, enabling \main to adapt its strategy as a campaign unfolds, mirroring the iterative, decision-driven nature of real-world drug discovery.
Each worker agent carries out a step critical in advancing a drug discovery campaign---from identifying hits and optimizing them, to proposing synthesis routes and evaluating the quality of molecules. 
% \st{molecule candidates' quality.}
% \amcomment{quality of candidate molecules}
% ---while \orchestrator enables \main to navigate the drug discovery pipeline end-to-end effectively. 
%
\main's worker agents are equipped with state-of-the-art \textit{in silico} tools, including generative models for \textit{de novo} molecular generations,
thus allowing \main to explore beyond known chemical space and propose novel molecules designed for each campaign's objectives.

We rigorously benchmark \main with a wide range of methods designed to accelerate drug discovery.
We highlight several insights from \main here:
% \yecomment{From my understanding, generally we list points to show the contributions rather than pure experimental findings. Also, I think the current introduction reflects the paper's issue: it feels discrete rather than a coherent paper. Each section is not closely related.} \rzcomment{I'm not sure about second sentence, do you have any examples why it feels discrete, and how to improve?}
% \yecomment{For example, this paragraph seems a summary of the experimental section. Beyond these, we need highlight the advantage of \main, not only through the number on performance, but the framework design, motivation.}
% \rzcomment{Thanks! It's revised now}
%
\textbf{(1)} \main outperforms existing baselines by a large margin,
generating molecules satisfying key pharmaceutical properties required for successful drugs across 50 targets, 
along with providing their synthesis routes (Section~\ref{sec:results:main}).
%
%<<<<<<< HEAD
%\textbf{(2)} \main performs strongly due to the collaborative efforts of each agent, emphasizing their importance in how they contribute to the satisfying molecules (Section~\ref{sec:results:ablation}). 
%Notably, \orchestrator contributes the most by enabling \main to focus on the most promising candidates, greatly reducing the molecule exploration needed for \main to achieve its goal; 
%=======
This is consistent with our motivation for the design of
% \amcomment{the design, implementation, development, etc. of} 
\main: 
an autonomous system that can scale to realistic tasks and adapt throughout a campaign to generate satisfying molecules across targets;
% \st{whereas} single-agent systems scale poorly to realistic tasks 
% and multi-agent systems relying on a predefined workflow cannot accommodate the iterative search that such campaigns require, 
% \main adapts its workflow as a campaign unfolds;
% \amcomment{instead maybe "a multi-agent system that can scale to realistic tasks and is adaptive throughout a campaign to achieve "success" spanning the breadth of difficulty of those tasks."}
%
\textbf{(2)} \main's performance arises from the collaborative efforts of each agent, emphasizing their importance in contributing to the satisfying molecules (Section~\ref{sec:results:ablation}). 
Each worker agent makes a distinct contribution, while \orchestrator contributes the most.
% \st{Notably, each worker agent makes a distinct contribution, with ORCHESTRATOR contributing the most:}
% \amcomment{Each worker agent makes a distinct contribution and, notably, \orchestrator contributes the most}: 
Notably, \orchestrator tracks the campaign through its evolving repository of molecules
% \st{candidates} \amcomment{candidate molecules (we haven't defined this yet}
and directs \main's effort towards the most promising ones, 
substantially reducing the number of molecules MADDIE must explore to reach its objective; 
\textbf{(3)} \main demonstrates its capability to adapt its workflow during campaigns for varying targets, employing distinct strategies required to produce the satisfying molecules (Section~\ref{sec:results:agent}).
Even on a target with no existing marketed FDA-approved drugs~\citep{zuo2024gene-abcd1}, \main can identify more than 100 promising molecules through its intelligent adaptation, emphasizing its capability; 
\textbf{(4)} \main generates one promising molecule for PSEN1, a critical target associated with Alzheimer's Disease, with the molecule's \textit{in silico} properties exceeding those of the reference ligand and the associated FDA-approved drug (Section~\ref{section:results:case_study});
and \textbf{(5)} \main serves as a low-cost and efficient autonomous drug discovery system, highlighting its potential to accelerate drug discovery (Section~\ref{section:results:efficiency}). 
Together, these findings indicate that dynamic orchestration of specialized worker agents through a central agent offers an effective design for autonomous drug discovery, 
and \main can serve as a promising system in this domain.
% \amcomment{this is true, but is this all that we are supposed to take away?}

% \newpage

%%%%%%%%%%%%%%%%%%%%%%%%%%%%%%%%%%%
\section{Related Works}\label{section:related_works}
%%%%%%%%%%%%%%%%%%%%%%%%%%%%%%%%%%%

% \begin{itemize}
%     \item \xia{names should not allow breaking across 
% lines.. fix this across the entire paper} \rzcomment{Revised}
%     \item \xia{method names should use non-broken hyphenation! correct this throughout the entire paper} \rzcomment{Revised}
%     \item \xia{following sectioning hierarchies. do not use textbf to create sectionings. } \rzcomment{Revised}
%     % \item \yecomment{conDitar?}\rzcomment{conDitar-dev is correct, we use conDitar, which is conDitar-dev without optimization module. More discussion on why we use conDitar, not conDitar-dev, is in the supplement.}
% \end{itemize}

% \noindent\textbf{Agents for Drug Discovery}\quad 
\subsection{Agents for Drug Discovery}
Agents powered by LLMs are capable of coordinating \textit{in silico} tools across broader drug discovery workflows. 
% Beyond direct molecular generation, LLM agents can coordinate \textit{in silico} tools across broader drug-discovery workflows. 
%
Existing systems operate at different levels of scope and autonomy, from task-specific assistants to end-to-end drug discovery. 
These systems are typically classified into two categories: \textbf{(1)} single-agent system and \textbf{(2)} multi-agent system. 

% \paragraph{Single-agent system}

% \xia{following sectioning hierarchies. do not use textbf to create sectionings. } \rzcomment{Revised} 
%
\mbox{CACTUS}~\citep{10.1021/acsomega.4c08408} focuses on assistive cheminformatics tasks such as estimating molecular properties, retrieving structurally similar molecules, and evaluating drug-likeness. 
\mbox{TxGemma}~\citep{wang2025txgemma} also proposes a conversational assistant for therapeutics predictions and question answering, allowing better interaction with medicinal chemists. 
AgentD~\citep{10.1021/acs.jcim.5c02454-agentd} spans a wider range of the early-stage computational pipeline in drug discovery, from target and ligand information retrieval, to molecular generation, property prediction, and refinement.
\liddia~\citep{averly-etal-2025-liddia} provides an autonomous workflow where LLM-based reasoning guides the iterative process of finding satisfying molecules that fit a specific profile.
%generation, evaluation, and refinement of candidate molecules. 
%
These examples demonstrate that LLMs can coordinate \textit{in silico} tools toward a larger drug discovery goal.
However, their performance is typically limited and does not scale well for more complex and realistic tasks (Section~\ref{sec:results:main}). 
% their decision-making generally occurs within a single agent. 

% \paragraph{Multi-agent system} 

% \hl{Multi-agent systems typically decompose a complex task into subtasks, each handled by a specialized agent, enabling them to tackle problems that exceed the reach of any single-agent system.} 
Multi-agent systems distribute their tasks across specialized agents, enabling them to tackle complex problems, such as drug discovery, that can exceed the reach of any single-agent system.
Their system designs and assigned roles for the agents vary among systems.
%\xia{I do not think this it is right to the point. please revise} \rzcomment{Revised (?)}
%
\mbox{DrugAgent}~\citep{liu2025drugagentautomatingaiaideddrug} coordinates planner and instructor agents to develop, implement, and refine machine-learning solutions using drug discovery domain knowledge.
Robin~\citep{Ghareeb2026} assigns literature-search and data-analysis to separate agents, using their output to loop through hypothesis formulation, experiment planning, and result interpretation.
MADD~\citep{solovev-etal-2025-madd} employs four coordinated agents designed for hit identification, using specialized tools across data processing, properties prediction and molecule generation.
More directly focused on target binding site identification and candidate generation, \mbox{F.A.D.E.}~\citep{Kantorow2026.06.20.733481} uses a hierarchical multi-agent setup to produce ranked candidates and demonstrates its effectiveness through case studies on targets EGFR and CRBP1. 
%handle protein-structure , identify binding sites, generate new molecules, and rank candidates in a user-friendly workflow. Its methods were validated with two case studies for EGFR and CRBP1
%
\prompttopill
% \xia{method names should use non-broken hyphenation! correct this throughout the entire paper}
~\citep{10.1093/bioadv/vbaf323-prompt-to-pill} and \mbox{PharmAgents}~\citep{gao2025pharmagents} work at a broader scope, coordinating  agents across drug discovery phases, from preclinical stage to clinical-trial simulation. 
\mbox{Mozi}~\citep{cao2026mozi} proposes a dual-layer multi-agent architecture for drug discovery, a layer for control and governance and another one for workflow.
\mbox{FROGENT}~\citep{pan2025frogent} shows an example of a decentralized multi-agent system, using an orchestrator agent that decomposes the task into a high-level plan, and the other agents to execute each subtask in the plan. 
% \hl{them} \xia{@Reza, what does ``them" refer to here? } 
% \rz{The plan itself is fixed and the orchestrate agent's role is to decompose the problem, unlike \orchestrator in \main that enables flexible workflow adaptation and its role in governing the entire campaign.}
% \hl{shows an example of a decentralized multi-agent system and how it can be used for end-to-end drug discovery.}
% \hl{demonstrates the capability of a decentralized multi-agent system various individual tasks in drug discovery.}
%\xia{how? does this introduction of a method provide any useful information to the audience} \rzcomment{Revised}
%
Together, these systems demonstrate the overall capability of collaboration between different specialized agents on drug discovery tasks.
% \hl{how specialization can extend agentic workflows across drug discovery tasks and stages. } 
%
However, their coordination generally centers on predefined workflows, and their system evaluations often focus on individual case studies.
This is in contrast to \main, which enables flexible workflow adaptation across different campaigns (Section~\ref{sec:results:agent}).

\subsection{Generative Models for Drug Design}
Generative models have recently been employed for \textit{de novo} drug design to accelerate drug discovery. 
These models generally fall under two categories: \textbf{(1)} structure-based drug design (SBDD) and \textbf{(2)} ligand-based drug design (LBDD).

SBDD methods generate ligand molecules conditioned on the three-dimensional structures of target binding pockets.
Early approaches such as \mbox{Pocket2Mol}~\citep{pmlr-v162-peng22b-pocket2mol} use autoregressive sampling to generate molecular atoms sequentially, 
while more recent approaches are diffusion-based~\citep{guan3d-targetdiff,NEURIPS2024_4ddfe69f}.
% and enable a more flexible generation
%
% However, sequential generation can suffer from error accumulation and limited sampling efficiency.
%
% Diffusion-based approaches have emerged as a more flexible alternative for molecular generation.
%
% More recently, diffusion-based approaches emerge as a more flexible alternative for generation, with \mbox{TargetDiff}~\citep{guan3d-targetdiff} and 
% \mbox{ALIDiff}~\citep{gu2024aligningtargetawaremoleculediffusion} as examples.
% \xia{names should not allow breaking across 
% lines.. fix this across the entire paper} as examples.
% For example, TargetDiff~\citep{guan3d-targetdiff} jointly generates ligand atom types and coordinates through geometric modeling of protein-ligand complexes throughout diffusion process.
%
% To further improve binding affinity, 
% ALIDiff~\citep{gu2024aligningtargetawaremoleculediffusion} finetunes a pretrained diffusion-based SBDD model using preference pairs selected based on binding affinity scores.
%
Notably, \mbox{conDitar-dev}~\citep{gao2026generatingdevelopable3dmolecules-conditar} decouples protein encoding from the diffusion process through a separately pretrained pocket encoder and performs inference-time noise optimization to improve molecular properties beyond binding affinity.
%
% Despite these advances, most conventional SBDD models operate as one-shot generators with fixed generative processes, offering limited flexibility to adapt generation based on downstream evaluation.
%
Several works augment SBDD models with LLM-based refinement~\citep{NEURIPS2025_14391f8f,hu2026steering,10.1145/3774904.3792605-elillm}.
% using generated molecules as starting points for iterative optimization.
ELILLM~\citep{10.1145/3774904.3792605-elillm}, one of the state-of-the-art methods, encodes the initial molecules into the LLM latent space and performs Bayesian optimization to identify promising molecular representations.
% Recent work therefore augments SBDD models with LLM-based refinement,
% using generated molecules as starting points for iterative optimization.
%
% CIDD~\citep{gao2025cidd} is among the first to validate this approach,
% by leveraging Chain-of-Thought reasoning to refine generated molecules for improved drug-likeness and structural rationality.
%
% followed by \mbox{K-BTS}~\citep{hu2026steering} and \mbox{ELILLM}~\citep{10.1145/3774904.3792605-elillm}. 
%
% Specifically, \mbox{ELILLM}~\citep{10.1145/3774904.3792605-elillm} encodes the initial molecules into the LLM latent space and performs Bayesian optimization to identify promising molecular representations. 
% K-BTS~\citep{hu2026steering} embeds LLM refinement within a bi-level Thompson sampling framework,
% where historical molecular transitions and docking feedback are used to update the beliefs and sampling parameters.
%
% To facilitate efficient and controllable refinement, 
% ELILLM~\citep{10.1145/3774904.3792605-elillm} encodes the initial molecules into the LLM latent space and performs Bayesian optimization to identify promising molecular representations.

LBDD generates molecules conditioned on the shape of the ligand that binds to the target structure's pocket. 
Previous approaches include \mbox{SQUID}~\citep{adams2023equivariant-squid}, using a fragment-based generative model to decode fragments and a scoring model to adjust the angles of rotatable bonds between fragments, and 
% \rx{\mbox{DESERT}, an image-to-sequence model with an encoder-decoder architecture that generates molecular structures conditioned on voxelized shapes.}
\mbox{DESERT}~\citep{NEURIPS2022_96ddbf81-shape2mol}, an image-to-sequence model with an encoder-decoder architecture that generates molecular structures conditioned on voxelized shapes.
% \hl{{\mbox{Shape2Mol}}~{\citep{NEURIPS2022_96ddbf81-shape2mol}}, using an encoder-decoder framework that encodes molecules into latent embeddings and uses them to sequentially build fragments to generate the molecules.}
%\xia{this is almost the same idea as DiffSMol} 
%% \rzcomment{Hey Ruoxi, can you help on how DiffSMol is different to Shape2Mol?}
%\rzcomment{Revised}
%
%Following these works, 
\mbox{DiffSMol}~\citep{chen2025generating-diffsmol}, one of state-of-the-art methods for LBDD, 
encodes the ligand shape into pretrained embeddings and uses these embeddings to condition a diffusion-based molecular generation model.
% \rz{explain here}.

Together, these generative models provide effective means of finding promising molecules. 
Generation alone, however, is only the first step of drug discovery, and the molecules these models produce often demand substantial manual downstream filtering and selection~\citep{zhavoronkov2019deep,gao2026generatingdevelopable3dmolecules-conditar}.
\main addresses this gap by integrating the filtering and selection into the system itself: 
\hitidentifier and \leadoptimizer propose the molecules, while \evaluator and
\synthesizer assess them---all under \orchestrator's coordination.
\section{Experimental Settings}\label{section:experimental_settings}
%%%%%%%%%%%%%%%%%%%%%%%%%%%%%%%%%%%

\paragraph{Dataset}
We curate a dataset to evaluate \main.
Concretely, we collect \textbf{(1)} a set of 50 protein targets, along with their 3D binding pocket structures, 
% to evaluate \main. 
%
% Compared to existing ones~\citep{averly-etal-2025-liddia,gao2026generatingdevelopable3dmolecules-conditar}, we also curate 
and \textbf{(2)} a set of drugs approved by U.S. Food and Drug Administration (FDA), each is associated with a target in our dataset. 
%
% \amcomment{"compared to existing ones" is awkward and maybe combine the (1) and (2) listed things all into one sentence, or leave the listing out}
%
These drugs are used to define what molecular property requirements \main's generated molecules should meet (Table~\ref{tbl:tools}). 
% and as a benchmark to compare with (Supplement~\ref{S-supp:prop_analysis}). 
%
We provide the full protein targets, their annotations, the curated FDA-approved drugs, and the protocol to collect them in this paper (Appendix~\ref{supp:dataset}).
% \xia{re-order sections in Supplementary materials according to the ordering in which they are referred to in the main text} \rzcomment{Revised}

% \textbf{Protein targets}. \amcomment{textbf needs to be removed (i think this section is the last place with it)}
%
The targets span major therapeutic areas including autoimmune disease, cancer, cardiovascular disease, diabetes, infectious disease, and neurological conditions. 
% h autoimmune disease (A), cancer (CA), cardio-
% vascular disease (CV), Diabetes (D), Infectious Disease (I), or Neurological Conditions
% (N).
%
The dataset includes 42 targets from existing datasets: \textbf{(1)} \mbox{LIDDIA}~\citep{averly-etal-2025-liddia} (12 targets) and \textbf{(2)} \mbox{CrossDocked2020}~\citep{10.1021/acs.jcim.0c00411} (30 targets), and additional 8 targets from OpenTargets~\citep{10.1093/nar/gkae1128}.
We ensure that these targets do not exist in \mbox{CrossDocked2020} training set, thus allowing \main to be compared against methods trained on \mbox{CrossDocked2020}. 
We use the experimentally validated 3D structures available~\citep{Berman2000} and preprocess them to extract the binding pockets, similar to \mbox{LIDDIA's} dataset preprocessing~\citep{averly-etal-2025-liddia}. 

% \textbf{Approved FDA drugs}.
%
For each target, we identify whether it has drugs that are \textbf{(1)} small molecules, \textbf{(2)} FDA-approved, and \textbf{(3)} currently marketed as of July 2026. 
Out of 50 targets, 22 targets are associated with at least one drug.
For the remaining targets (28), no drug associations are identified, indicating opportunities for drug development for these targets. 
We use these drugs to calculate the property requirements for our experiment (Appendix~\ref{supp:reqs}).

\paragraph{Evaluation Metrics}
We use several metrics to evaluate the performance of \main. 
These metrics are based on 
% whether the molecules can satisfy multiple constraints commonly considered in real-world drug discovery.  
\textbf{(1)} the molecular properties often considered in real-world drug discovery, 
\textbf{(2)} the synthesizability of the molecule, and
\textbf{(3)} whether the molecule can satisfy multiple property and synthesizability requirements simultaneously. 

% \textbf{Molecular properties}. \amcomment{remove the bf, I think the two bf sections could be combined into one paragraph}
%
% The first set of constraints concerns whether a molecule exhibits key properties similar to a successful drug. 
%
We evaluate the generated molecules on several key properties critical for a successful drug.  
% each computed using a well-established, high-performing computational tool (Table~\ref{X}). 
%
We first consider properties commonly used in the literature~\citep{gao2026generatingdevelopable3dmolecules-conditar}: 
\textbf{(1)} binding affinity, measured by Vina docking score~\citep{https://doi.org/10.1002/jcc.21334-vina} (\vina); 
\textbf{(2)} drug-likeness, or desirability of a molecule using quantitative estimate of drug-likeness~\citep{bickerton2012quantifying-qed} (\qedd); 
and \textbf{(3)} synthesizability estimate, using synthetic accessibility score~\citep{ertl2009estimation-sascore} (\sascore).
In addition to common properties, we further assess the toxicities of these molecules.
Concretely, we measure the probability that a molecule may \textbf{(1)} induce mutations in DNA~\citep{10.1021/ci300400a-mut} (\mut), 
\textbf{(2)} fail clinical trials owing to toxicity~\citep{gayvert2016data-clintox} (\clintox), 
and \textbf{(3)} inhibit the hERG potassium channel~\citep{10.1021/acs.molpharmaceut.6b00471-herg} (\herg). 
%
% Each measures whether the molecule can likely cause genetic damage~\citep{X}, side effects on the cardiovascular system~\citep{X} or other harmful effects severe enough to derail the clinical development~\citep{X}, consecutively.
%
Finally, we apply Lilly medicinal filters~\citep{10.1021/jm301008n-lilly} (\lilly) to flag whether the molecules have undesirable traits, such as interfering with assay measurements (e.g., fluorescence) or lacking druggability (e.g., molecules lacking both oxygen and nitrogen). 
%
% Evaluating these properties on the generated molecules provides a comprehensive evaluation for \main. 
%
% In addition to the previously described metrics, we also compute the per-constraint pass rates (\%\vina, \%\qedd, \%\sascore, \%\mut, \%\clintox, \%\herg,
% \%\lilly) by measuring the fraction of output molecules satisfying that single property constraint for each target, then averaging them across targets.
%
% \rz{To evaluate whether the molecules can be experimentally validated during biological assay testing (e.g., not promiscuous and not damaging the proteins), we use Lilly medicinal filters. This part needs revision. }
%
% They use a structural demerit system that assigns a score conveying how "unlikable" the molecule is. 
%
% \textbf{Synthesizability}. 
% %
% The second set of constraints consider whether the molecule can be made in practice. 
% %
% Specifically, we consider whether a synthesis route can be generated for that molecule (\syn). 
We assess the synthesizability of a molecule by determining whether a synthesis route can be generated (\syn). 
We employ the same approach as \synthesizer in \main to measure the synthesizability of the molecule (Section~\ref{sec:method:synthesizer}).
%, with more details provided in Section~\ref{sec:method:synthesizer}. 
% , using two complementary strategies: retrosynthesis planning~\citep{zhao2024efficient-meea} and synthesizable analog generation~\citep{luo2026efficient-prexsyn}. 
%
% For each strategy, we use a state-of-the-art planning tool in both performance and efficiency~\citep{X,X}.  
% %
% A molecule is considered synthesizable if either tool can find a route.
%
% More details is provided in Section~\ref{sec:method:synthesizer}. 
%
This is a more rigorous assessment than synthesizability estimate (\sascore; Section~\ref{section:method:evaluator}). 
%as discussed in  Section~\ref{section:method:evaluator}. 
%
% As with molecular properties, we also compute the fraction of output molecules that satisfy the synthesizability constraint (\%\syn) for each target, then average them across targets.

% Specifically, we first apply a retrosynthesis model~\citep{zhao2024efficient-meea} to search for a route to that molecule. 
% %
% If no route is found, we use a synthesizable analog generation model~\citep{luo2026efficient-prexsyn} to find whether. 
%
% Combining the two models increase the chance of finding a synthesis route, since a route missed by one model can be recovered by the other, while still remaining tractable for large scale synthesizability assessment. 

%
% \rz{justification why this way}
% measure the binding affinity against the target pocket using Vina Docking Scores, following literature~\citep{}. Lower indicates stronger binding affinity. 
%
% For drug-likeness and safety, we report \emph{QED} (higher is better),
% \emph{SAScore} (SAS, lower is better), mutagenicity (\emph{MUT}),
% clinical toxicity (\emph{CTOX}), and hERG blockage (\emph{HERG}), all of which
% are toxicity risks for which lower is better, as well as the
% Lilly medicinal-chemistry rules (\emph{LILLY}).
%
% For synthesizability, we assess whether each molecule is retrosynthesizable
% (\emph{SYN}).

For every molecular property and synthesizability defined previously, we identify a threshold as a requirement for a molecule to meet. 
% \amcomment{ does the Syn get a "threshold" ?}
%
These thresholds are determined using existing FDA-approved drugs curated in our dataset as references (Appendix~\ref{supp:reqs}), 
% \amcomment{or just "These thresholds are informed by the properties of FDA-approved drugs associated with targets in out dataset."}
ensuring that they are realistic rather than arbitrary cutoffs. 
We define a molecule as \textit{satisfying} only if it \emph{jointly} meets all requirements. 

We report the performance of \main using several metrics: 
\textbf{(1)} \tgtout, the number of targets for which at least one output molecule is produced, as a measure of \main's coverage in delivering molecules for the targets;
\textbf{(2)} \out and \sat, the average number of output
and satisfying molecules per target, respectively;
\textbf{(3)} \%\pass, the average percentage of satisfying molecules among the output molecules per target;
% \yecomment{should we unify the \% in \pass and \%\vina?}
% \rzcomment{good idea}
%
and \textbf{(4)} the per-requirement pass rates (\%\vina, \%\qedd, \%\sascore, \%\mut, \%\clintox, \%\herg,
\%\lilly, \%\syn), by measuring the average percentage of output molecules satisfying each respective requirement per target. 

\paragraph{Baselines}
%
% Requires: \usepackage{booktabs,threeparttable,amssymb}
\begin{table*}[t]
  \centering
  \caption{The thresholds used in the experiments and whether each method can measure the corresponding properties and synthesizability during generation (\checkmark\ indicates it can).
  % that the method can measure \hl{them} \xia{what?}. 
  The thresholds are calculated using marketed FDA-approved drugs in our dataset (Appendix~\ref{supp:reqs}).}
  %, as discussed in Supplement~\ref{supp:reqs}.}
  % \caption{Property and synthesizability tools invoked during molecule design. \checkmark\ indicates that the method invokes the tool itself while designing its molecules, and -- that it does not, in which case we apply the tool as a post hoc filter afterwards and discard the molecules violating the corresponding constraint.}
  % \vspace{-4pt}
  \label{tbl:tools}
  \footnotesize
  \begin{threeparttable}
    \begin{tabular}{
      @{\hspace{6pt}}l@{\hspace{6pt}}
      @{\hspace{5pt}}c@{\hspace{5pt}}
      @{\hspace{5pt}}c@{\hspace{5pt}}
      @{\hspace{5pt}}c@{\hspace{5pt}}
      @{\hspace{5pt}}c@{\hspace{5pt}}
      @{\hspace{5pt}}c@{\hspace{5pt}}
      @{\hspace{5pt}}c@{\hspace{5pt}}
      @{\hspace{5pt}}c@{\hspace{5pt}}
      @{\hspace{5pt}}c@{\hspace{6pt}}
    }
      \toprule
      \multirow{2}{*}{Method} & \vina & \qedd & \sascore & \mut & \clintox & \herg & \lilly & \syn \\
      \cmidrule{2-9}
      & $<$-6.53 & $>$0.36 & $<$5.35 & $<$0.51 & $<$0.71 & $<$0.96 & True & True \\
      \midrule
      \multicolumn{9}{l}{\textit{Agents for Drug Discovery}} \\
      \midrule
      \quad \main (ours)& \checkmark & \checkmark & \checkmark & \checkmark & \checkmark & \checkmark & \checkmark & \checkmark \\
      \quad \liddia~\citep{averly-etal-2025-liddia} &
      \checkmark & \checkmark & \checkmark & -- & -- & -- & -- & -- \\
      % ==============================================================
      % \midrule
      % ==============================================================
      \quad Prompt-to-Pill~\citep{10.1093/bioadv/vbaf323-prompt-to-pill} &
      \checkmark & \checkmark & -- & \checkmark & -- & \checkmark & -- & -- \\
      % ==============================================================
      \quad AgentD~\citep{10.1021/acs.jcim.5c02454-agentd} &
      -- & \checkmark & -- & -- & -- & -- & -- & -- \\
      % ==============================================================
      \midrule
      \multicolumn{9}{l}{\textit{Generative Models for Drug Design}} \\
      \midrule
      % ==============================================================
      \quad Pocket2Mol~\citep{pmlr-v162-peng22b-pocket2mol} &
      -- & -- & -- & -- & -- & -- & -- & -- \\
      % ==============================================================
      \quad TargetDiff~\citep{guan3d-targetdiff} &
      -- & -- & -- & -- & -- & -- & -- & -- \\
      % ==============================================================
      \quad conDitar~\citep{gao2026generatingdevelopable3dmolecules-conditar} &
      -- & -- & -- & -- & -- & -- & -- & -- \\
      % ==============================================================
      \quad DiffSMol~\citep{chen2025generating-diffsmol} &
      -- & -- & -- & -- & -- & -- & -- & -- \\
      % ==============================================================
      % \midrule
      \quad ELILLM~\citep{10.1145/3774904.3792605-elillm} &
      -- & -- & -- & -- & -- & -- & -- & -- \\
      % ==============================================================
      \midrule
      % \multicolumn{9}{l}{\textit{LLM for Drug Discovery}} \\
      % \midrule
      % ==============================================================
      
      \multicolumn{9}{l}{\textit{Generalist LLMs}} \\
      \midrule
      % ==============================================================
      \quad Nemotron-3-Ultra~\citep{blakeman2026nemotron-nemotron3} &
      -- & -- & -- & -- & -- & -- & -- & -- \\
      % ==============================================================
      \quad Qwen3-30B~\citep{yang2025qwen3} &
      -- & -- & -- & -- & -- & -- & -- & -- \\
      % ==============================================================
      \bottomrule
    \end{tabular}
  \end{threeparttable}
\end{table*}
To evaluate the effectiveness of \main, 
we compare it with a wide range of baselines,
each cast as a \emph{generate-then-filter} pipeline: a molecular generation method followed by a filtering stage. 
The molecule generation methods span three categories:
\textbf{(1)} \emph{Agents for Drug Discovery}, \textbf{(2)} \emph{Generative Models for Drug Design} and \textbf{(3)} \emph{Generalist LLMs}.
% %
% \textbf{(1)} \textit{Agents for Drug Discovery}, including 
% \liddia~\citep{averly-etal-2025-liddia}, \mbox{Prompt-to-Pill}~\citep{10.1093/bioadv/vbaf323-prompt-to-pill} and AgentD~\citep{10.1021/acs.jcim.5c02454-agentd};
% % 
% \textbf{(2)} \textit{Generative Models for Drug Design}, including 
% Pocket2Mol~\citep{pmlr-v162-peng22b-pocket2mol}, TargetDiff~\citep{guan3d-targetdiff}, conDitar~\citep{gao2026generatingdevelopable3dmolecules-conditar},
% DiffSMol~\citep{chen2025generating-diffsmol},
% and ELILLM~\citep{10.1145/3774904.3792605-elillm};
% % 
% and \textbf{(3)} \textit{Generalist LLMs}, including 
% Nemotron-3~\citep{blakeman2025nvidia-nemotron3} and Qwen3-30B~\citep{yang2025qwen3}.
%
The molecules generated from each method are then filtered against each requirement, if the method cannot assess the corresponding property or synthesizability during generation. 
% After generation, molecules from each method are then filtered using a set of constraints, 
% depending on whether the method can compute that constraint during its generation.
%
Table~\ref{tbl:tools} presents which property or synthesizability each method can assess during its generation. 
%
%\hl{Every molecule generation method is filtered against \emph{every} requirement after generation, except agent for drug discovery baselines which are filtered against \emph{some} requirements. } \xia{need to be more precise here -- 
%what does it mean ``same requirements"?}
Every molecule from generative models for drug design and generalist LLMs is filtered against 
\emph{every} requirement after generation. 
In contrast, molecules from agents for drug discovery baselines are filtered only against  
the requirements that cannot be assessed by the method during generation.
Only molecules that pass the filtering are considered as the output molecules of that method. 
%
% Not all output molecules are satisfying, as presented in Table~\ref{tbl:overall-wout} and discussed in Section~\ref{sec:results:main}. 
%
Coupling a generation method with a filtering stage enables a fair comparison with \main, while also being a widely adopted paradigm in drug discovery~\citep{gao2026generatingdevelopable3dmolecules-conditar,zhavoronkov2019deep}. 

We use a variety of methods for each category of the baselines. 
\textbf{(1)} For \emph{Agents for Drug Discovery}, we compare against two state-of-the-art single-agent systems, \mbox{LIDDIA}~\citep{averly-etal-2025-liddia} and \mbox{AgentD}~\citep{10.1021/acs.jcim.5c02454-agentd}, and one multi-agent system, \prompttopill~\citep{10.1093/bioadv/vbaf323-prompt-to-pill}. 
These agents are designed for end-to-end drug discovery, similar to ours, and serve as direct points of comparison for \main. 
% \hl{We discuss more in details why the remaining multi-agent systems are not included in Supplement~{\ref{supp:baseline-selection}}.}
They are also selected since their implementations are publicly accessible (Appendix~\ref{supp:baseline-selection}).
\textbf{(2)} For \emph{Generative Models for Drug Design}, we compare against a wide variety of models: \mbox{Pocket2Mol}~\citep{pmlr-v162-peng22b-pocket2mol}, \mbox{TargetDiff}~\citep{guan3d-targetdiff}, \mbox{conDitar}~\citep{gao2026generatingdevelopable3dmolecules-conditar}, \mbox{DiffSMol}~\citep{chen2025generating-diffsmol} and \mbox{ELILLM}~\citep{10.1145/3774904.3792605-elillm}. 
These models include both SBDD (\mbox{Pocket2Mol}, \mbox{TargetDiff}, \mbox{conDitar}, \mbox{ELILLM}) and LBDD (\mbox{DiffSMol}) approaches.
% \st{We treat conDitar as the strongest baseline as it reports both strong performance in predicted binding affinity and through experimental validation}~\citep{gao2026generatingdevelopable3dmolecules-conditar}. 
%
\textbf{(3)} For \emph{Generalist LLMs}, we compare against two open-source LLMs, \mbox{Qwen3-30B}~\citep{yang2025qwen3} and \mbox{Nemotron-3-Ultra}~\citep{blakeman2026nemotron-nemotron3}. 
% both of which are competitive across general reasoning and scientific benchmarks~\citep{yang2025qwen3,blakeman2026nemotron-nemotron3}. 
We focus our comparison of \main to open-source LLMs for reproducibility and accessibility: their weights are publicly available and are less likely to be discontinued than proprietary LLMs. 
Each LLM is competitive on general reasoning and scientific benchmarks~\citep{yang2025qwen3,blakeman2026nemotron-nemotron3}, making them strong baselines. 
\mbox{Qwen3-30B} is also used as the reasoning engine for \main, while \mbox{Nemotron-3-Ultra} (550B) is among the largest and state-of-the-art open-source LLMs available.
% The two LLMs are selected since they are both competitive across general reasoning and scientific benchmarks~\citep{yang2025qwen3,blakeman2026nemotron-nemotron3} while occupying opposite ends in model size: 
% \mbox{Qwen3-30B} represents a compact LLM that can be deployed on a few local machines, 
% while \mbox{Nemotron-3-Ultra} is among the largest and state-of-the-art open-source LLMs available~\citep{blakeman2026nemotron-nemotron3}.
We only select two due to our experimentation budget.

Each method is instructed to produce the same fixed number of output molecules per target as \main for a fair comparison.
% \amcomment{I'm not sure I understand why fixed number is emphasized here?}
%
Methods that cannot be instructed in this way, 
including the five generative models for drug design and \mbox{AgentD}, 
are instead instructed to produce 
% \textit{as many} 
as many \textit{intermediate} molecules as \main produces, per target.
% during \hl{its} 
% entire generation process, per target.   
% W
Specifically,
we define an \textit{intermediate} molecule as any molecule that \hitidentifier, \leadoptimizer, and \synthesizer produce during a campaign, whether or not they serve as an output molecule. 
% \amcomment{just want to double check that "output molecule" is used consistently}
%
The number of \textit{intermediate} molecules is used as a basis for how many molecules the generative models and \mbox{AgentD} should generate, and also for analysis (Section~\ref{sec:results:ablation}).
%
% \yecomment{Should we need define the \textit{intermediate} molecules in formal? It will be used many times in the experimental results.} \rzcomment{Currently it's used in "Ablation Study" and "Intermediate Molecule Analysis". If we move the last one to Supplement, there's only one section that uses \textit{intermediate} molecules. We can formally define it there.}
% The number reflects how many \textit{intermediate} molecules \main produces. 
%
This ensures that these baselines receive the same opportunities as \main to explore the chemical space (by generating the same number of molecules), thus supporting a 
% fairer 
fair comparison.

\paragraph{Implementation Details}
%
% \rzcomment{Let's put the first paragraph here, and then the rest of implementation details at Supplement. First paragraph is required because we compare \main (Qwen3-30B) vs Qwen3-30B. Second and third paragraph will make more sense after Method section.}
%
In \main, we adopt \mbox{Qwen3-30B} \citep{yang2025qwen3} as the backbone LLM for \orchestrator and all four worker agents, an open-source LLM with strong performance across varying tasks while also small enough to be deployed locally.
The model is deployed with vLLM~\citep{kwon2023efficient-vllm} and decoded with a temperature of 0.7, a top-$p$ of 0.8, and a context window of 64K tokens. 
These are the recommended parameters for \mbox{Qwen3-30B}~\citep{yang2025qwen3}. 
We also provide implementation details for the baselines (Appendix~\ref{supp:baseline-imp}), along with more details on \main (Appendix~\ref{supp:main-imp}).
\section{Experimental Results}\label{section:results}
%%%%%%%%%%%%%%%%%%%%%%%%%%%%%%%%%%%

% \ye{
% Discussion: \\
% (1) List the experimental results we'd like to show. What observations / conclusions we want to discuss or demonstrate.\\
% (2) Suggested experiments are similar to LIDDIA1's case studies illustrated in Figure 2,3,4,6,7 in LIDDIA1.\\
% (3) Find multiple representative but different cases to show the workflow of \main.\\
% (4) Ablation study: remove the \leadoptimizer and \synthesizer, and a sequential variant of \main.\\
% (5) Compare the output molecules with approved drugs. (For example, for the drugs whose docking vina is above the threshold, we could analyze it.)\\
% (6) For a particular target with only one drug, we could analyze it.\\
% (7) Computational cost.\\
% (8) Expert evaluation.\\
% (9) Case study: binding.\\
% (10) The distribution analysis on the sources of output molecules (\hitidentifier, \leadoptimizer, or \synthesizer).\\
% }

% \ye{
% Priority Discussion:
% \begin{itemize}
%     \item Main Results (\ref{sec:results:main})
%     \item Ablation Study (\ref{sec:results:ablation}), Agent Analysis (\ref{sec:results:agent})
%     \item Computational Cost (\ref{sec:results:cost}), Expert Evaluation (\ref{sec:results:expert})
% \end{itemize}
% }

% \newpage
%===================================================
\subsection{Overall Performance}
\label{sec:results:main}
%===================================================
% Requires: \usepackage{booktabs,multirow,threeparttable}
\begin{table*}[t]
  \centering
  \caption{Main experimental results of \main against baselines.}
  \label{tbl:overall-wout}
  \footnotesize
  \begin{threeparttable}
    \begin{tabular}{
      @{\hspace{1.5pt}}l@{\hspace{-3pt}}
      @{\hspace{-3pt}}r@{\hspace{7pt}}
      @{\hspace{7pt}}r@{\hspace{1pt}}
      @{\hspace{1pt}}r@{\hspace{1pt}}
      @{\hspace{1pt}}r@{\hspace{7pt}}
      @{\hspace{7pt}}r@{\hspace{1.5pt}}
      @{\hspace{1.5pt}}r@{\hspace{1.5pt}}
      @{\hspace{1.5pt}}r@{\hspace{1.5pt}}
      @{\hspace{1.5pt}}r@{\hspace{1.5pt}}
      @{\hspace{1.5pt}}r@{\hspace{0.5pt}}
      @{\hspace{0pt}}r@{\hspace{1.5pt}}
      @{\hspace{1.5pt}}r@{\hspace{1.5pt}}
      @{\hspace{1.5pt}}r@{\hspace{1.5pt}}
    }
      \toprule
      % \multirow{2}{*}{Method} & \multirow{2}{*}{\#Tgt$_{\text{out}}$\tnote{a}} & \multicolumn{11}{c}{Average value per target with outputs} \\
      % \cmidrule(l){3-13}
      Method & \tgtout & \out & \sat & \%\pass & \%\vina & \%\qedd & \%\sascore & \%\mut & \%\clintox & \%\herg & \%\lilly & \%\syn \\
      % \midrule
      \midrule
      \multicolumn{13}{l}{\textit{Agents for Drug Discovery}} \\
      \midrule
      \quad\; \main & 50 & 100.0 & 100.0 & 100.0 & 100.0 & 100.0 & 100.0 & 100.0 & 100.0 & 100.0 & 100.0 & 100.0 \\
      \quad\; \liddia  & 49 & 36.0 & 28.0 & 80.0 & 83.1 & 96.1 & 100.0 & 100.0 & 100.0 & 100.0 & 100.0 & 100.0 \\
      \quad\; Prompt-to-Pill  & 31 & 14.2 & 2.9 & 20.9 & 55.0 & 88.5 & 100.0 & 41.0 & 100.0 & 98.1 & 100.0 & 100.0 \\
      \quad\; AgentD & 35 & 44.2 & 44.2 & 100.0 & 100.0 & 100.0 & 100.0 & 100.0 & 100.0 & 100.0 & 100.0 & 100.0 \\
      \midrule
      \multicolumn{13}{l}{\textit{Generative Models for Drug Design}} \\
      \midrule
      \quad\; Pocket2Mol  & 50 & 33.1 & 33.1 & 100.0 & 100.0 & 100.0 & 100.0 & 100.0 & 100.0 & 100.0 & 100.0 & 100.0 \\
      \quad\; TargetDiff  & 48 & 15.8 & 15.8 & 100.0 & 100.0 & 100.0 & 100.0 & 100.0 & 100.0 & 100.0 & 100.0 & 100.0 \\
      \quad\; conDitar  & 49 & 56.7 & 56.7 & 100.0 & 100.0 & 100.0 & 100.0 & 100.0 & 100.0 & 100.0 & 100.0 & 100.0 \\
      \quad\; DiffSMol  & 27 & 24.7 & 24.7 & 100.0 & 100.0 & 100.0 & 100.0 & 100.0 & 100.0 & 100.0 & 100.0 & 100.0 \\
      \quad\; ELILLM  & 27 & 4.5 & 4.5 & 100.0 & 100.0 & 100.0 & 100.0 & 100.0 & 100.0 & 100.0 & 100.0 & 100.0 \\
      \midrule
      
      % \multicolumn{13}{l}{\textit{LLM for Drug Discovery}} \\
      % \midrule
      % ELILLM  & 27 & 4.5 & 4.5 & 100.0 & 100.0 & 100.0 & 100.0 & 100.0 & 100.0 & 100.0 & 100.0 & 100.0 \\
      % \midrule
      \multicolumn{13}{l}{\textit{Generalist LLMs}} \\
      \midrule
      \quad\; Nemotron-3-Ultra  & 38 & 23.2 & 23.2 & 100.0 & 100.0 & 100.0 & 100.0 & 100.0 & 100.0 & 100.0 & 100.0 & 100.0 \\
      \quad\; Qwen3-30B  & 3 & 8.0 & 8.0 & 100.0 & 100.0 & 100.0 & 100.0 & 100.0 & 100.0 & 100.0 & 100.0 & 100.0 \\
      \bottomrule
    \end{tabular}
    % \begin{tablenotes}[flushleft]
    %   \footnotesize
    %   \item[a] .
    % \end{tablenotes}
  \end{threeparttable}
\end{table*}
Table~\ref{tbl:overall-wout} reports the performance of \main and the baselines.
For each target, every method is instructed to produce 100 output 
% \amcomment{repeating again that every method is instructed this way but over half are not? We are now talking about molecules you are evaluating now not generating "For each target, we evaluate up to 100 molecules per method, a number.... For methods that support requested output count, we evaluate the returned molecules and for the remaining methods we select the top 100 molecules or less if less available. (or whatever the procedure is) }
% \yecomment{maybe use a hyper-parameter?} \rzcomment{I don't think this is a hyperparameter, it's more like a task} 
molecules, a number large enough for further exploration (e.g., downstream selection for wet-lab validation) and small enough to keep evaluation tractable. 
In this setting, we compute the metrics only on targets for which a method produces at least one output molecule (e.g., generated molecules after filtering for baselines), ensuring fairness across methods. 
We also compare the property values of molecules generated by \main against baselines, the reference ligands, and our curated FDA-approved drugs (Appendix~\ref{supp:prop_analysis}). 

\textbf{\textit{Only \main delivers the requested 100 satisfying molecules on all 50 targets.}}
%
% \main is the only method that reaches the requested number on every target. %thereby saturating the maximum attainable under the evaluation.
%
The closest baselines fall well short: \mbox{conDitar} delivers 56.7 satisfying molecules over the 49 targets it covers, \mbox{Pocket2Mol} 33.1 over 50, and \mbox{AgentD} 44.2 over 35.
This demonstrates \main's advantage in consistently finding satisfying molecules across targets.
Such consistency arises from \main's multi-agent design: 
\main can adaptively generate, optimize, and evaluate according to what the current molecules lack, using its \orchestrator to decide which worker agent to delegate next. 
% \am{the next task to}.  
%
This forms a dynamic decision-making loop that improves molecules over time, thereby succeeding on a variety of targets. 
% % 
% in \main, \orchestrator delegates each subtask to \hitidentifier, \evaluator, \leadoptimizer or \synthesizer according to what the target currently lacks.
% %
% This forms a dynamic closed loop that revises the plan as evidence accumulates, thus succeeding on targets of widely differing difficulty.

% \textbf{\textit{\main delivers the requested 100 satisfying molecules on all 50 targets.}}
% %
% \main is the only method that reaches the requested number on every target, thereby saturating the maximum attainable under the evaluation.
% %
% The closest baselines fall well short: conDitar delivers 56.7 satisfying molecules over the 49 targets it covers, Pocket2Mol 33.1 over 50, and AgentD 44.2 over 35.
% %
% This demonstrates the advantage of \main in finding satisfying molecules consistently across various targets.
% %
% Such consistency arises from \main's multi-agent design:
% % 
% rather than following a fixed pipeline,
% % 
% in \main, \orchestrator delegates each subtask to \hitidentifier, \evaluator, \leadoptimizer or \synthesizer according to what the target currently lacks.
% %
% This forms a dynamic closed loop that revises the plan as evidence accumulates, thus succeeding on targets of widely differing difficulty.

\textbf{\textit{Agentic baselines employ evaluation tools during generation but still leave requirements unsatisfied.}}
As illustrated in Table~\ref{tbl:tools}, each agent is capable of measuring some properties for the requirements: \mbox{LIDDIA} measures \vina, \qedd and \sascore during its campaign, \prompttopill measures \vina, \qedd, \mut and \herg, and \mbox{AgentD} measures \qedd.
For \mbox{LIDDIA} and \prompttopill, however, the per-requirement pass rates still fall short on exactly these properties,
83.1\% on \vina for \liddia and 41.0\% on \mut for \prompttopill.
%
% Such failures arise largely from the LLM itself. 
These failures likely arise from the reasoning engine powering these agents: the LLMs themselves.
LLMs may misinterpret the tool outputs or hallucinate, leading them to accept a molecule that does not satisfy the requirements---a well-known weakness of LLM-based agent systems~\citep{lin2025llm}.
% \amcomment{change the tense of this- LLMs may misinterpret tool outputs or hallucinate, leading them to accept molecules that do not satisfy the requirements---a well-known weakness of LLM-based agent systems~\citep{lin2025llm}.}
% the instability of an LLM in interpreting tool outputs presented as text, 
% 
% where a misread or hallucinated value leads the LLM to accept a molecule that does not in fact satisfy the constraint --- a well-known weakness of LLM-based agent systems \ye{[Citation]}.
%
In contrast, \main addresses this weakness by accepting a molecule only after it has been evaluated by \evaluator and \synthesizer (Section~\ref{section:method}).
% \amcomment{overall note- check for verb tense when describing methods and unify. goes btwn past/present sometimes.}

\textbf{\textit{Generative models for drug design achieve perfect passing rates by filtering, but retain far fewer molecules than requested.}}
For these baselines, filtering is applied against every requirement after generation, 
%since they don't have access to any evaluation tools during generation. 
and molecules failing any requirement are discarded, leading to 100\% pass rates for every target (with at least one output molecule).
Filtering, however, cannot improve on what the methods produce: \mbox{conDitar} and \mbox{Pocket2Mol} retain only 56.7 and 33.1 satisfying molecules on average, respectively, both well below the requested 100.
When the method rarely produces molecules meeting the requirements, filtering may retain none of them, as with \mbox{DiffSMol} for 23 of the targets.
This exposes a fundamental limitation for methods without feedback from tools during generation: the pass rate can always be enforced to 100\% through filtering, but the effective yield, or 
% \hl{how many number of molecules survive the filtering} 
the number of satisfying molecules cannot be improved.

\textbf{\textit{Generalist LLMs produce very few satisfying molecules.}}
%
% ELILLM, which is specially designed for drug discovery, covers only 27 targets and delivers 4.5 satisfying molecules on them, fewer than the general-purpose Nemotron-3 with 38 targets and 23.2 molecules.
%
For instance, \mbox{Nemotron-3-Ultra}, one of the \mbox{state-of-the-art} \mbox{open-source} LLMs~\citep{blakeman2026nemotron-nemotron3}, produces on average only 23.2 satisfying molecules on 38 targets.  
%
% These models are not equipped with evaluation tools, and filtering further reveals that for many targets the produced molecules are not satisfying.
% 
% None of these models can compute a property while generating, so their molecules are filtered only after the fact, and for many targets none survives.
%
% \amcomment{maybe: "These models are not equipped with evaluation capabilities, so in post-generation filtering it is revealed that for many targets the produced molecules are unsatisfying."  }
%
\mbox{Qwen3-30B}, the LLM that drives every agent in \main, covers only 3 targets on its own, emphasizing that the gap to the 50 reached by \main is attributable to its design rather than solely to the capability of the underlying language model.
\subsection{Ablation Study}
\label{sec:results:ablation}
%===================================================

We conduct ablation experiments to assess the effectiveness of the different agents in the \main framework. 
Specifically, we remove \synthesizer (Section~\ref{sec:method:synthesizer}), \leadoptimizer (Section~\ref{section:method:leadoptimizer}) and \orchestrator (Section~\ref{section:method:orchestrator}) individually, 
resulting in three variants of \main: 
\textbf{(1)} \mbox{\textit{w/o \synthesizer}}, without the capability to assess synthesizability; 
\textbf{(2)} \mbox{\textit{w/o \leadoptimizer}}, which cannot optimize any property; 
and \textbf{(3)} \mbox{\textit{w/o \orchestrator}}, which only allows a fixed workflow during a campaign, respectively. 
In \main \textit{w/o {\orchestrator}}, we set the workflow such that each agent is invoked sequentially in this order: {\hitidentifier}, {\leadoptimizer}, {\evaluator}, then {\synthesizer}.
%
%\hl{In \textit{w/o {\synthesizer}}, no synthesis route is sought and the candidate molecules are returned without any synthesizability assessment.
%%
%In \textit{w/o {\leadoptimizer}}, a candidate that fails a property constraint can no longer be optimized, and {\main} must solely rely on {\hitidentifier} for molecule generation.
%%
%For \textit{w/o {\orchestrator}}, we set the workflow such that each agent is invoked in a fixed cyclic order: {\hitidentifier}, {\leadoptimizer}, {\evaluator}, then {\synthesizer}, sequentially. } \xia{combine and simplify the discussion here with 
%the above ``remove \synthesizer", etc., descriptions}
%
The workflow for \main \mbox{\textit{w/o {\orchestrator}}} 
% \hl{The process} \xia{it is not clear what ``the" process refers to} 
is repeated until the requested number of satisfying molecules is obtained. 
We do not remove \hitidentifier as it serves as the starting point of a campaign and \evaluator as every candidate molecule in \main is required to pass that worker agent. 
% \yecomment{I feel here we do not need the (1) (2) idx.}
%
% For this experiment, we also compute the number of \textit{intermediate} molecules from \main, defined as any molecule that \hitidentifier, \leadoptimizer and \synthesizer produces during a campaign, whether or not it serves as the output molecule. 
%
%\amcomment{why should the reader care about this? Based on conclusions in following paragraphs, we are using it as a proxy for efficiency. maybe this should be primed to the reader here.}
%
We measure \interm, the average number of intermediate molecules generated per target, and we provide the results in Table~\ref{tbl:ablation}. 
% In this subsection, we conduct ablation experiments to assess the effectiveness of the different components in the \main framework.
% %
% We remove \synthesizer (Section~\ref{sec:method:synthesizer}), \leadoptimizer (Section~\ref{section:method:leadoptimizer}) and \orchestrator (Section~\ref{section:method:orchestrator}) in turn, resulting in three variants: \textit{-w/o \synthesizer}, \textit{-w/o \leadoptimizer} and \textit{-w/o \orchestrator}, respectively.
% %
% Specifically, in \textit{-w/o \synthesizer}, no synthesis route is sought and the candidate molecules are returned without any synthesizability assessment.
% %
% In \textit{-w/o \leadoptimizer}, a candidate that fails a constraint can no longer be optimized and has to be replaced by a newly generated one.
% %
% In \textit{-w/o \orchestrator}, the agents are invoked in the fixed cyclic order “\hitidentifier $\rightarrow$ \leadoptimizer $\rightarrow$ \evaluator $\rightarrow$ \synthesizer”, repeated until the requested molecules are obtained, rather than dispatched according to the current state of the target.

% We provide the results in Table~\ref{tbl:ablation} and highlight several observations here:

% \textbf{\textit{\main generates less than half of the requested number of satisfying molecules when \synthesizer is removed.}} 

\textbf{\textit{\synthesizer is responsible for more than half of \main's satisfying molecules}}.
\main \mbox{\textit{w/o \synthesizer}} generates only 40.2 instead of 100 satisfying molecules, mostly because they fail the synthesis requirement (\%\syn).
% From generating 100 satisfying molecules on all targets, \main 
% \mbox{\textit{w/o \synthesizer}} generates only 40.2 on average afterward, mostly because they fail the synthesis requirement (\%\syn).  
%
\synthesizer also generates synthesizable analogs for candidate molecules without a route, and removing it causes \hitidentifier and \leadoptimizer to make up the shortfall by generating and optimizing more candidates. 
This is shown by the number of intermediate molecules generated: 1,554.8 molecules as opposed to 1,043.8 with \synthesizer on average. 

% \textbf{\textit{\main without \leadoptimizer generates almost 30\% more intermediate molecules}}.
\textbf{\textit{\leadoptimizer reduces the number of intermediate molecules \main generates by nearly 30\%}}.
Without it, the number increases from 1,043.8 to 1,327.7 as presented in Table~\ref{tbl:ablation}. 
In \main \textit{w/o \leadoptimizer}, a candidate molecule that fails a requirement cannot be improved further, forcing this variant to rely on \hitidentifier for generation.
In addition, for challenging targets that \hitidentifier struggles with (e.g., ABCD1\_ALD), \main 
\textit{w/o \leadoptimizer} cannot deliver the requested molecules, resulting in only covering 49 instead of 50 targets.  
% This removes \main's performance consistency, since for some challenging targets (e.g., \rz{XXX}), it has no option but to generate new molecules via \hitidentifier, limiting its adaptability and performance. 

% Requires: \usepackage{booktabs,multirow,threeparttable}
\begin{table*}[t]
  \centering
  \caption{Ablation study.}
  \label{tbl:ablation}
  \footnotesize
  \begin{threeparttable}
    \begin{tabular}{
      @{\hspace{6pt}}l@{\hspace{1pt}}
      @{\hspace{1pt}}r@{\hspace{7.5pt}}
      @{\hspace{7.5pt}}r@{\hspace{2.5pt}}
      @{\hspace{2.5pt}}r@{\hspace{2.5pt}}
      @{\hspace{2.5pt}}r@{\hspace{7.5pt}}
      % @{\hspace{5pt}}r@{\hspace{2.5pt}}
      @{\hspace{7.5pt}}r@{\hspace{7.5pt}}
      @{\hspace{7.5pt}}r@{\hspace{6pt}}
    }
      \toprule
      Method & \tgtout & \out & \sat & \%\pass & \%\syn & \interm \\
      \midrule
      \main                & 50 & 100.0 & 100.0 & 100.0  & 100.0 & 1{,}043.8 \\
      \quad \textit{w/o \synthesizer} & 50 & 100.0 & 40.2  & 40.2   & 40.2  & 1{,}554.8 \\
      \quad \textit{w/o \leadoptimizer}   & 49 & 100.0 & 100.0  & 100.0 & 100.0  & 1{,}327.7 \\
      \quad \textit{w/o \orchestrator}   & 50 & 100.0 & 100.0   & 100.0 & 100.0  & 1{,}743.8 \\
      \bottomrule
    \end{tabular}
    % \begin{tablenotes}[flushleft]
      % \footnotesize
      % \item[a] \interm: average number of intermediate molecules explored per target before producing the final outputs. \yecomment{After adding the formal definition in the text, could remove this footnote.}
      % \xia{Can someone up fix this instead of commenting?} \rzcomment{Revised}
    % \end{tablenotes}
  \end{threeparttable}
\end{table*}

% \textbf{\textit{Removing {\orchestrator} leads to the highest increase in the number of intermediate molecule generations}}.
% \hl{\textbf{\textit{{\orchestrator} contributes the largest reduction in intermediate molecule generations.}}}
\textbf{\textit{{\orchestrator} greatly decreases the number of molecules explored to find the satisfying molecules.}}
A campaign without \orchestrator requires generating 1,743.8 molecules on average, around 70\% more than with {\orchestrator}, to reach the same goal of 100 molecules satisfying all requirements. 
This gap reflects the role of \orchestrator: rather than generating molecules blindly, \orchestrator intelligently uses feedback from its prior decisions and accumulated molecules to decide what to generate next, concentrating its effort on molecules likely to satisfy the requirements. 
As a result, fewer molecules are required to explore, leading to a more efficient and effective method.

Taken together, these results show that every agent contributes by reducing the amount of exploration a campaign requires. 
They differ, however, in their role of reducing the intermediate molecules: 
\textbf{(1)} \synthesizer generates synthesizable analogs, allowing other agents to focus on the other requirements;
\textbf{(2)} \leadoptimizer improves existing candidates, providing a more efficient exploration for \main;
and \textbf{(3)} \orchestrator intelligently adapts to what existing molecules require, enabling \main to focus on the most promising candidate molecules.
\subsection{Campaign Dynamics Analysis}
\label{sec:results:agent}
%===================================================

We analyze how a campaign unfolds in \main: 
first through the transition probabilities across the worker agents aggregated over all the targets (Section~\ref{sec:results:agent:prob}), 
then through the flow of molecules from the worker agent that produces them to the output (Section~\ref{sec:results:agent:molecule_flow}), 
and finally through the trajectories of the worker agents on representative targets (Section~\ref{sec:results:agent:case}).
% In this subsection, we analyze how a campaign unfolds over the agents of \main: first through the transition probabilities aggregated over all the targets (Section~\ref{sec:results:agent:prob}), then through the flow of molecules from the agent that produces them to the output (Section~\ref{sec:results:agent:molecule_flow}), and finally through the trajectories on representative targets (Section~\ref{sec:results:agent:case}).

Each transition below is a delegation made by \orchestrator, 
% which decides the next subtask once the current worker agent returns its results.
%
and the agent itself is excluded in these figures as it serves as the central agent to which all worker agents must report. 
Only the four specialized worker agents are presented. 
% Each transition below is a delegation made by \orchestrator, which decides the next subtask once the current agent returns its results.
% %
% \orchestrator is not drawn in these figures, in which it would otherwise lie on every edge, and only the four specialized agents are shown.

% ------------------------------------------------------
\subsubsection{Overall Transition Probabilities} 
\label{sec:results:agent:prob}
% ------------------------------------------------------

\begin{figure}[ht]
    \centering
    \includegraphics[width=0.85\linewidth]{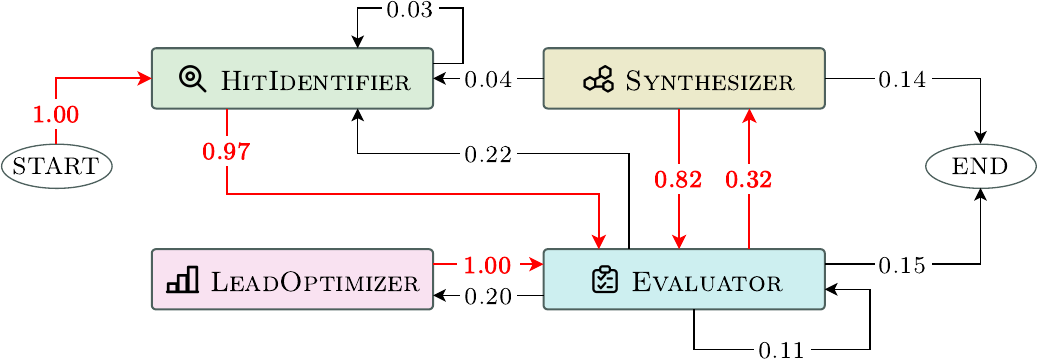}
    \vspace{5pt}
    \caption{
    Transition probabilities among worker agents during a campaign. 
    Red color represents the transition with the highest probability out of each worker agent.
    % \hl{Agent transition probabilities in \main} 
%    \xia{need a better term than ``agent transition probablities"}. \rzcomment{Revised (?)}
%    % 
%    % Transitions are calculated over all 50 targets.  
%    %the probability on an outgoing edge is the total number of times that transition occurs divided by the total number of invocations of the source agent, so the outgoing probabilities of each agent sum to one.
%    % 
%    % Red color marks the \hl{dominant} 
%    \xia{it is not accurate. need a better word} \rzcomment{Revised (?)}
%    % transition out of each worker agent.
%    % \yecomment{Delegation probabilities between \main's worker agents, aggregated over all 50 targets. Each edge gives the probability that one worker agent is followed by another in \orchestrator's delegation sequence. Red marks the most likely outgoing delegation from each worker agent.}
    } 
    % Every run starts at \hitidentifier and ends at SUCCESS.}
    \label{fig:transition_prob}
\end{figure}

Figure~\ref{fig:transition_prob} shows the empirical transition probabilities among the worker agents, calculated from the delegation decisions \orchestrator makes over the 50 targets.
Each edge represents a transition probability. 
% Figure~\ref{fig:transition_prob} shows the empirical transition probabilities between the agents, estimated from the delegation decisions that \orchestrator makes over the 50 targets.
% %
% Each edge gives the probability that one agent is followed by another, with every run starting at \hitidentifier and terminating at SUCCESS.

\textbf{\textit{\evaluator is the most frequently involved worker agent during a campaign.}}
% \textbf{\textit{\evaluator is the hub through which every worker agent passes, after \orchestrator.}}
%
After being invoked, \hitidentifier goes to \evaluator with probability 0.97, \leadoptimizer with probability 1.00 and \synthesizer with probability 0.82.
Every molecule that is generated, either by \hitidentifier or \synthesizer (e.g., synthesizable 
analogs), or optimized is therefore assessed before it is processed any further.
Any decision made in \main on which molecules to progress is attributed to the evaluated properties from \evaluator, emphasizing its role as a quality checkpoint for {\main}.
% \textbf{\textit{\evaluator is the central hub through which every agent passes.}}
% %
% After completing its task, \hitidentifier returns to \evaluator with probability 0.97, \leadoptimizer with probability 1.00 and \synthesizer with probability 0.82.
% %
% Every molecule that is generated, optimized or assigned a synthesis route is therefore assessed before it is used any further.
% %
% This is the empirical counterpart of the separation between production and verification on which \main is built.

% \textbf{\textit{Transitions out of \evaluator are almost equal in probability across worker agents}.}
% \textbf{\textit{Transitions out of \evaluator are \hl{spread???} almost evenly across worker agents.}}
% \fb{\textbf{\textit{\orchestrator adapts its delegation based on feedback from \evaluator}.}
\textbf{\textit{\main adapts its campaign based on feedback from \evaluator}}.
In Figure~\ref{fig:transition_prob}, 
transitions out of {\evaluator} (except to itself) are almost equal in probability across worker agents. 
Following \evaluator, the campaign returns to \hitidentifier to seek new molecules with probability 0.22, moves to \leadoptimizer to optimize one or more deficient properties with probability 0.20, or moves to \synthesizer to search for a synthesis route with probability 0.32.
% re-evaluates with probability 0.11, and terminates with probability 0.15.
%
%[TODO: @Frazier]
Since no transition probability dominates (out of \evaluator), a typical campaign in \main does not follow a fixed trajectory after evaluation.
Instead, \main adapts,
invoking the next worker agent according to \evaluator's feedback throughout a campaign.
% \fb{
% This indicates that the \orchestrator's delegation does not follow any one  single trajectory after evaluation by \evaluator.
% %
% Instead, the \orchestrator adapts the multi-agent workflow to the feedback from the \evaluator,
% leveraging each worker agent to solve distinct problems for the campaign.
% [Question: do the types of problems we would expect to be solved by each worker appear with (roughly) equal frequency?]
% }
% \hl{This indicates how {\main} can effectively adapt to different demands occurring during a campaign, since each worker agent is equally likely to be called next after {\evaluator}.}
% \xia{I do not get the point here... }
% \rz{Not yet.}\textbf{\textit{Control leaving \evaluator is distributed rather than concentrated on a single next step.}}
% %
% From \evaluator, the campaign returns to \hitidentifier to seek new hits with probability 0.22, moves to \leadoptimizer to optimize one or more deficient properties with probability 0.20, moves to \synthesizer to search for a synthesis route with probability 0.32, re-evaluates with probability 0.11, and terminates with probability 0.15.
% %
% No single next step accounts for more than a third of these transitions, and together with the near-deterministic returns to \evaluator this forms the closed loop that allows \main to reach the requested number of molecules on every target.

\textbf{\textit{Every campaign is guardrailed by either \evaluator or \synthesizer before it finishes}}. 
%
%\hl{SUCCESS} \xia{never defined; should not use it here} 
% is never reached directly from \hitidentifier or \leadoptimizer.
A campaign never finishes until either of them is called at the end.
This behavior follows similarly to real-world drug discovery, since the last step of a campaign 
is typically an assessment of properties and synthesizability, ensuring that the molecules are
grounded in evidence. 
\subsubsection{Molecule Flow Across Agents} 
\label{sec:results:agent:molecule_flow}
% ==============================================
\begin{figure}[h]
  \centering
  \begin{subfigure}[t]{0.49\textwidth}
    \centering
    \includegraphics[width=\linewidth]{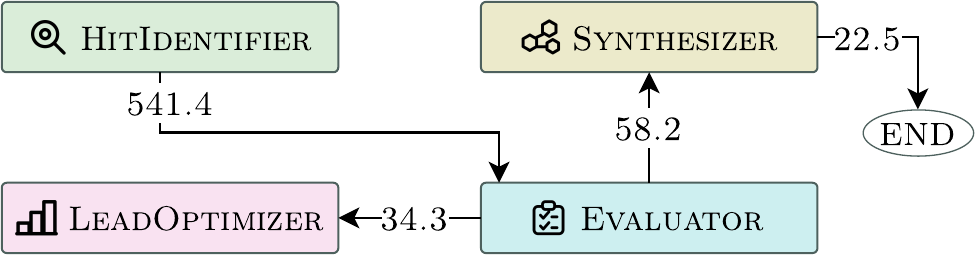}
    \caption{Molecules from \hitidentifier}
    \label{fig:hi}
  \end{subfigure}
  \hfill
  % \hspace{0.05\textwidth}
  \begin{subfigure}[t]{0.49\textwidth}
    \centering
    \includegraphics[width=\linewidth]{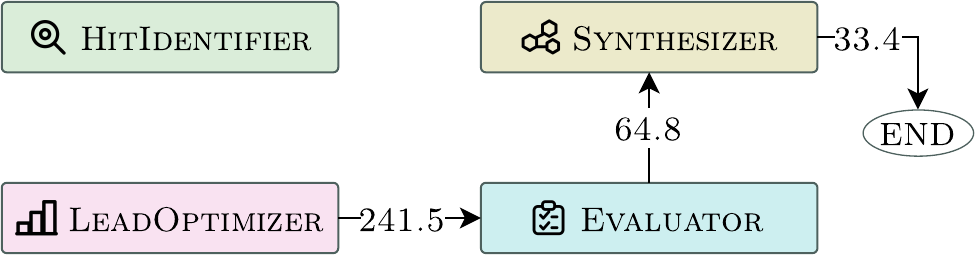}
    \caption{Molecules from \leadoptimizer}
    \label{fig:lo}
  \end{subfigure}
  % \hfill
  
  \vspace{10pt}
  \begin{subfigure}[t]{0.49\textwidth}
    \centering
    \includegraphics[width=\linewidth]{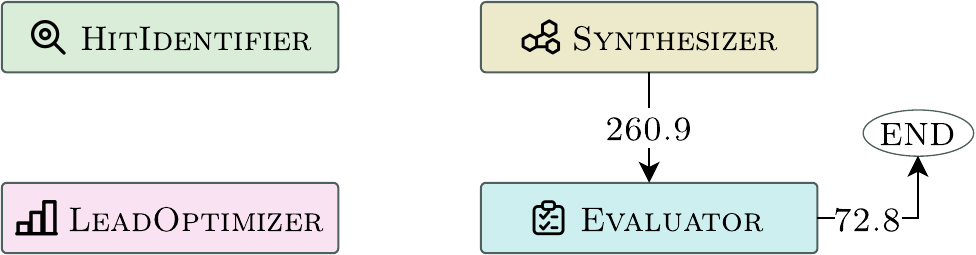}
    \caption{Molecules from \synthesizer}
    \label{fig:sy}
  \end{subfigure}
  \caption{Flow of molecules from the worker agent at which they originate to the campaign output. Each subfigure traces the molecules originating from one worker agent; each edge gives the number of those molecules subsequently passed to another worker agent or reaching the output, averaged over the 50 targets.
}
  \label{fig:flow}
  \vspace{-15pt}
\end{figure}

Figure~\ref{fig:flow} traces the molecules produced by \hitidentifier, \leadoptimizer, and \synthesizer during a campaign.
The figure illustrates how each worker agent contributes to the satisfying molecules, and how these worker agents collaborate to achieve those molecules.
% \hl{The counts represent the average number of molecules over all targets. } 
%
% We also compare the number of molecules stored in \orchestrator's \textit{synthesizable and property satisfying candidate set} $\mathcal{S}$ (Section~\ref{section:method:orchestrator}) and the output molecules as presented in Table~\ref{tbl:agent-source}.
%
% \#Mol. denotes the average number of molecules per target. 
%
% In some cases, \main can find more than 100 satisfying molecules in a campaign but returns only the first 100, since the task requires only 100 molecules (Section~\ref{sec:results:main}). 
% \rz{A campaign runs until at least 100 molecules satisfy all the constraints and the first 100 per target are then returned, so the counts in the figure are those of the satisfying molecules rather than of the molecules finally reported.}
% Figure~\ref{fig:flow} traces the molecules produced by each of \hitidentifier, \leadoptimizer and \synthesizer through the remainder of the campaign, with every molecule attributed to the agent that created it.
% %
% A campaign continues until at least 100 molecules satisfy all the constraints and the first 100 per target are then returned, so the counts in the figure are those of the satisfying molecules rather than of the molecules finally reported.

\textbf{\textit{\hitidentifier generates the most molecules but contributes the fewest satisfying ones.}}
\hitidentifier sends 541.4 molecules to \evaluator on average (Figure~\ref{fig:hi}), more than twice as many as {\leadoptimizer} and {\synthesizer}, which send 241.5 (Figure~\ref{fig:lo}) and 260.9 (Figure~\ref{fig:sy}), respectively.
Yet only 22.5 of them end up satisfying all the requirements, compared with 33.4 and 72.8 for the other two agents, respectively. 
% \amcomment{maybe using percentages here would be easier to understand: 4 percent vs 14 and 28 percent or agent molecules make it to output }
%
This contrast reflects a division of effort among these agents: \hitidentifier explores, providing a wide range of candidates other agents can use, while \leadoptimizer and \synthesizer exploit, using these candidates to generate molecules more likely to meet the requirements. 
It is through this coupling of exploration and exploitation that \main can produce satisfying molecules in its campaign. 

% This contrast reflects a division of effort among these agents: \hitidentifier explores, generating broadly at lower quality, whereas \leadoptimizer and \synthesizer exploit, concentrating on higher-quality candidate molecules more likely to meet the requirements. 
% \rz{Not yet.}\textbf{\textit{\hitidentifier generates the most molecules but contributes the fewest satisfying ones.}}
% %
% \hitidentifier sends 27,072 molecules to \evaluator, more than twice as many as \leadoptimizer and \synthesizer, which send 12,076 and 13,043, yet only 1,124 of them end up satisfying all the constraints, against 1,669 and 3,641 for the other two agents.
% %
% Exploration is thus concentrated at the entry of the campaign, whereas the molecules that survive to the end come mostly from the refinement stages.

\textbf{\textit{The satisfying molecules are drawn from all three agents, with \synthesizer contributing the largest share.}}
The three contributions are not independent: 
\leadoptimizer optimizes the molecules that \hitidentifier generates, and \synthesizer generates its analogs from molecules that \hitidentifier and \leadoptimizer produce, after being assessed by \evaluator.
\synthesizer produces the most satisfying molecules because it often acts last, generating analogs from candidate molecules that already satisfy most requirements.
Each agent thus contributes to the final molecules, either directly or through another agent. 
\subsubsection{Agent Execution Across Campaigns}
\label{sec:results:agent:case}

\begin{figure}[t]
  \centering
  \captionsetup[subfigure]{skip=1pt}
  \newlength{\tu}\setlength{\tu}{\dimexpr\textwidth/185\relax}
  \newcommand{\trajgap}{\hspace{10.37\tu}}

  \includegraphics[width=176.87\tu]{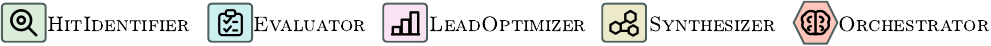}\\[3pt]

  \makebox[176.87\tu][l]{%
  \begin{subfigure}[b]{165.86\tu}
    \centering
    \includegraphics[width=\linewidth]{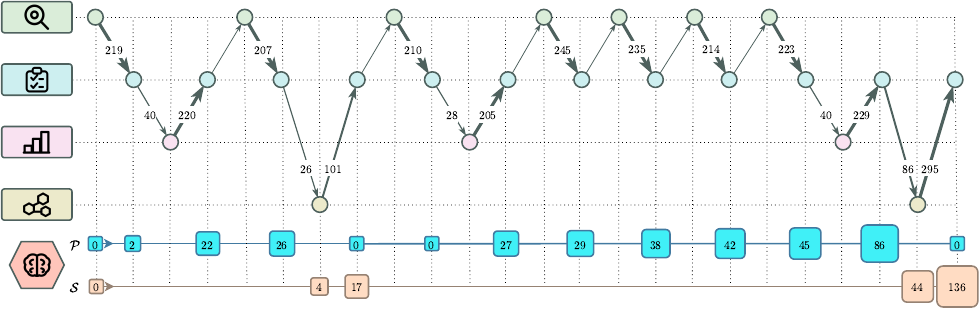}
    \caption{ABCD1\_ALD}\label{fig:flow-abcd1}
  \end{subfigure}}

  \vspace{-3pt}
  \begin{subfigure}[b]{80.48\tu}
    \centering
    \includegraphics[width=\linewidth]{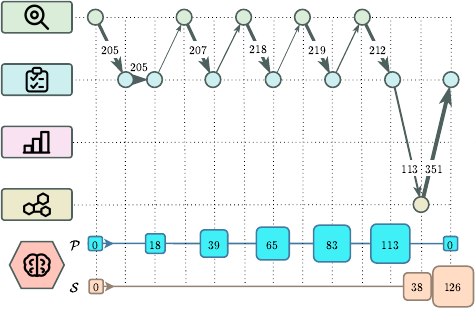}
    \caption{BCHE}\label{fig:flow-bche}
  \end{subfigure}%
  \trajgap%
  \begin{subfigure}[b]{86.02\tu}
    \centering
    \includegraphics[width=\linewidth]{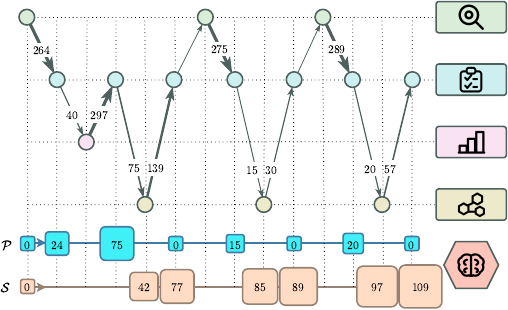}
    \caption{ADAM10}\label{fig:flow-adam10}
  \end{subfigure}

  \vspace{-3pt}
  \begin{subfigure}[b]{48.18\tu}
    \centering
    \includegraphics[width=\linewidth]{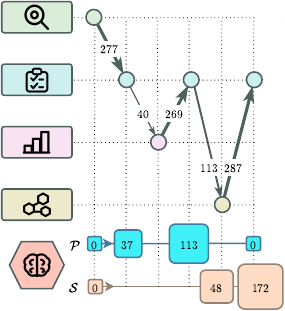}
    \caption{ADRB1}\label{fig:flow-adrb1}
  \end{subfigure}%
  \trajgap%
  \begin{subfigure}[b]{46.91\tu}
    \centering
    \includegraphics[width=\linewidth]{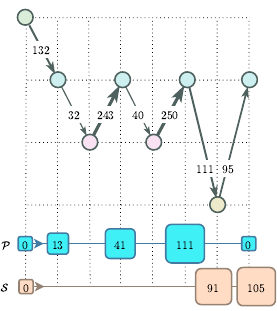}
    \caption{IDH2}\label{fig:flow-idh2}
  \end{subfigure}%
  \trajgap%
  \begin{subfigure}[b]{61.04\tu}
    \centering
    \includegraphics[width=\linewidth]{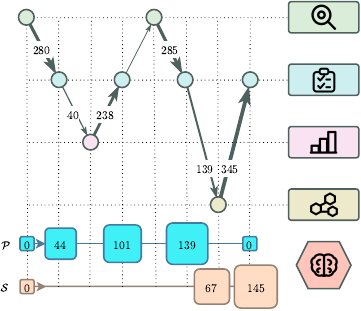}
    \caption{PAK4}\label{fig:flow-pak4}
  \end{subfigure}
  % \captionsetup[subfigure]{skip=1pt}
  % \includegraphics[width=\textwidth]{figure/case/trajectory/legend.pdf}\\[5pt]
  % \makebox[\textwidth][l]{%
  % \begin{subfigure}[b]{0.9843\textwidth}
  %   \centering
  %   \includegraphics[width=\linewidth]{figure/case/trajectory/ABCD1_ALD_flow.pdf}
  %   %\vspace{-6pt}
  %   \caption{ABCD1\_ALD}
  %   \label{fig:flow-abcd1}
  % \end{subfigure}
  % }

  % \vspace{-10pt}
  % \begin{subfigure}[b]{0.4776\textwidth}
  %   \centering
  %   \includegraphics[width=\linewidth]{figure/case/trajectory/BCHE_flow.pdf}
  %   \captionsetup{skip=-3pt}
  %   \caption{BCHE}
  %   \label{fig:flow-bche}
  % \end{subfigure}
  % \hfill
  % % \vspace{-6pt}
  % \begin{subfigure}[b]{0.5105\textwidth}
  %   \centering
  %   \includegraphics[width=\linewidth]{figure/case/trajectory/ADAM10_flow.pdf}
  %   \captionsetup{skip=-3pt}
  %   \caption{ADAM10}
  %   \label{fig:flow-adam10}
  % \end{subfigure}

  % \vspace{-10pt}
  % \begin{subfigure}[b]{0.2859\textwidth}
  %   \centering
  %   \includegraphics[width=\linewidth]{figure/case/trajectory/ADRB1_flow.pdf}
  %   % \vspace{-6pt}
  %   \caption{ADRB1}
  %   \label{fig:flow-adrb1}
  % \end{subfigure}
  % % \hspace{1.5em}% 
  % \hfill
  % \begin{subfigure}[b]{0.2784\textwidth}
  %   \centering
  %   \includegraphics[width=\linewidth]{figure/case/trajectory/IDH2_flow.pdf}
  %   % \vspace{-6pt}
  %   \caption{IDH2}
  %   \label{fig:flow-idh2}
  % \end{subfigure}
  % % \hspace{1.5em}% 
  % \hfill
  % \begin{subfigure}[b]{0.3623\textwidth}
  %   \centering
  %   \includegraphics[width=\linewidth]{figure/case/trajectory/PAK4_flow.pdf}
  %   % \vspace{-6pt}
  %   \caption{PAK4}
  %   \label{fig:flow-pak4}
  % \end{subfigure}
  \vspace{-5pt}
  \caption{
  Trajectories of \main over six representative targets: \textbf{(a)} ABCD1\_ALD, \textbf{(b)} BCHE, \textbf{(c)} ADAM10, \textbf{(d)} ADRB1, \textbf{(e)} IDH2, and \textbf{(f)} PAK4.
  Rows represent the agents in \main and columns represent one delegation or step during the campaign.
  % Each circle denotes one delegation to that worker agent. 
  The number on the arrow denotes the number of molecules passed to the next agent.
  The number on the box denotes the number of molecules stored in each repository after the delegation.
  The arrow width (worker agents) and box size (\orchestrator) are proportional to the number of molecules.
  % Transition trajectories of \main on six representative targets.
% Within each subfigure, steps run left to right and each row corresponds to one worker agent or to one of the two candidate sets ($\mathcal{P}$, $\mathcal{S}$), with every node and block placed at a gridline intersection. In the upper four rows, each node marks one invocation of a worker agent and each arrow one delegation issued by \orchestrator, which is omitted from the trajectory as every arrow originates from it; numbers on arrows give the molecules passed in that step, with arrow width proportional to that number. In the lower two rows, each block gives the size of $\mathcal{P}$ or $\mathcal{S}$ after that step, with block size proportional to the count. 
% \xia{The ChatGPT writing this caption is not a friend of the ChatGPT writing the manuscript, so it speaks in a very different style... need to unify. } \rzcomment{Revised}
% \yecomment{I think the rationale of previous writing is to give the explanation of each part shown in the figure. You need to revise it rather then directly remove it. If you feel it is not suitable, you can introduce this in the main text.....}
%\rzcomment{Other comments removed temporarily so figure can fit into the paper.}
}
  \label{fig:flow-cases}
  \vspace{-10pt}
\end{figure}

We further examine campaigns for different targets in \main to examine how they unfold. 
We also provide the number of molecules stored in $\mathcal{P}$ and $\mathcal{S}$ under \orchestrator (Section~\ref{section:method}), representing the number of molecules satisfying (1) all requirements except synthesizability, and (2) all requirements.
During a campaign, \main can generate more satisfying molecules than requested; in that case, it returns only the first satisfying molecules generated, up to the number requested. 

\textbf{\textit{\main adaptively employs different strategies throughout a campaign.}}
Figure~\ref{fig:flow-abcd1} follows a campaign against ABCD1\_ALD, a challenging target~\citep{zuo2024gene-abcd1} with no marketed FDA-approved drugs as of July 2026. 
% (Table~\ref{tbl:liddia2_drug_associations})
% along with having the longest trajectory (e.g., highest number of delegations).
%
Over successive rounds of delegation (by \orchestrator), generation (by \hitidentifier), optimization (by \leadoptimizer), and assessment (by \evaluator and \synthesizer), 
\main identifies a total of 136 satisfying molecules, demonstrating its capability.
%
%\amcomment{unify the "action word" for lead optimizer- is it generation or optimization?}
%\rzcomment{Revised}
%
We highlight several patterns of \main in this campaign: 
\textbf{(1)} the most common strategy is \emph{generate-then-optimize}, which invokes \hitidentifier followed by \leadoptimizer, with \evaluator assessing the molecules in between generations. 
\main applies this strategy twice in the first half of the campaign and once near the end;
\textbf{(2)} \orchestrator adapts its strategy throughout the campaign. 
Most notably, \orchestrator employs \emph{generate-then-optimize} during the first half, 
observes the results (from the molecules stored in $\mathcal{P}$ and $\mathcal{S}$), 
then adopts a new strategy, \emph{multi-generate-then-optimize}, which calls \hitidentifier multiple times before optimizing via \leadoptimizer at the end (with \evaluator in between). 
This strategy dominates the second half of the campaign, accumulating 86 property satisfying molecules in $\mathcal{P}$;
% \textbf{(2)} \orchestrator learns that this is not the most effective approach judging by the number of property satisfying molecules (first add 22, second add 27). 
% %
% Instead it learns and adapt its strategy, and employ a new strategy \textit{many-generate-then-filter}, which essentially runs \hitidentifier multiple times and then optimize at the end.
% %
% This is what happens during the second half of the campaign.
%
% These strategies leads to many property satisfying molecules, accumulating 86 molecules in $\mathcal{P}$ satisfying and ends with 136 satisfying molecules in total, after \synthesizer and \evaluator.  
%
\textbf{(3)} \evaluator is called after every generation---whether by \hitidentifier, \leadoptimizer or \synthesizer---emphasizing its importance;
\textbf{(4)} {\synthesizer} is the least frequently called agent (only twice). 
%
% However, it is only after {\synthesizer} that the molecules in $\mathcal{S}$ increases, and it is \synthesizer's synthesis analog generation that ultimately increases the number of satisfying molecules dramatically, from 44 to 136, thus ending the campaign. 
However, it is only after \synthesizer does $\mathcal{S}$ increase: 
% \hl{its} 
\synthesizer's
% \xia{what does ``its" refer to?} \rzcomment{Revised} 
synthesis analog generation drives the sharp increase in satisfying molecules, from 44 to 136, thus ending the campaign.
% However, it is only after \synthesizer's assessment that the number of molecules $\mathcal{S}$ increase, and it is only through \synthesizer's synthesis analog generation that it can increase the satisfying molecules massively from 44 to 136 at the end. 
%
This highlights \synthesizer as a critical agent despite its infrequent use.
Together, the campaign illustrates how worker agents, coordinated through \orchestrator's feedback-driven decisions, can solve a difficult task---generating 100 satisfying molecules for a target with no existing marketed FDA-approved drugs.
% \textbf{\textit{A single campaign combines all four agents under the coordination of \orchestrator.}}
% %
% Figure~\ref{fig:PAK4} follows the campaign on PAK4.
% %
% \hitidentifier first generates 280 molecules (step 1), of which \evaluator finds 44 satisfying the constraints (step 2).
% %
% \orchestrator then delegates 40 of them to \leadoptimizer, which returns 238 candidates, 57 of them satisfying (step 3 - 4), and a second call to \hitidentifier adds 285 molecules and a further 38 (step 5 - 6).
% %
% The 139 satisfying candidates accumulated so far are passed to \synthesizer, which finds routes for 67 of them and builds 345 analogs (step 7), 78 of which \evaluator finds satisfying (step 8).
% %
% The campaign ends with 145 satisfying molecules, of which the first 100 are returned (step 9).
% %
% The requested molecules are therefore assembled over several rounds and from more than one agent, rather than produced in a single pass.

% \rz{
% \textbf{\textit{Trajectory analysis on a target from the case study paragraph.}}}
%
% \rzcomment{This is a placeholder. There will be one more paragraph on a target in the case study.}

\textbf{\textit{\main dynamically adapts its campaign to different targets.}}
We present five more campaigns, each of which \main approaches with distinct strategies.
\textbf{(1)} The campaign on BCHE~\citep{pub.1069188445-bche,Pajk01012020-bche2}, presented in Figure~\ref{fig:flow-bche}, 
% \xia{refer to suvbfigure} 
% \rzcomment{Revised}
is notable in that \leadoptimizer is never called: 
\main relies solely on \hitidentifier to generate molecules, with \synthesizer
invoked only at the end to produce synthesizable analogs.
\textbf{(2)} For target ADAM10~\citep{10.3389/fimmu.2020.00499-adam10}, shown in Figure~\ref{fig:flow-adam10},
\main pursues two kinds of strategies:
\hitidentifier followed by \leadoptimizer and then \synthesizer, and \hitidentifier
followed directly by \synthesizer. 
Each strategy is interleaved with \evaluator between successive generations. 
% \hl{These strategies enable exploration beyond only using one specific strategy.}
The two strategies differ in whether \leadoptimizer is invoked, suggesting that \orchestrator observes whether the molecules require optimization and adapts accordingly.
%\xia{does this provide any concrete, meaningful insight? the insight is ``multiple" strategies 
%enable exploration beyond ``one" strategy? where is the notation ``beyond" from? is it better 
%than just one strategy? } \rzcomment{Revised}
%
\textbf{(3)} ADRB1, IDH2, and PAK4, each illustrated in Figure~\ref{fig:flow-adrb1},~\ref{fig:flow-idh2}, and~\ref{fig:flow-pak4} respectively, follow shorter trajectories with a similar strategy,
all beginning with \hitidentifier and then \leadoptimizer, but diverging after:
ADRB1 proceeds directly to \synthesizer and finishes, while   
IDH2 performs an additional round of optimization, and
PAK4 an additional round of generation, before each concludes with a
final \synthesizer call.
Together, these campaigns show that \main does not execute a fixed workflow, but instead, it adapts, mirroring the adaptive workflows a medicinal chemist typically employs in real-world drug discovery.
We discuss additional representative trajectories on two targets, ABO\_BGAT and HTR2A (Appendix~\ref{supp:addtl_trajs}).
\subsection{Case Study on PSEN1}
\label{section:results:case_study}
%===================================================
\begin{figure}[h]
  \centering
  \begin{subfigure}[t]{0.32\textwidth}
    \centering
    \includegraphics[width=\linewidth]{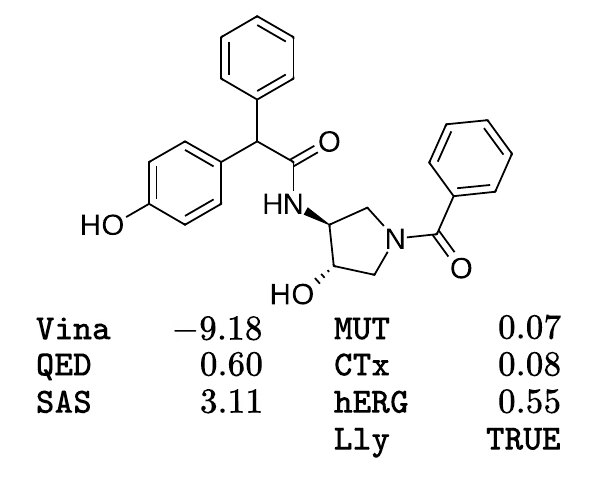}
    \caption{PSEN1-01}
    \label{fig:psen1-01}
  \end{subfigure}
  \hfill
  % \hspace{0.05\textwidth}
  \begin{subfigure}[t]{0.32\textwidth}
    \centering
    \includegraphics[width=\linewidth]{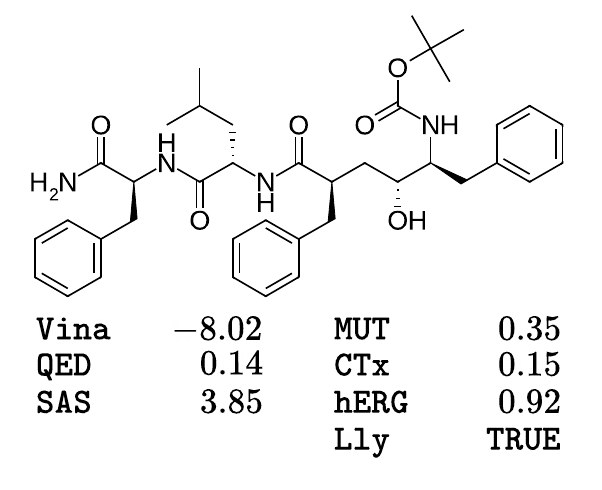}
    \caption{PSEN1 reference ligand}
    \label{fig:psen1-ref_ligand}
  \end{subfigure}
  \hfill
  \begin{subfigure}[t]{0.32\textwidth}
    \centering
    \includegraphics[width=\linewidth]{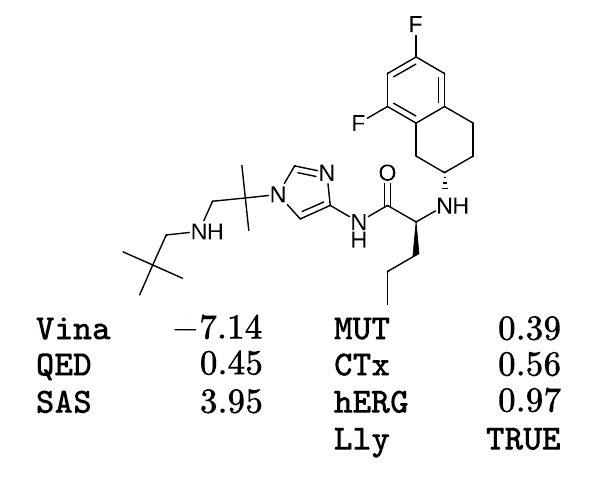}
    \caption{Nirogacestat}
    \label{fig:psen1-nirogacestat}
  \end{subfigure}
  
  \begin{subfigure}[t]{0.49\textwidth}
    \centering
    \includegraphics[width=\linewidth]{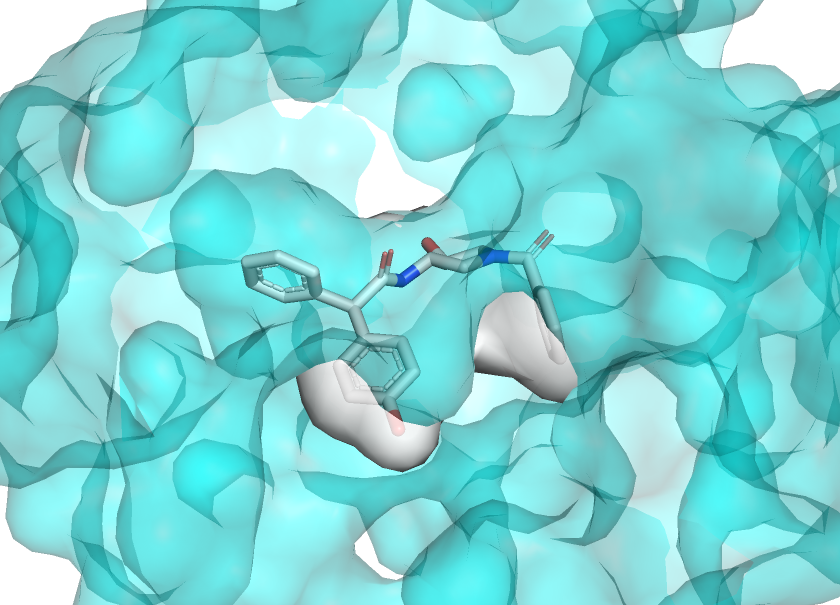}
    \caption{PSEN1-01 and PSEN1 pocket binding interaction (surface representation).}
    \label{fig:psen1-01-binding-cloud}
  \end{subfigure} 
  \hfill
  \begin{subfigure}[t]{0.49\textwidth}
    \centering
    \includegraphics[width=\linewidth]{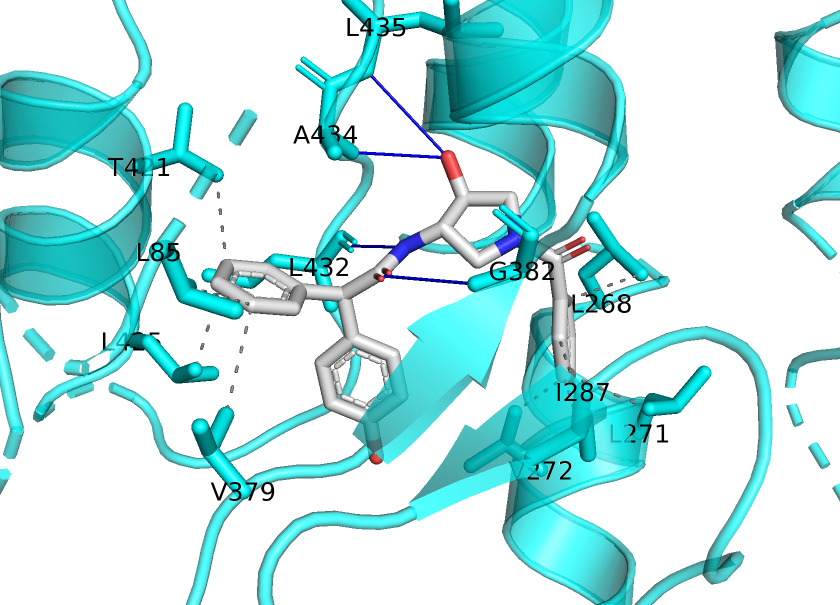}
    \caption{PSEN1-01 and PSEN1 pocket binding interaction (cartoon representation). Hydrogen bonds and hydrophobic interactions are denoted as solid blue and dashed gray lines, respectively.}
    \label{fig:psen1-01-binding-cartoon}
  \end{subfigure} 
  % \vspace{-20pt}
  \begin{subfigure}[t]{\textwidth}
    \centering
    \includegraphics[width=\linewidth]{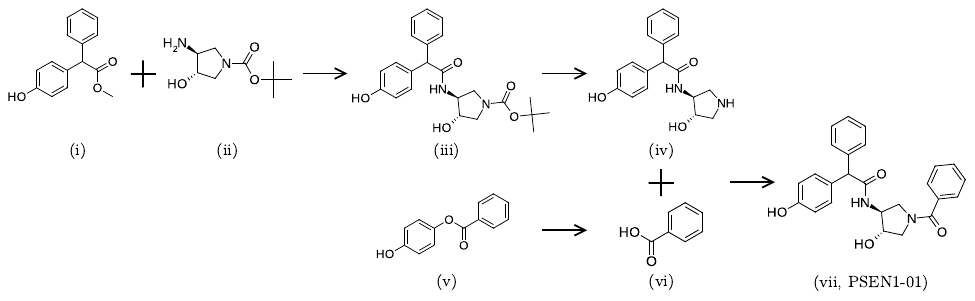}
    \vspace{-15pt}
    \caption{PSEN1-01 predicted synthesis route. Compounds~(i)--(vi)
denote the starting materials and intermediates along the route;
(vii) is the final product, PSEN1-01.}
    \label{fig:route}
  \end{subfigure} 
  \vspace{-8pt}
  \caption{PSEN1 case study.}
  \label{fig:psen1-case_study}
  \vspace{-15pt}
\end{figure}

To assess \main's generated molecules beyond computational predictions, we ask a domain expert to evaluate them.
%
% We focus on six targets associated with either of the two major disease areas: \textbf{(1)} PSEN1~\citep{X}, ACHE~\citep{X}, and BCHE~\citep{X} for neurological conditions, particularly Alzheimer's disease~\citep{X}, and \textbf{(2)} MET/HGFR~\citep{X}, FLT3~\citep{X}, and KIT~\citep{X} for cancer~\citep{X}.
We focus on three targets associated with Alzheimer's disease: PSEN1~\citep{ijms231810970-psen1}, ACHE~\citep{https://doi.org/10.1002/pro.70297-ache}, and BCHE~\citep{biom9100592-bche}.
We screen the molecules according to two criteria before expert evaluation.
The criteria are \textbf{(1)} whether their \textit{in silico} properties exceed those of reference ligands and at least one of the associated FDA-approved drugs for the target, and \textbf{(2)} whether their structures are chemically unstable through manual validation.
% \amcomment{how did we determine this?} 
%
We discuss the results for PSEN1 here, and the remaining targets in Appendix~\ref{supp:addtl_case_studies}.

\textbf{\textit{On PSEN1, \main proposes a
molecule that surpasses both the co-crystallized ligand and the approved drug across all six in silico properties}}. 
PSEN1 encodes the catalytic subunit of Gamma-secretase, an enzyme complex linked to Alzheimer's disease and certain tumors because it processes multiple critical membrane proteins, including the amyloid precursor protein (APP) and the cancer-related signaling protein~\citep{de2012presenilins-psen1}. 
Despite decades of medicinal chemistry~\citep{sherrington1995cloning-alzheimer}, there is only one marketed FDA-approved drug (as of July 2026) associated with PSEN1, Nirogacestat---originally developed for Alzheimer's disease, though currently it is used for the treatment of patients with desmoid tumors~\citep{gounder2023nirogacestat}.
Against this target, \main proposes 16 molecules with \textit{in silico} properties better than either the co-crystallized reference ligand or Nirogacestat on binding affinity (\vina), drug-likeness (\qedd), synthesizability (\sascore), and toxicity (\mut, \clintox, \herg). 
We illustrate all 16 molecules in Figure~\ref{fig:psen1_addtl_mols}.
%
% Some of the molecules are stereoisomers (e.g., PSEN1-01 and PSEN1-02) or regioisomers (e.g., PSEN1-04 and PSEN1-05) of one another;
% the remaining targets are structurally distinct, indicating that \main is capable of generating diverse molecules without diversity being imposed as an explicit requirement.
%
Notably, PSEN1-01 (Figure~\ref{fig:psen1-01}) surpass both the reference ligand (Figure~\ref{fig:psen1-ref_ligand}) and Nirogacestat (Figure~\ref{fig:psen1-nirogacestat}) across all properties \textit{in silico}, demonstrating its potential for this target.  
%
% \rz{Further discussion on the in silico properties}

\textbf{\textit{PSEN1-01 interacts favorably withing the binding site of PSEN1}}.
Figure~\ref{fig:psen1-01-binding-cloud} and~\ref{fig:psen1-01-binding-cartoon} illustrates the binding site and how PSEN1-01 interacts with it.  
PSEN1-01 buries the majority of its molecular surface into the site, with the 4-hydroxyphenyl ring left partially exposed. 
Notably, the molecule engages with the two sequence motifs of PSEN1: the hydroxyl on its pyrrolidine ring accepts hydrogen bonds with A434 and L435 of the PAL motif~\citep{Sato6264-psen1-pal}, while the acetamide carbonyl accepts from G382 of the GxGD motif~\citep{Yamasaki3821-psen1-gxgd}.   
These hydrogen bonds are complemented by nine hydrophobic contacts that is distributed over two distinct subsites: the pendant phenyl against L85, V379, T421 and L425, and the benzoyl phenyl against L268, L271, V272 and I287 (2 contacts).
Together, these interactions account for the binding improvement of PSEN1-01 over both the reference ligand and Nirogacestat,
and show that it can achieve the shape complementarity required by this binding site
% satisfy the shape complementarity this binding site demands 
while still satisfying other key properties.

\textbf{\textit{\main proposes a four-step synthesis route for synthesizing PSEN1-01}}. 
% \yecomment{@Frazier, could you double check the content in this paragraph? I do not have much knowledge about the synthesis route, but I asked Claude and it seems two things may be off:
% ``1. The stereodescriptors should be (3S,4S) rather than (3R,4R) — in both places.
% 2. The benzoic acid precursor is 4-hydroxyphenyl benzoate, not phenyl 4-hydroxybenzoate (the acyl half is benzoyl, the alcohol half is hydroquinone). If so, plain hydrolysis already gives benzoic acid and the "deoxygenation" shouldn't be there.''
% Just wanted to make sure there are no errors before our submission.
% }
% \fb{[I copied the phenyl 4-hydroxybenzoate from our slack discussion and only did a cursory check, missing the switch of the alcohol.  you are correct that the alcohol is on the other side.  I have fixed this.]
% [For (ii), I see Claude's point, but I need to check what flips may have been done during figure editing, as this can affect stereochemistry.] [update: after checking the original SMILES shared in slack, 3S,4S is accurate.  The current figure also matches 3S,4S.]}
% \yecomment{Also, you can update the writing with the idx of compounds in Figure~\ref{fig:route}.}
%
Figure~\ref{fig:route} presents the route,
starting with an ester aminolysis to join
\iupac{methyl 2-(4-hydroxy\allowbreak{}phenyl)-2-phenyl\allowbreak{}acetate} (i)
and
\iupac{(3S,4S)-3-amino-1-Boc-4-hydroxy\allowbreak{}pyrrolidine} (ii)
to form
\iupac{tert-butyl \allowbreak{}(3S,4S)-3-hydroxy-4-[\allowbreak{}[2-(4-hydroxy\allowbreak{}phenyl)-2-phenyl\allowbreak{}acetyl]amino]\allowbreak{}pyrrolidine-1-carboxylate} (iii),
which already contains the \iupac{4-hydroxy\allowbreak{}diphenyl\allowbreak{}methyl} and the aminated \iupac{hydroxy\allowbreak{}pyrrolidine} moieties of the PSEN1-01 structure.
N-Boc deprotection of molecule (iii) exposes the pyrrolidine nitrogen to yield \iupac{2-(4-hydroxyphenyl)-2-phenyl-N-[(3S,4S)-3-hydroxy\allowbreak{}pyrrolidin-4-yl]acetamide} (iv), which is subsequently benzoylated with benzoic acid (vi) to yield PSEN1-01 (vii). 
\synthesizer does not assume benzoic acid (vi) to be purchasable (due to the assumptions within the tool~\citep{luo2026efficient-prexsyn}),
so it proposes preparing benzoic acid via ester hydrolysis of \iupac{4-hydroxy\allowbreak{}phenyl benzoate} (v).
This route highlights a key feature of \main:
it provides an interpretable way for evaluating synthesizability in the form of a proposed synthesis route via \synthesizer. 
Furthermore, this demonstrates that \main can propose synthesis routes even for molecules with regioselectivity requirements and multiple stereocenters.
\subsection{Computational Efficiency and Cost 
% \xia{need a better title} \rzcomment{Revised (?)}
} 
\label{section:results:efficiency}
%===================================================
We assess the efficiency of \main along two dimensions:
% \amcomment{word choice. maybe "using two criteria" or just "in terms of" }: 
\textbf{(1)} the number of intermediate molecules generated to achieve the requested number of satisfying molecules (Section~\ref{sec:results:efficiency:interm}), and \textbf{(2)} the computational cost of a campaign, measured by running time and token consumption (Section~\ref{sec:results:cost}). 

% We discuss the efficiency of \main \rz{by observing the number of intermediate molecules generated to achieve the requested satisfying molecules} \xia{informal and procedural. revise please} \rzcomment{Revised}(Section~\ref{sec:results:efficiency:interm}), 
% \hl{along with how long \main takes to finish a campaign and the tokens consumed} \xia{not sure what this means... } \rzcomment{Revised} (Section~\ref{sec:results:cost}).

%

% \yecomment{From Avery: could analyze the details about the 5 difficult targets.}

% \noindent \yecomment{This subsection can be put in the appendix.}

% ------------------------------------------------------
\subsubsection{Intermediate Molecules Explored by \main}
\label{sec:results:efficiency:interm}
% ------------------------------------------------------
% Requires: \usepackage{booktabs}
\begin{table}[ht]
  \centering
  \caption{
  % \hl{Target distribution over intermediate molecules explored.}
  Number of targets grouped by the number of intermediate molecules explored.
  % \xia{what does this mean? } \rzcomment{Revised}.
  }
  % \yecomment{Could we move this sentence to the main text instead of the caption?} \rzcomment{Revised}
  \label{tbl:interm-dist}
  \footnotesize
  \begin{tabular}{
    @{\hspace{10pt}}l@{\hspace{10pt}}
    @{\hspace{6pt}}r@{\hspace{6pt}}
    @{\hspace{6pt}}r@{\hspace{10pt}}
  }
    \toprule
    \interm & Targets & Share (\%) \\
    \midrule
    $<750$               & 10 & 20.0 \\
    $750$--$1{,}000$     & 18 & 36.0 \\
    $1{,}000$--$1{,}250$ & 9  & 18.0 \\
    $1{,}250$--$1{,}500$ & 8  & 16.0 \\
    $\geq 1{,}500$       & 5  & 10.0 \\
    \bottomrule
  \end{tabular}
\end{table}
%

% \hl{In {\main}, an intermediate molecule is any molecule that {\hitidentifier}, {\leadoptimizer} or {\synthesizer} produces during a campaign and passes to {\evaluator}, whether or not it serves as the output molecule.}

Table~\ref{tbl:interm-dist} reports the number of intermediate molecules produced over the 50 targets.
The targets are grouped into several categories based on the intermediate molecules, and we observe how many targets exist in these categories.
% The number of intermediate molecules measures how many molecules \main explores for a target.  %or how much exploration a target demands to finally arrive to the requested number of molecules.
%
%\amcomment{is this section being moved to appendix? It seems like this should go before ablation study to properly introduce intermediate molecules/why we report them/using them as a proxy for efficiency}
%
% Table~\ref{tbl:interm-dist} reports the quantities over the 50 targets.
% In \main, an intermediate molecule is any molecule that \hitidentifier, \leadoptimizer or \synthesizer produces during a campaign and passes to \evaluator, whether or not it is eventually returned.
% %
% Since \main delivers exactly 100 satisfying molecules on every target, the number of intermediate molecules a campaign consumes measures directly how much search a target demands.
% %
% Table~\ref{tbl:interm-dist} reports the distribution of this quantity over the 50 targets.

\textbf{\textit{A majority ($>$50\%) of the targets require less than 1,000 intermediate molecules to achieve the requested satisfying molecules.}} 
%
% The majority of them (90\%) requires fewer than 1,500. 
% More than half of the targets require fewer than 1,000 intermediate molecules and 90\% with fewer than 1,500.
% More than half of the targets require fewer than 1,000 intermediate molecules and 90\% with fewer than 1,500.
%
Averaged over all targets, \main generates only 1,043.8 molecules to deliver 100 satisfying molecules, indicating that it explores the chemical space intelligently rather than through exhaustive enumeration.
For the hardest 10\% of targets, \main generates more than 1,500 molecules (up to 2,601 molecules), 
yet still delivers the requested satisfying molecules, emphasizing its robustness on even challenging targets. 
\subsubsection{Computational Cost}
\label{sec:results:cost}
%===================================================

%\yecomment{As we discussed before, we need to add the runtime and the output token analysis of baselines as well, to be compared with \main.}

% \xia{Is this section done or not? if yes, please remove comments} \rzcomment{Revised}

% \yecomment{We can use Table~\ref{tbl:efficiency} and we need add the time and token usage, some of which may be estimated.}

We evaluate the computational cost of \main by measuring
% \amcomment{measuring} 
\textbf{(1)} the running time and output token consumption, \textbf{(2)} the number of actions (e.g., \emph{process} and \emph{delegate}; Section~\ref{section:method:orchestrator}) taken by \orchestrator, and \textbf{(3)} the number of delegations per worker agent. 
% \st{the worker agent is delegated} \amcomment{the worker agents are delegated tasks}. 
%
We also include the computational cost for the shortest and longest campaigns, and present the results in Table~\ref{tbl:cost}.

% Table~\ref{tbl:cost} reports the wall-clock time, the number of output tokens, and the number of actions taken by \orchestrator over a campaign, averaged over the 50 targets.
% %
% We also include same information for both shortest and longest campaigns. 
% Table~\ref{tbl:cost} reports the wall-clock time, the number of output tokens, and the number of actions taken by \orchestrator over a campaign, averaged over the 50 targets and for the shortest and longest runs.

% \textbf{\textit{\main serves as a fast and low-cost method \xia {engine} for drug discovery.}}
\textbf{\textit{\main serves as a fast and low-cost autonomous system for drug discovery.}}
Across the 50 targets, \main finishes a campaign in 3.3 hours and consumes 21,125 output tokens on average. 
%and the 50 targets together require 165 hours and 678,409 output tokens.
%
\main takes at least about one hour on ADAMTS5~\citep{10.3389/fmolb.2021.703110-adamts5} and almost half a day on HDAC8~\citep{cells11193161-hdac8}, each represents the shortest and longest campaign of \main, respectively. 
Relative to the satisfying molecules delivered, the cost of generating one satisfying molecules is roughly 2 minutes and 211 output tokens, a cost negligible against conventional drug discovery~\citep{10.1001/jamanetworkopen.2024.53275-drug-discovery-billions}.
% \textbf{\textit{A campaign is inexpensive in absolute terms.}}
% %
% On average \main completes a target in 3.30 hours and 13,568 output tokens, and the 50 targets together require 165 hours and 678,409 output tokens.
% %
% Relative to the molecules delivered, this amounts to roughly two minutes and 136 output tokens per molecule, a cost that is negligible against the months a conventional hit-to-lead campaign requires for the same stage \ye{[Citation]}.

\textbf{\textit{Half of the delegations are made to \evaluator.}}
Out of 10.0 delegations issued per target on average, 5.10 go to \evaluator, while \hitidentifier, \synthesizer and \leadoptimizer receive 2.24, 1.66 and 1.02, respectively.
\evaluator is thus delegated as often as
% the same amount of time as 
the remaining three agents combined,
consistent with the transition probability discussion across worker agents (Section~\ref{sec:results:agent:prob}).
% \hl{consistent with the ablation study presented previously} \yecomment{Needs updates} (Section~\ref{sec:results:ablation}).
%
The number of delegations, however, does not determine the campaign duration: ADAMTS5 and HDAC8 both takes 1.2 and 11.6 hours, respectively, yet they differ by only two delegations (4 and 6).
This indicates that campaign cost is determined by the work performed within each delegation rather than the number of delegations issued.
\begin{table*}[t]
  \centering
  \caption{Computational cost of \main over 50 targets. 
  % \xia{need to update table format} \rzcomment{Revised}.
  }
  % \yecomment{We can directly use Table~\ref{tbl:efficiency}.}}
  \label{tbl:cost}
  \footnotesize
  \begin{threeparttable}
    \begin{tabular}{
      @{\hspace{6pt}}l@{\hspace{6pt}}
      @{\hspace{2pt}}r@{\hspace{2pt}}
      @{\hspace{2pt}}r@{\hspace{2pt}}
      @{\hspace{6pt}}r@{\hspace{6pt}}
      @{\hspace{2pt}}r@{\hspace{2pt}}
      @{\hspace{2pt}}r@{\hspace{2pt}}
      @{\hspace{2pt}}r@{\hspace{2pt}}
      @{\hspace{6pt}}r@{\hspace{6pt}}
      @{\hspace{2pt}}r@{\hspace{5.5pt}}
      @{\hspace{5.5pt}}r@{\hspace{5.5pt}}
      @{\hspace{5.5pt}}r@{\hspace{5.5pt}}
      @{\hspace{5.5pt}}r@{\hspace{6pt}}
    }
      \toprule
      \multirow{2}{*}{Metric} & \multicolumn{2}{c}{Cost} &  & \multicolumn{3}{c}{Action} &  & \multicolumn{4}{c}{Delegation per agent} \\
      % \multirow{2}{*}{Target} & \multicolumn{2}{c}{Cost} & \multicolumn{3}{c}{Action} & \multicolumn{4}{c}{Delegation per agent} \\
      \cmidrule(r){2-3} \cmidrule(r){5-7} \cmidrule(r){9-12}
      & Time (h) & Out. token &  &  Total & Process & Delegate &  & HI & LO & EV & SY \\
      \midrule
      Average             & 3.3  & 21,125 &  & 24.9 & 14.9 & 10.0 &  & 2.2 & 1.0 & 5.1 & 1.7 \\
      \midrule
      Shortest (ADAMTS5)  & 1.2  & 7,508  &  & 9    & 5    & 4    & &  1    & 0    & 2    & 1 \\
      Longest (HDAC8)     & 11.6 & 30,448 &  & 44   & 38   & 6    & &  1    & 1    & 3    & 1 \\
      % Mean value             & 3.30  & 21,125 & 24.9 & 14.9 & 10.0 & 2.24 & 1.02 & 5.10 & 1.66 \\
      % Shortest (ADAMTS5)  & 1.15  & 7,508  & 9    & 5    & 4    & 1    & 0    & 2    & 1 \\
      % Longest (HDAC8)     & 11.64 & 30,448 & 44   & 38   & 6    & 1    & 1    & 3    & 1 \\
      % \midrule
      % LIDDIA & 3.17 & 31,428 & - & - & - & - & - & - \\
      \bottomrule
    \end{tabular}
    % \begin{tablenotes}[flushleft]
    %   \footnotesize
    %   % \item[a] An action is a single decision taken by \orchestrator, either processing the artifacts returned by an agent or delegating a subtask to one.
    % \end{tablenotes}
  \end{threeparttable}
\end{table*}
% \input{table/computation}

% \newpage
% %===================================================
% \subsection{Expert Evaluation}
% \label{sec:results:expert}
% %===================================================
% Communicate with Daniel for more details. For example, let him review the output molecules, or evaluate the generation process of \main.

% \begin{itemize}
%     \item Known ligands and approved drugs
%     \item How many molecules we want to evaluate? 100 molecules? How to select candidates to evaluate? (Random selection / Top-N molecules)
%     \item 2D / 3D structures?
%     \item Baselines?
% \end{itemize}

\section{Discussions}\label{section:discussion}

We discuss the current limitations of \main and highlight several potential directions 
for future work.

\paragraph{In vitro validation for \main's generated molecules} 
\main is an autonomous multi-agent system for \textit{in silico} drug discovery---
with dedicated agents equipped with state-of-the-art \textit{in silico} tools to generate, optimize, evaluate, and plan the synthesis of molecules.
\main has demonstrated its ability to generate molecules that satisfy requirements grounded in these tools.
However, such computational tools remain imperfect proxies for laboratory synthesis and \textit{in vitro} experiments,
which are indispensable next steps to validate the real-world potential of molecules generated by \main.
Despite their importance,
both synthesis and \textit{in vitro} experimentation
remain challenging to perform systematically and at scale~\cite{chen2025generating-diffsmol}.
This motivates future research into how to better scale and incorporate \textit{in vitro} validation into \main.
In anticipation of such validation, \main provides synthesis routes, predicted binding affinity, and ADMET profiling for molecules it generates, providing extensive information needed to prioritize which candidates to advance.
Furthermore, the results from \textit{in vitro} experimentation can also be provided back to \main, enabling the system to adapt its workflow and revise its strategy to produce molecules more likely to advance in a real-world drug discovery campaign.

\paragraph{Human-in-the-loop design for \main}

Domain experts are essential for successful drug discovery---a domain where the reliable safety assessment is the highest priority and in-depth understanding is required. 
In this study, we couple {\main's} evaluation with assessments from a domain expert,  
% molecules produced by {\main} are examined by an expert on whether their structures are reasonable as hits or leads in a drug discovery campaign and their synthesizability.
%
serving primarily to verify the quality of the generated molecules, rather than to drive the generation process itself.
A natural extension for \main is integrating domain experts' feedback directly into the system, enabling an interactive framework in which expert judgments can steer the generation process toward more desirable regions of chemical space. 
The current design for \main is well-positioned to integrate this feedback: because \orchestrator already maintains an evolving
repository of candidates and decides which to pursue,
expert judgments can be supplied as additional information for \orchestrator to shape its decision.
For instance, domain experts can flag undesirable substructures and scaffolds on grounds that current \textit{in silico} tools do not capture, 
or indicate preferences among a set of candidates, 
thus steering the campaign towards more desirable molecules.
More importantly, such feedback need not be time-consuming: a domain expert can provide occasional feedback while still meaningfully shaping the drug discovery campaign. 
% \amcomment{and further, \main can keep record of this expert qualitative review and store/learn from it}
%
Through a human-in-the-loop design, 
\main can improve its reliability of producing molecules likely to advance in a real-world drug discovery campaign (e.g., \textit{in vitro} validation), while serving as a collaborator for domain experts (rather than a system that can only be inspected).
% Such a human-in-the-loop design improves both the reliability and applicability of \main.
%
% Concretely, these feedback may allow \main's generated molecules to be more likely to advance in a real-world drug discovery campaign (e.g., to \textit{in vitro} validation), and position \main as a collaborator rather than a system that can only be inspected.  

\paragraph{\main as a fully autonomous and self-driving system with laboratory integration}
A more ambitious direction is to integrate automated laboratory platforms directly into {\main}, enabling a fully autonomous, self-driving drug discovery system. 
Such a system allows \main to synthesize molecules 
% using the synthesis routes proposed by \synthesizer 
on automated synthesis platforms and submit the synthesized molecules for \textit{in vitro} validation,
closing the loop between \textit{in silico} design and \textit{in vitro} validation. 
\synthesizer and \evaluator are the natural integration points, by equipping both agents with automated synthesis and \textit{in vitro} validation platforms, respectively.
%
% The routes proposed by \synthesizer, along with the properties predicted by \evaluator, can inform which candidates to prioritize for synthesis on the platform (e.g., prioritizing molecules with higher binding affinity and least expensive starting materials).
%
\orchestrator can also serve as another point of integration.
Since synthesis is costly and time-consuming, 
this makes candidate selection a resource allocation problem (e.g., selecting only the most promising molecules to synthesize under a fixed budget), a decision that requires a component with a global-level view of a campaign, which \orchestrator provides.
Only molecules \orchestrator deems satisfactory should proceed to synthesis, with synthetic
feasibility from \synthesizer and predicted properties from \evaluator grounding
that decision. 
%
% The laboratory results can also be used as feedback to \orchestrator to ground its decision, allowing \orchestrator to revise its strategy.

Several safeguards must be implemented before a fully autonomous self-driving system can be deployed responsibly.
A system that acts in the physical world introduces safety concerns absent from a system that works purely \textit{in silico}.
For instance, a molecule deemed unsatisfactory \textit{in silico} can simply be discarded, whereas a synthesized molecule must be handled, stored, and disposed of responsibly.
Safeguards are also needed to ensure molecules generated by \main are not hazardous, whether due to error or misuse of the system.
Such safeguards may include structural screening against molecules known to be toxic and hazardous and including medicinal chemists in the critical step of laboratory testing.
Even so, the potential benefits are substantial.
Grounding \main's decisions in laboratory feedback can let the system steer its workflow toward real-world outcomes, and the resulting laboratory measurements can accumulate into data well suited to improving \main.
More broadly, such a system can shift the role of agents in drug discovery from informing medicinal chemists to driving the drug discovery process itself, with the potential to substantially reduce the time and cost of bringing new medicine to patients.

% Future work includes evaluating the molecules or including wet-lab inside \main.

% \textbf{Considering more requirements for properties}.

% \textbf{Using commercial LLMs, known to be "smart" for drug discovery, for \main}.

\section{Conclusions}\label{section:conclusion}

We present \main, an autonomous multi-agent system for \textit{in silico} drug discovery capable of generating molecules with satisfying properties for drug development. 
Rather than executing a fixed workflow, \main decides which worker agent to invoke through \orchestrator, drawing on its memory module and evolving repositories of candidates for its decision-making.
Across 50 targets, \main outperforms state-of-the-art baselines spanning generative models for drug design, agents for drug discovery, and generalist LLMs, successfully generating 100 satisfying molecules per target that meet key molecular property requirements while also providing their synthesis routes.
Our ablation reveals that this performance is attributed to \main's collaboration between agents, 
where each worker agent contributes distinctly, 
while \orchestrator contributes the most by concentrating effort on the most promising candidates 
that greatly reduce the chemical space \main must explore to find the satisfying molecules.
Consistent with its design, \main exhibits strong adaptability across campaigns, effectively employing different strategies as each target demands.
Notably, \main demonstrates intelligent strategy adaptations for a challenging target with no existing marketed FDA-approved drugs.
On PSEN1, a critical target associated with Alzheimer's disease, \main generates PSEN1-01, a promising molecule with \textit{in silico} properties outperforming those of the reference ligand and Nirogacestat, while also demonstrating favorable binding interaction with the challenging site of the target.
\main achieves strong performance at low cost and time, with one campaign roughly producing 21K output tokens and taking 3.3 hours on average.
We envision \main as a foundation for future autonomous drug discovery systems that can carry a molecule \textit{in silico} to potential drug candidates, ultimately accelerating the timeline and cost of bringing new drugs to patients.

%%=============================================%%
%% METHOD
%%=============================================%%

% \newpage
\section{Methods}\label{section:method}

We introduce \main, a multi-agent system for \textit{in silico} drug discovery.
It is composed of a central agent, \orchestrator, and four specialized worker agents.
Given a target protein and a set of desired molecular properties, 
\main aims to identify molecules that bind to the target and satisfy all desired properties. 
\main performs the task by distributing subtasks across worker agents, with \orchestrator deciding which subtask to pursue next.
This design enables \main to decompose a complex, multi-objective problem typical in drug discovery into manageable subtasks, enabling each agent to collectively advance the drug discovery campaign. 

% We introduce \main, an agentic framework for \textit{in silico} drug discovery.
% %
% By coupling the reasoning capabilities of a large language model with a suite of specialized computational tools 
% for drug discovery, \main operates as an intelligent, autonomous system capable of navigating the drug discovery pipeline end-to-end.
% % Given a target protein and a set of desired molecular properties, \main couples the reasoning capabilities of a large language model with a suite of specialized computational tools to find molecules that satisfy all of them.
% % 
% Figure~\ref{fig:method-main} presents its architecture.
% %
% Given a target protein and a set of desired molecular properties, 
% \main aims to identify molecules that bind to the target and satisfy all desired properties. 
% %
% To do so, \main is designed as a multi-agent system composed of a central agent, \orchestrator, and four specialized worker agents.
%
\orchestrator (Section~\ref{section:method:orchestrator}) is responsible for the entire workflow of \main, deciding how a drug discovery campaign proceeds by delegating to a worker agent and processing the molecules. 
Under its coordination, the worker agents perform distinct subtasks critical to a successful drug discovery campaign:
\textbf{(1)} \hitidentifier (Section~\ref{section:method:hitidentifier}) serves as the entry point of a campaign, providing candidate molecules from which a campaign proceeds;
\textbf{(2)} \leadoptimizer (Section~\ref{section:method:leadoptimizer}) serves as the refinement stage, advancing existing candidate molecules towards those with satisfying properties;
\textbf{(3)} \synthesizer (Section~\ref{sec:method:synthesizer}) bridges the gap between molecules designed \textit{in silico} and those that can be made in practice, proposing synthesis routes and finding synthesizable analogs;
and \textbf{(4)} \evaluator (Section~\ref{section:method:evaluator}) functions as a quality control layer, assessing the key properties that define a successful drug, including binding affinity, drug-likeness, and safety-related properties.  %\citep{swanson2024admet-admetai}; 
Acting in concert, these components allow \main to emulate the iterative, collaborative nature of a real-world campaign within a single autonomous system.

% \main maintains a persistent \workspace that records all artifacts supplied by the users and generated by the individual agents throughout a campaign. 

%
Each agent (\orchestrator and the four worker agents) maintains its own \emph{workspace}, each is used to store relevant artifacts required for their tasks (Table~\ref{tbl:worker_components}).
% \yecomment{need specify that \orchestrator and four worker agents all have their own \emph{workspace}, and Table~\ref{tbl:worker_components} is not used to introduce the tools used by four worker agents rather than workspace.} \rzcomment{Revised}
%
The workspaces are dynamically changing and growing continuously, with more artifacts (e.g., molecules) generated as the campaign progresses. 
Each worker agent operates within its own \workspace, with no access to the others. 
This prevents any alteration of artifacts from worker agents outside their own \workspace, preventing error propagation throughout \main. 
Meanwhile, \orchestrator alone has access to each worker agent's \workspace, granting it a global view required to lead the campaign. 

\subsection{\orchestrator}
\label{section:method:orchestrator}

At the center of \main is \orchestrator, an agent responsible for the decision-making of the entire drug discovery campaign. 
% \rz{\orchestrator serves as the central component in \main, guiding the drug discovery campaign and ensuring that the worker agents act in concert.}
%
\orchestrator decides how a campaign proceeds: which worker agent to call next and which candidate molecules to focus on. 
Each decision is conditioned on two sources of information: 
\textbf{(1)} the existing candidate molecules generated throughout the campaign, 
% (pooled as \textit{property satisfying candidate set} and \textit{synthesizable and property satisfying candidate set} described later),  
% \yecomment{In the following, we use the ``candidate pool", we could unify these. Also, the reader may be confused about what is the candidate set/pool here as it is the first time we mention this concept. Do you think we need a preliminary paragraph to introduce some important concepts, such as ``candidate pool", ``\workspace"?}
which captures the progress of the campaign, and 
\textbf{(2)} the profiles 
% \yecomment{Add a reference to its content.} 
% \rzcomment{Revised} 
of the worker agents (Appendix~\ref{supp:worker_profile}), which describe the capabilities that each worker agent can contribute to the campaign. 
Together, these information allow \orchestrator to reason jointly about what the campaign needs and which worker agent 
% \hl{who} 
% \yecomment{``which worker agent" is more appropriate?}
% \fb{[agreed, let's not personify the agents in formal writing.]} 
is best suited to provide it. 
Furthermore, this information enables dynamic decision-making for \orchestrator, choosing which drug discovery step to focus on next and calling the worker agent dedicated to that step.
This design makes \main's campaign non-linear and adaptive, mirroring the iterative, feedback-driven nature of real-world drug discovery~\citep{hughes2011principles}. 
% \yecomment{Citation}. 

Operationally, \orchestrator makes two kinds of decisions. 
It can choose to advance the campaign by issuing a \delegate action to a worker agent, or 
to collect information (e.g., the properties of existing candidate molecules) by issuing a \process action. 
\textbf{(1)} Using a \emph{delegate} action, \orchestrator sends a worker agent a prompt consisting of a task description and references to the artifacts stored in each worker agent's \workspace (Section \ref{section:method:worker}). 
In return, it receives a structured report of what the worker agent has done. 
These reports, along with the information gathered from \process action(s) and the profiles of worker agents will determine which worker agent should be called next via \delegate. 
\textbf{(2)} Using a \emph{process} action, \orchestrator reads the artifacts produced from the worker agent it has just called.
What it takes from them depends on the worker agent: the molecules generated by \hitidentifier or modified by \leadoptimizer, the property values computed by \evaluator, or the synthesis routes found by \synthesizer. 
From these artifacts it updates its view of which candidate molecules meet the requirements and which of them still fall short. 
%
% \orchestrator alone has access to all the artifacts generated throughout a campaign, granting it a global view required to lead the campaign. 

With these two actions, 
% By alternating \fb{interleaving?} these two actions, 
%
\orchestrator drives the campaign forward (through \textit{delegate)} while continually grounding its decision (through \textit{process}). 
Carrying these actions out over many steps rests on two elements of its design: \textbf{(1)} the memory that keeps its own record tractable and \textbf{(2)} the satisfying candidate repositories in which it keeps the molecules worth pursuing, stored in the workspace.
% \yecomment{Budget and termination is on the same level, but it is not introduced here.}
% \yecomment{We need specify these are in \orchestrator's workspace, otherwise, readers will get confused whether \orchestrator has a workspace, and if so, what is its workspace's usage...} \rzcomment{Both revised}
% and the budget to finish a campaign. 
% \hl{within which it must complete it}\fb{[this is grammatically correct, but not very easy to read (``it'' is used twice for 2 different things)]}.

% \yecomment{Use paragraph}
\paragraph{Memory module}
A defining feature of \orchestrator is its capability to sustain a long-horizon campaign for \main (Section \ref{sec:results:agent:case}). 
This is enabled by a dedicated memory module that dynamically stores the most relevant information. 
%
% Over many steps, the record on which \orchestrator conditions its decisions grows beyond what the underlying LLM can attend to.
%
% \rz{The memory module therefore serves to store and surface the most relevant information to the \orchestrator.}
%
The module holds the records of the campaign itself---the information of the artifacts produced, the reports the worker agents return and \orchestrator's own prior decisions---kept in full. 
Other information is stored and pruned to the most recent entries, such as system-level messages generated throughout the campaign (e.g., \orchestrator calling a worker agent with incorrect format). 
%
% This includes errors generated from the underlying LLM (e.g., calling a worker agent incorrectly) and other system-level messages (e.g., addition of artifacts to \workspace). 
%
When the memory module exceeds what the underlying LLM can hold in its context, the module compresses its records by summarizing them, selectively storing and pruning information based on its recency and relevance.  
The module thus provides \orchestrator a compact, decision-relevant context, allowing it to condition each decision on the salient history of the campaign while remaining within the effective reasoning capacity of the underlying LLM. 
%
% Further details is available in \rz{(refer to Supplement)}.

% \yecomment{Use paragraph}
\paragraph{Satisfying candidate repositories} 
%
% \rz{Property pool comes from HI and LO AND pass EV. They will go to SY and if pass, go to Syn pool. In Syn pool, there are two mols then: (1) from HI and LO that pass SY and (2) analogs from SY that pass EV. The Syn pool is like the main bar, and property pool is like intermediate molecules that only contain "good ones" without synthesizabily, while Syn pool contain both good ones and synthesizable}
%
Over the course of a campaign, \orchestrator maintains two repositories of molecules. 
The first repository $\mathcal{P}$ holds the candidate molecules that satisfy every property evaluated by \evaluator. 
Concretely, these are molecules from \hitidentifier and \leadoptimizer that pass each property requirement. 
The second repository $\mathcal{S}$ holds candidate molecules that both satisfy every property and are also synthesizable as verified by \synthesizer. 
This set has two sources: 
\textbf{(1)} molecules from $\mathcal{P}$ that are synthesizable, and 
\textbf{(2)} their synthesizable analogs (Section \ref{sec:method:synthesizer}) found by \synthesizer that satisfy every property as evaluated by \evaluator. 
\orchestrator takes both repositories into account when choosing each subsequent action.
% The two sets form the state of a campaign, from which \orchestrator chooses each subsequent action. 
%
These repositories serve several purposes: $\mathcal{P}$ tracks which candidate molecules are worth considering for synthesis assessments, while $\mathcal{S}$ tracks which of them are both satisfying and also synthesizable. 
They also inform \orchestrator on which worker agent it should call next. 
For instance, a large $\mathcal{P}$ means there are many candidate molecules with satisfying properties, which informs \orchestrator that it is better to call \synthesizer to verify their synthesizability next.  

% \yecomment{Should we add one paragraph to introduce the \textbf{worker agent profile} here?}

% \yecomment{Use paragraph}
% \textbf{Budget and termination}.
%
\orchestrator is allowed a budget of actions per campaign, and stops once $\mathcal{S}$ holds the requested number of molecules or this budget is exhausted.
At each step \orchestrator is informed of how many actions remain, and once a few of them are left it stops searching for new molecules and delegates those in $\mathcal{P}$ to \synthesizer, ensuring that the candidates accumulated so far can still enter $\mathcal{S}$ before the budget runs out.

% \textbf{Budget and termination}.
% %
% \orchestrator is allowed a budget of $T_{\max}$ actions per campaign, and stops once $\mathcal{S}$ holds the requested number of molecules or this budget is exhausted.
% %
% At each step \orchestrator is told how many actions remain, and once fewer than $b$ of them are left it stops searching for new molecules and delegates those in $\mathcal{P}$ to \synthesizer, ensuring that the candidates accumulated so far can still enter $\mathcal{S}$ before the budget runs out.

\bigskip\noindent The combination of these capabilities in \orchestrator is what distinguishes \main from prior approaches.
Together they allow \main to operate autonomously, 
to accelerate drug discovery process (Section~\ref{sec:results:cost}), and 
to achieve strong performance in our benchmark (Table~\ref{tbl:overall-wout}). 
Critically, they allow \main to complete long-horizon campaigns on varying targets (Section~\ref{sec:results:agent:case}). 
\subsection{Worker Agents}
\label{section:method:worker}
%%=============================================%%
% \yecomment{Should we explicitly explain the worker agent profile in each subsubsection? Currently, it seems only appears as a sentence \textit{``Each worker agent maintains a static profile, containing a brief summary of its capabilities and what subtask they support"}}

% \yecomment{Also, since we have removed the \workspace introduction in the current version, as we discussed, we need to add the \workspace definition/explanation to each worker agent.}

Each worker agent carries out one specialized subtask in a drug discovery campaign, 
each critical for a successful campaign. 
Each worker agent maintains a static profile, containing a brief summary of its capabilities and what subtask they support (Appendix~\ref{supp:worker_profile}). 
These profiles are used by \orchestrator during its decision-making. 

% When carrying out a subtask, each worker agent operates within its own \workspace, with no access to the others. 
%
% This prevents any alteration of artifacts from worker agents outside their own \workspace, preventing error propagation throughout \main. 
%
% Meanwhile, \orchestrator alone has access to each worker agent's \workspace, granting it a global view required to lead the campaign. 

Each worker agent receives from \orchestrator a 
% \yecomment{prompt xxx, make it the same as the expression as the \orchestrator section.} 
% \rzcomment{Revised} 
prompt containing a task description with references to the artifacts in its \workspace, executes the subtask with its own tools and stores the artifacts it produces in its \workspace.
The worker agent then issues a \emph{report} by providing a structured summary of what it has done back to \orchestrator. 
Table~\ref{tbl:worker_components} lists the artifacts each worker agent requires and produces in its \workspace and the tools used. 
% Among these artifacts are the two pools: \evaluator's property values are what admit a molecule into $\mathcal{P}_t$, and the synthesis routes \synthesizer finds for the molecules in $\mathcal{P}_t$ are what admit them into $\mathcal{S}_t$.
The subsections below describe the strategy with which each worker agent approaches its subtask.
% Requires: \usepackage{booktabs,threeparttable}
\begin{table*}[t]
  \centering
  \caption{Artifacts required and produced by each worker agent, along with their respective tools.}
% Artifacts are not passed between agents directly: \orchestrator's prompt names the paths at which a worker agent's inputs are stored in the \workspace, and the outputs it produces are written back to its own folder there.}
  \label{tbl:worker_components}
  \footnotesize
  \begin{threeparttable}
    \begin{tabular}{
      @{\hspace{2pt}}p{0.125\textwidth}@{\hspace{10pt}}
      @{\hspace{2pt}}p{0.2\textwidth}@{\hspace{10pt}}
      @{\hspace{2pt}}p{0.175\textwidth}@{\hspace{10pt}}
      @{\hspace{2pt}}p{0.4\textwidth}@{\hspace{2pt}}
    }
      \toprule
      Worker agent & Input & Output & Tools \\
      \midrule
      \hitidentifier &
      Target structure. &
      Candidate molecules. &
      conDitar~\citep{gao2026generatingdevelopable3dmolecules-conditar} for SBDD; DiffSMol~\citep{chen2025generating-diffsmol} for LBDD; AutoDock~\citep{autodock} for predicting binding affinity.
      % \yecomment{not the unidocking, need to verifiy this with Ruoxi.} 
      % QED~\citep{bickerton2012quantifying-qed} and SAScore~\citep{ertl2009estimation-sascore} for filtering.
      \\
      \midrule
      \evaluator &
      Candidate molecules. &
      Property values. &
      UniDock~\citep{10.1021/acs.jctc.2c01145-unidock}, QED~\citep{bickerton2012quantifying-qed}, SAScore~\citep{ertl2009estimation-sascore}, ADMET-AI~\citep{swanson2024admet-admetai} and Lilly filter~\citep{10.1021/jm301008n-lilly}. \\
      \midrule
      \leadoptimizer &
      Candidate molecules that fail one or more requirements. &
      Optimized candidate molecules. &
      GeLLMO-C~\citep{Dey2025}; MolOpt algorithms~\citep{NEURIPS2022_8644353f-pmo}; RDKit and \evaluator-backed oracles for the target properties. \\
      \midrule
      \synthesizer &
      Candidate molecules. &
      Synthesis routes or synthesizable analogs. &
      MEEA$^{*}$~\citep{zhao2024efficient-meea} and PrexSyn~\citep{luo2026efficient-prexsyn}. \\
      \bottomrule
    \end{tabular}
  \end{threeparttable}
\end{table*}

\subsubsection{\hitidentifier}
\label{section:method:hitidentifier}
%%---------------------------------------------%%

% \rz{Highlights of HitIdentifier: }
% \rz{
% \begin{itemize}
%     \item HitIdentifier supplies the candidate molecules from which the rest of campaign of LIDDIA2 will proceed
%     \item HitIdentifier serves as the entry point and the component that ensure that the molecules can bind well to the target. Emphasize on binding because it is the first important property (citation) before it can be considered worth pursuing to other properties, just like in practice. 
%     \item HitIdentifier search for these promising binding molecules by two methods: SBDD+details and LBDD+details. These are also analogous to real world drug discovery. 
%     % \item Explain details after: what kind of task orchestrator can delegate, what decision making happen in HitIdentifier, etc
%     % \item Add what artifacts generated in workspace. 
% \end{itemize}
% }

\hitidentifier serves as the entry point of \main's campaign, supplying the candidate molecules from which the rest of the campaign proceeds.
It is the only worker agent that conditions the molecule generation directly on the target structure, enabling \main to propose molecules likely to bind the target.  
Similar to hit identification step in drug discovery, \hitidentifier prioritizes binding affinity, a key property for molecules to have the intended therapeutic effects~\citep{sawada2024predicting}. 
Every molecule \main proposes is derived (e.g., optimized) from what \hitidentifier finds, emphasizing its importance. 

%
% Everything \main later assesses, optimizes or replaces comes from what \hitidentifier finds, emphasizing its importance. 
%
% \hitidentifier aims to find strong binding molecules, or those likely to interact strongly with the binding pocket of the target structure. 

\hitidentifier first receives a subtask from \orchestrator containing the references to artifacts in its \workspace (e.g., target structure), a number of molecules to return and a budget of iterations to find them.
% some property requirements. 
% commonly considered during hit identification in drug discovery~\citep{XXX}. 
%
% The requirements include QED~\citep{bickerton2012quantifying-qed} and SAScore~\citep{ertl2009estimation-sascore}.   
% \rzcomment{and some (optional) property constraints: Vina, QED, SAScore)}
% 
% \hitidentifier returns $n$ hits: molecules predicted to bind $P$ and passing the constraints set by the \orchestrator. \hl{passing preliminary filters on drug-likeness and synthetic accessibility.}
% \yecomment{We need to verify this with Ruoxi.}
% \rzcomment{the more correct one is "passing constraints set by the \orchestrator's prompt (if any)"}
% 
\hitidentifier then finds these molecules over a series of iterations using two strategies: structure-based drug design (SBDD) \citep{ijms20112783-sbdd} and ligand-based drug design (LBDD) \citep{10-1039-c1md00044f-lbdd}. 
At each iteration, it selects a strategy, invokes the corresponding tool, and accumulates the resulting molecules alongside those from previous iterations. 
The molecules are then scored using their predicted binding affinities, stored in its \workspace, and accumulated to guide the selection of the next strategy.
\hitidentifier stops once the molecules satisfying the constraints reaches or exceeds the requested number from \orchestrator or it exhausts the budget of iterations. 
% In each iteration, it reasons over the molecules it has found so far and chooses between two strategies: structure-based drug design (SBDD) \citep{ijms20112783-sbdd} and ligand-based drug design (LBDD) \citep{10-1039-c1md00044f-lbdd}. 
% 
% The corresponding tool for each strategy is run and the generated molecules are scored with its internal evaluation tools. 
%

%
% the molecules it generates with UniDock}~\citep{10.1021/acs.jctc.2c01145-unidock}, \hl{QED}~\citep{bickerton2012quantifying-qed} \hl{and SAScore}~\citep{ertl2009estimation-sascore}, \hl{keeping the best of them.}
% \yecomment{Needs further check this with Ruoxi.}
% \rzcomment{keeping the ones that satisfy the constraints listed in the \orchestrator's prompt (if any)}
% 
% \hitidentifier stops once it has kept $n$ molecules or exhausted $I_{\max}$, and returns the strongest-binding among them.
% \yecomment{A little confusing here. In practice, if \hitidentifier have generated more than $n$ molecules, then it will return all of them; Otherwise, if it cannot generate enough molecules within $I_{\max}$ iterations, it will be seen as a failed run of \hitidentifier and return the error to \orchestrator. We could make sure this with Ruoxi.}

% \textbf{Structure-based drug design}. SBDD enables \hitidentifier to design candidate molecules that bind into the binding pocket by exploiting its spatial configurations, offering a mechanistic approach to hit discovery grounded in pocket-ligand interactions. SBDD further provides a principled starting point for ligand-based modeling, complementing LBDD in finding hits. 

\paragraph{Structure-based drug design}
SBDD generates molecules conditioned on the three-dimensional structure of the target pocket.
It enables \hitidentifier to design molecules by exploiting the binding pocket's spatial configurations, offering a mechanistic approach to molecular discovery grounded in pocket-ligand interactions. 
SBDD further provides a principled starting point for ligand-based modeling, or LBDD, and thus complements each other. 
%
% \hitidentifier uses a generative model trained on large-scale experimental data, enabling \textit{de novo} generation that increases the likelihood of finding novel and more effective molecules. 

% It is available to \hitidentifier from the first iteration onwards, and is the strategy by which \hitidentifier reaches binding modes it has not yet tried.
% \hitidentifier uses conDitar~\citep{gao2026generatingdevelopable3dmolecules-conditar} for this, a \textit{de novo} generative model trained on large-scale experimental data, which lets it propose molecules outside any existing library.

\paragraph{Ligand-based drug design}
LBDD instead generates molecules conditioned on the shape of one that already binds to the pocket.  
% using DiffSMol~\citep{chen2025generating-diffsmol}.
% Such a molecule is either supplied by \orchestrator or generated by \hitidentifier, so LBDD becomes available to \hitidentifier only once a binder is at hand, and \hitidentifier chooses which one to condition on along with the strategy itself.
Where SBDD generates molecules from the binding pocket, LBDD expands around the known binding molecules. 
Since it relies on different prior information (e.g., shape-conditioned instead of pocket-conditioned), LBDD broadens \hitidentifier's exploration when combined with SBDD and thus, further increases the likelihood to find more and stronger binding molecules. 
\subsubsection{\leadoptimizer}
\label{section:method:leadoptimizer}

\leadoptimizer is the molecular-refinement component of \main. 
It complements the molecular generative stages as done by \hitidentifier (e.g., generates candidates from a binding pocket) and \synthesizer (e.g., generates synthesizable analogs from existing candidates). 
Specifically, their candidate molecules are used as starting points by \leadoptimizer to produce derivatives with improved properties. 
\leadoptimizer 
%adapts to 
addresses the
different property requirements a campaign requires through two different families of optimizer: 
\textbf{(1)} pretrained controllable multi-property optimizer~\citep{Dey2025} and 
\textbf{(2)} oracle-guided single-property optimizer~\citep{NEURIPS2022_8644353f-pmo}, each is distinct and complements each other. 
They provide both 
%the 
adaptability and control 
over
% whatever \main's campaign demands,
the property improvements to whatever a campaign demands, 
% \am{property optimization,} 
enabling \leadoptimizer to
transform molecules in \main to those with satisfying properties. 
% \am{meet campaign-specific objectives.}
%
% \leadoptimizer therefore transforms the candidate molecules in \main to those with satisfying properties, closing the gap between initial candidates to promising ones.

\leadoptimizer performs
%the
a
subtask by first formulating a plan and then selecting the most suitable optimizer.
Concretely, after \orchestrator identifies which candidate molecules are needed to improve, what properties require %improvements, 
improvement, and the number of derivatives to produce for each molecule,
\leadoptimizer 
%first \amcomment{redundent}
turns the request into a structured plan with objectives.
It contains both properties to be improved and the required values or ranges. 
The plan then guides \leadoptimizer to select the best optimizer from one of two families, based on which properties to optimize, how many of them, and the performance of that optimizer, when such records are stored by \leadoptimizer. 
%
% \amcomment{run on sentence}
%
Once an optimizer is selected, \leadoptimizer applies it to each candidate molecule, producing molecule derivatives that are then validated, de-duplicated, stored in its workspace, and reported back to \orchestrator. 
If the initial attempt doesn’t achieve the requested property improvement(s) or produce enough valid derivatives, \leadoptimizer may run follow-up attempts including retrying or switching optimizers, subject to an execution budget.

\paragraph{Pretrained controllable multi-property optimizer} The pretrained controllable multi-property optimizer family is well suited when the subtask requires improving multiple properties while attempting to preserve 
%the 
other desirable molecular properties. 
This family offers \leadoptimizer greater control over 
%the property improvements, such as
which properties to maintain and improve. 
%and how much improvement is needed. 
% \am{and the extent of change required.}
%
In contrast to using multiple single-property optimizers (the second family), a pretrained multi-property optimizer 
%benefits more by providing a 
provides
more balanced improvement (e.g., improving while maintaining) and efficiency (e.g., one instead of multiple runs). 
% \amcomment{I think this misses the point a little bit, it is less about the single/multi and more about the basis in LLM/pretrained}
%
However, 
%since they are 
because this family is often 
pretrained on specific properties~\citep{Dey2025}, 
it limits their adaptability for the wide range of properties a drug discovery campaign may require,
% \am{their applicability is limited when a campaign requires properties outside their learned objectives,}
%
necessitating \leadoptimizer to use another family of optimizers.  

\paragraph{Oracle-guided single-property optimizer} 
Oracle-guided single-property optimizers refine molecules using an \textit{in silico} property predictor available to \leadoptimizer (Table \ref{tbl:worker_components}). 
This family is primarily used 
%as a fallback option, when properties to optimize are not supported by the first family. 
when the properties to optimize are not supported by the first family.
% \am{when the request centers on a single property that can be directly scored, or when the requested property is not supported by the first family.}
%
% Each candidate molecule is scored by the predictor, modified by an optimizer, and rescored again to observe the improvement. 
%
\leadoptimizer has access to more than one optimizer in this family. 
To select the best one, it uses its
available
property predictors, 
%previous results of optimization,
prior optimization results,
and 
%the
historical performance records of each tool as guidance (Appendix~\ref{supp:leadoptimizer_record}). 

\subsubsection{\synthesizer}
\label{sec:method:synthesizer}
%==========================

% In \main,
\synthesizer investigates the synthesis of molecules,
assessing whether the candidate molecules in \main can be made in practice.
This is critical for \emph{in silico} drug discovery,
as synthesizability is a major bottleneck to development of AI-generated molecules~\citep{blass_basic_2021}.
\synthesizer assesses synthesizability based on whether it can generate a synthesis route for a given molecule,
grounding its assessments in interpretable reaction chemistry
rather than relying solely on estimated synthetic accessibility scores~\citep{ertl2009estimation-sascore}. 
\synthesizer uses two complementary strategies to generate synthesis routes: \textbf{(1)} retrosynthesis planning~\citep{zhao2024efficient-meea}, and \textbf{(2)} synthesizable analog generation~\citep{luo2026efficient-prexsyn}.
%
% Concretely, \synthesizer aims to find synthesis routes for candidate molecules using retrosynthesis planning, and if the routes cannot be found, identify the candidate molecules' synthesizable analogs --- structurally similar molecules with known synthesis routes --- using synthesizable analog generation. 
%
% When \synthesizer identifies a synthesis route for a given molecule using retrosynthesis planning,
% it concludes the molecule is synthesizable.
%
% When this first strategy fails to find a route,
% \synthesizer instead generates synthesizable analogs---molecules with high structural similarity to the given molecule and identifiable synthesis routes.
%
This two-strategy approach mirrors conventional practice in drug discovery,
where molecules that are difficult to synthesize directly are used to identify analogs from established libraries of synthesizable molecules.

\paragraph{Retrosynthesis planning}
\synthesizer uses retrosynthesis planning to generate
% generates
synthesis routes by starting from a given candidate molecule and working through the reverse logic of synthesis reactions to reach purchasable starting materials.
Using this approach, \synthesizer can identify synthesis routes to a specific molecule,
mimicking the retrosynthetic analyses performed by chemists in conventional drug discovery. 
% mimicking the retrosynthetic analyses performed by synthetic chemists in conventional drug discovery.
% This planning process relies on heuristic guidance and reaction prediction to iteratively navigate the vast space of chemical reactions.
% 
% A molecule for which a complete route is found is synthesizable, and \synthesizer goes no further with it; one for which the search fails is passed to the second strategy.
%
% [We need some explanation as to why \synthesizer stops here, rather than just stating the logic above.  Perhaps something like:]
If this planning process can identify a route for a given molecule, 
\synthesizer uses the route as the evidence to conclude that the molecule is synthesizable and terminates. 
% the \synthesizer has sufficient evidence to conclude that the molecule is synthesizable and terminate. 
%
However, if no route is found, \synthesizer investigates further using the synthesizable analog generation strategy. 

\paragraph{Synthesizable analog generation}
In synthesizable analog generation, \synthesizer identifies analogs by constructing synthesis routes to molecules that are structurally similar to the given candidate molecule.
Unlike retrosynthesis planning, the route generation starts from purchasable starting materials and works through the forward logic of synthesis reactions to produce analogs.
This forward-synthesis approach naturally complements retrosynthesis planning,
and can uncover synthesis routes that retrosynthesis planning may struggle to identify.
In some cases, the product of the generated synthesis route is not an analog,
but the given candidate molecule itself.
When this occurs, \synthesizer concludes that the candidate molecule is synthesizable.
Otherwise, \synthesizer concludes that the candidate molecule is unsynthesizable and provides the analogs and their routes to \orchestrator as potential alternatives to the candidate molecule.
Thus, \synthesizer goes beyond synthesizability assessment,
converting unsynthesizable molecules into potentially promising new leads to further the drug discovery campaign.

\subsubsection{\evaluator}
\label{section:method:evaluator}
%%---------------------------------------------%%

% \rz{Highlights of Evaluator: }
% \rz{
% \begin{itemize}
%     \item Evaluator serves as the quality mechanism in LIDDIA2. 
%     \item Whether the molecules are good or not is dependent on Evaluator, from which Orchestrator can make decision whether they are worth pursuing or not. The decision of Orchestrator is ultimately dependent from Evaluator.
%     \item Evaluator can evaluate many properties, each of them is critical and is also property that is often considered in real world drug discovery: Docking, whether the molecule can bind well, QED to measure drug-likeness (how close they are to a drug), SAScore for quick estimate of synthesizability, ADMET to measure how the drug will behave in the body, LILLY filter for structural liabilities (whether they are promiscuous, etc), etc. 
%     % \item add workspace
% \end{itemize}
% }

\evaluator serves as the quality control mechanism in \main. 
Every decision made by \orchestrator regarding the candidate molecules is grounded in \evaluator's assessments, thus shaping the trajectory of \main's campaign. 
% \yecomment{May be too strong here. Only the decisions concerning the properties are grounded with \evaluator.}
%
\evaluator performs a comprehensive assessment of each candidate molecule and supplies the property values on which the rest of a campaign depends. 
% 
% To do so, it performs a comprehensive assessment of each candidate molecule on properties essential to a successful drug. 
%
% \rz{also mention it different than HitIdentifier evaluation tool.}
%
The properties \evaluator supports include binding affinity, drug-likeness, synthesizability, ADMET and structural liabilities. 
Each is critical to a successful drug and computed with its own respective tool (Table \ref{tbl:worker_components}). 
Its synthesizability assessment differs from \synthesizer since \evaluator uses a rapid heuristic scorer~\citep{ertl2009estimation-sascore} rather than finding an entire synthesis route from purchasable starting molecules to the candidate molecule. 
The score complements the synthesis route---it can convey information on the ease of synthesis~\citep{ertl2009estimation-sascore} as a quantitative number that can help in prioritizing molecules.

\section*{Acknowledgements}

This project was made possible, in part, by support from 
the National Science Foundation grant no. IIS-2435819 (X.N., D.A.-A.), 
the National Library of Medicine grant no. R01LM014385 (X.N.), 
and the National Institute of Aging grant no. R01AG082044 (X.N.). 
Any opinions, findings, conclusions or recommendations expressed in this paper are those of the authors and do not necessarily reflect the views of the funding agency.

\section*{Conflict of interests}
The authors claim no conflict of interests.

\section*{Ethics Statement}

\main is an autonomous system for drug discovery, enabling users to generate molecules that meet the specified property requirements prompted in natural language.
We acknowledge that not all generated molecules are safe, and \main can potentially be used to generate harmful molecules. 
We take a step to minimize the potential negative impact and design \main to operate purely \textit{in silico}, with no capability for the system to execute synthesis and produce physical molecules.
Nonetheless, we highly encourage supervision by responsible domain experts when using \main.
All users of \main are advised to exercise discretion and to comply with applicable safety guidelines, regulations, and established professional best practices.

\section*{Code Availability}
All the source code and implementation for \main are available in  \mbox{\url{https://github.com/ninglab/MADDIE}}.

\section*{Data Availability}
The preprocessed 3D structures of the protein targets, their annotations, and the associated FDA-approved drugs are available in \mbox{\url{https://github.com/ninglab/MADDIE/tree/main/dataset}}.

\section*{Author Contributions}
X.N. conceived the research and conducted the project administration.
X.N. investigated the research, obtained funding and resources for the research, and supervised the student and postdoctoral authors.
Y.L., R.A. and X.N. designed the computational methodologies of \main.
Y.L. and R.A. designed and implemented the computational methodologies of \orchestrator.
R.G. designed and implemented the computational methodologies of \hitidentifier.
A.M. designed and implemented the computational methodologies of \leadoptimizer.
F.B. designed and implemented the computational methodologies of \synthesizer.
Y.L. designed and implemented the computational methodologies of \evaluator.
Y.L. conducted the main experiments.
R.A., A.M., F.B. and R.G. conducted the baseline experiments.
R.A. and A.M. conducted data curation.
Y.L. and R.A. conducted formal analysis and result analysis.
D.A.-A. contributed medicinal chemistry expertise to the result analysis.
Y.L. conducted visualization; R.A. and A.M. contributed to visualization.
R.A. drafted the original manuscript.
Y.L., A.M., F.B., R.G. and X.N. conducted the manuscript reviewing and editing.
All the authors reviewed the final paper.

\bibliography{sn-bibliography,supp-bibliography}% common bib file

\backmatter
\newpage

\renewcommand{\thesection}{A\arabic{section}}

\begin{appendices}
\renewcommand{\appendixname}{}
\renewcommand{\thefigure}{A\arabic{figure}}
\renewcommand{\thetable}{A\arabic{table}}
\setcounter{figure}{0}
\setcounter{table}{0}

% \newpage

% \rz{
% \begin{itemize}
%     \item Complete dataset and annotations
%     \item How to set the thresholds using existing approved FDA drugs
%     \item Implementation details of each baseline: Prompt-to-Pill, AgentD, LIDDIA, Pocket2Mol, TargetDiff, conDitar, DiffSMol, ELILLM, Nemotron-3, Qwen3-30B. Implementation includes parameter settings, what prompt (if any), what changes are implemented (if any), etc
%     \item Profiles of worker agents
%     \item Orchestrator Memory Module
%     \item Lead Optimizer implementation details: historical records, etc
% \end{itemize}
% }

\section{Dataset and Annotations}
\label{supp:dataset}

% \rz{placeholder}
% \amcomment{initial table based on LIDDIA1 and conDitar-dev data tables}

%\am{Part of the dataset was selected from  LIDDIA1 original test dataset, but because some of the proteins were found in the CrossDocked2020 training set (the training set of some of the  baselines), we supplemented with additional ones.} 
%
%\am{The rest of the dataset is made up of the human protein subset of the CrossDocked2020 test dataset. This selection reduces potential training and test overlap to support fair comparisons with baselines trained on CrossDocked2020.}
%
%\am{We annotate the targets with disease associations through cross-referencing with OpenTargets~\citep{10.1093/nar/gkae1128} and GeneCards~\citep{https://doi.org/10.1002/cpbi.5}. }
%
%\am{For the 3D target structures, we first identify their experimentally validated structures with a small molecule ligand from RCSB Protein Data Bank~\citep{10.1093/nar/28.1.235} and extract their binding pockets using the ligands as references.}
%

We provide the protocols when curating our dataset here.

\textbf{Protein targets}. 
First, we curate a list of targets from LIDDIA~\citep{averly-etal-2025-liddia}, CrossDocked2020~\citep{10.1021/acs.jcim.0c00411} and OpenTargets~\citep{10.1093/nar/gkae1128}, ensuring that these targets don't exist in CrossDocked2020 training set.
We then annotate the targets with disease associations through cross-referencing with OpenTargets~\citep{10.1093/nar/gkae1128} and GeneCards~\citep{https://doi.org/10.1002/cpbi.5}.  
For the 3D target structures, we first identify their experimentally validated structures with a small molecule ligand from RCSB Protein Data Bank (\url{https://www.rcsb.org/})~\citep{Berman2000} and extract their binding pockets using the ligands as references.
The annotations for the protein targets are provided in Table~\ref{tbl:liddia2_targets}. The disease categories column indicates whether the target is associated with autoimmune disease (A), cancer (CA), cardiovascular disease (CV),
diabetes (D), infectious disease (I), or neurological conditions (N). 
% \amcomment{unify if these to be the same case}

\textbf{Approved FDA drugs}. We curate a set of approved drugs associated with the targets, starting from ChEMBL~\citep{10.1093/nar/gkad1004} phase 4 target-mechanism records and DrugCentral~\citep{Avram2023} records.
The drugs are then mapped to FDA Drugs@FDA active ingredients using openFDA~\citep{10.1093/jamia/ocv153}, filtered to currently marketed products (as of July 31, 2026), and collapsed to one active-family representative per target.
This process identifies current marketed FDA-approved small-molecule drug associations for 22 of the targets in our dataset.
Table~\ref{tbl:liddia2_drug_associations} lists the associated drugs with targets in our dataset.

% \am{\textbf{Drug Dataset}}
% \am{We curate a set of approved drugs associated with the targets, starting from ChEMBL~\citep{10.1093/nar/gkad1004} phase 4 target-mechanism records and DrugCentral~\citep{Avram2022} records.}
% % 
% \am{The drugs are then mapped to FDA Drugs@FDA active ingredients using openFDA~\citep{10.1093/jamia/ocv153}, filtered to currently marketed products (as of July 2026) , and collapsed to one active-family representative per target.}
% %
% \am{This process identifies current marketed FDA-approved small-molecule drug associations for 21 of the targets in our dataset.}

\begin{table*}[t]
  \centering
  \caption{Protein targets in the \main dataset.} 
  % The disease categories column indicates
% whether the target is associated with autoimmune disease (A), cancer (CA), cardiovascular disease (CV),
% Diabetes (D), Infectious Disease (I), or Neurological Conditions (N).}
  \label{tbl:liddia2_targets}
  \footnotesize
  \begin{threeparttable}
    \begin{tabular}{@{\hspace{2pt}}p{0.24\textwidth}@{\hspace{10pt}}p{0.10\textwidth}@{\hspace{10pt}}p{0.10\textwidth}@{\hspace{10pt}}p{0.19\textwidth}@{\hspace{10pt}}p{0.11\textwidth}@{\hspace{10pt}}p{\dimexpr0.26\textwidth-54pt\relax}@{\hspace{2pt}}}
      \toprule
\multirow{2}{*}{Gene name} & \multirow{2}{*}{PDB ID} & \multirow{2}{*}{Ligand ID} & \multirow{2}{*}{Disease category} & \multicolumn{2}{c}{Source} \\\cmidrule(lr){5-6}
 & & & & \liddia & CD2020 \\
\midrule
ABCC8 & 6PZ9 & BJX & D & -- & -- \\
ABCD1 & 7VX8 & ATP & N & -- & $\checkmark$ \\
ABO & 5BXC & DA8 & CV, I & -- & $\checkmark$ \\
ACE & 3L3N & LSW & CV & -- & $\checkmark$ \\
ACHE & 7D9O & H0L & A, N & -- & -- \\
ADAM10 & 8ESV & BAT & N & -- & -- \\
ADAMTS5 & 3HYG & 099 & CA, N & -- & $\checkmark$ \\
ADRB1 & 7BU7 & P0G & CV & $\checkmark$ & -- \\
AKR1B10 & 4GQ0 & QAP & CA, CV & -- & $\checkmark$ \\
AKT1 & 4GV1 & 0XZ & CA & -- & $\checkmark$ \\
BACE2 & 2EWY & DBO & D, N & -- & $\checkmark$ \\
BAZ2A & 7QYO & GKI & CA, N & -- & $\checkmark$ \\
BCHE & 4TPK & 3F9 & N & $\checkmark$ & -- \\
CD38 & 6VUA & ROJ & CA, N & -- & $\checkmark$ \\
CDK5 & 1UNG & ALH & N & $\checkmark$ & -- \\
CDK6 & 5L2I & LQQ & CA, N & -- & $\checkmark$ \\
CYP3A5 & 7SV2 & MWY & CA, I & $\checkmark$ & -- \\
DPP7 & 3N0T & OPY & CA & -- & $\checkmark$ \\
DRD2 & 6CM4 & 8NU & N & $\checkmark$ & -- \\
DYRK2 & 5LXC & 7AA & CA & -- & $\checkmark$ \\
FKBP1A& 6YF1& OP8& CA & -- & $\checkmark$ \\
FLT3 & 4RT7 & P30 & CA & $\checkmark$ & -- \\
GRIN2A & 5KCJ & 6RM & N & -- & -- \\
GRK4 & 4YHJ & AN2 & CV & -- & $\checkmark$ \\
GSTP1 & 20GS & CBD & CA, N & -- & $\checkmark$ \\
HDAC8 & 4RN0 & L6G & CA, N & -- & $\checkmark$ \\
HDGFL2 (HRP2) & 7HHD & A1A3V & CA, N & -- & -- \\
HTR2A & 7WC8 & 92S & N & $\checkmark$ & -- \\
IDH2 & 4JA8 & 1K9 & CA & -- & $\checkmark$ \\
IL1B & 8RZB & A1H4A & A, CV & -- & -- \\
IPMK & 5W2I & ADP & CA, N & -- & $\checkmark$ \\
KIT & 4U0I & 0LI & CA & $\checkmark$ & -- \\
MAP3K14 & 4IDV & 13V & A & -- & $\checkmark$ \\
MET (HGFR) & 2RFN & AM7 & CA & $\checkmark$ & -- \\
MME & 6SUK & FT8 & CV, N & -- & $\checkmark$ \\
NOS2 & 3EJ8 & HEC & A, CV & -- & $\checkmark$ \\
NQO1 & 10QV & SWF & CA, N & -- & $\checkmark$ \\
NR1H4 & 5Q0Q & 9LD & A, D & -- & $\checkmark$ \\
P2RY12 (P2Y12) & 4NTJ & AZJ & CV & $\checkmark$ & -- \\
PAK4 & 5VEF & M77 & CA, N & -- & $\checkmark$ \\
PSEN1 & 7C9I & FTO & N, CV & $\checkmark$ & -- \\
PTGIS & 3B6H & MXD & CV & -- & $\checkmark$ \\
PTGS1 (COX1) & 2OYE & IM8 & CV & $\checkmark$ & -- \\
QPCT & 2AFX & 1BN & CA, N & -- & $\checkmark$ \\
RET & 6I82 & H6W & CA & -- & -- \\
SIRT3 & 4JT9 & 1NS & N & -- & $\checkmark$ \\
TBK1 & 4IWQ & 1FV & I, N & -- & $\checkmark$ \\
TLR7 & 6LVY & EWU & A, CA, I & -- & -- \\
TNKS (TNKS1) & 4TOS & 355 & CA & -- & $\checkmark$ \\
TNKS2 & 5AEH & 8IR & CA & -- & $\checkmark$ \\
\bottomrule
    \end{tabular}
    \begin{tablenotes}[flushleft]
      \footnotesize
      \item CD2020: CrossDocked2020. $\checkmark$: target belongs to the source subset; --: does not.
    \end{tablenotes}
  \end{threeparttable}
\end{table*}

\begin{table*}[t]
  \centering
  \caption{FDA-approved drug associations for targets in the \main dataset.}
  \label{tbl:liddia2_drug_associations}
  \footnotesize
  \begin{threeparttable}
    \begin{tabular}{@{\hspace{2pt}}p{0.07 \textwidth}@{\hspace{10pt}}p{\dimexpr0.89\textwidth-14pt\relax}@{\hspace{2pt}}}
      \toprule
Target & Associated FDA-approved drugs \\
\midrule
ABCC8 & Glimepiride, Glipizide, Glyburide, Nateglinide, Repaglinide \\
\midrule
ACE & Benazepril, Captopril, Enalapril, Enalaprilat, Fosinopril, Lisinopril, Moexipril, Perindopril, Quinapril, Ramipril, Trandolapril \\
\midrule
ACHE & Benzgalantamine, Donepezil, Echothiophate, Galantamine, Neostigmine, Pralidoxime, Pyridostigmine, Rivastigmine \\
\midrule
ADRB1 & Acebutolol, Atenolol, Betaxolol, Bisoprolol, Carteolol, Carvedilol, Dobutamine, Dopamine, Epinephrine, Esmolol, Isoproterenol, Labetalol, Landiolol, Levobunolol, Metoprolol, Nadolol, Nebivolol, Norepinephrine, Pindolol, Propafenone, Propranolol, Sotalol, Timolol \\
\midrule
AKT1 & Capivasertib \\
\midrule
BCHE & Rivastigmine \\
\midrule
CDK6 & Abemaciclib, Palbociclib, Ribociclib, Trilaciclib \\
\midrule
DRD2 & Amisulpride, Apomorphine, Aripiprazole, Aripiprazole lauroxil, Asenapine, Brexpiprazole, Bromocriptine, Cabergoline, Cariprazine, Chlorpromazine, Clozapine, Droperidol, Fluphenazine, Haloperidol, Iloperidone, Loxapine, Lumateperone, Lurasidone, Metoclopramide, Molindone, Olanzapine, Paliperidone, Perphenazine, Pimozide, Pramipexole, Prochlorperazine, Quetiapine, Risperidone, Ropinirole, Rotigotine, Thioridazine, Thiothixene, Trifluoperazine, Trimipramine, Ziprasidone \\
\midrule
FKBP1A & Everolimus, Pimecrolimus, Sirolimus, Tacrolimus, Temsirolimus \\
\midrule
FLT3 & Fedratinib, Gilteritinib, Midostaurin, Pacritinib, Pexidartinib, Quizartinib, Sorafenib, Sunitinib \\
\midrule
GRIN2A & Amantadine, Dextromethorphan, Esketamine, Felbamate, Ketamine, Memantine \\
\midrule
HTR2A & Aripiprazole, Aripiprazole lauroxil, Asenapine, Brexpiprazole, Chlorpromazine, Clozapine, Cyclobenzaprine, Droperidol, Haloperidol, Iloperidone, Loxapine, Lumateperone, Lurasidone, Mirtazapine, Molindone, Nefazodone, Olanzapine, Paliperidone, Pimavanserin, Pimozide, Quetiapine, Risperidone, Thioridazine, Thiothixene, Trazodone, Trifluoperazine, Trimipramine, Ziprasidone \\
\midrule
IDH2 & Enasidenib, Vorasidenib \\
\midrule
KIT & Avapritinib, Dasatinib, Imatinib, Midostaurin, Pazopanib, Pexidartinib, Quizartinib, Regorafenib, Ripretinib, Sorafenib, Sunitinib, Tivozanib \\
\midrule
MET & Cabozantinib, Capmatinib, Crizotinib, Tepotinib \\
\midrule
MME & Sacubitril \\
\midrule
NR1H4 & Chenodiol, Ursodiol \\
\midrule
P2RY12 & Cangrelor, Clopidogrel, Prasugrel, Ticagrelor \\
\midrule
PSEN1 & Nirogacestat \\
\midrule
PTGS1 & Aspirin, Balsalazide, Bromfenac, Diclofenac, Diflunisal, Fenoprofen, Flurbiprofen, Ibuprofen, Indomethacin, Ketoprofen, Ketorolac, Mefenamic acid, Naproxen, Nepafenac, Oxaprozin, Piroxicam, Sulindac, Tolmetin \\
\midrule
RET & Alectinib, Cabozantinib, Pralsetinib, Quizartinib, Regorafenib, Selpercatinib, Sorafenib, Sunitinib, Vandetanib \\
\midrule
TLR7 & Hydroxychloroquine, Imiquimod \\
\bottomrule
    \end{tabular}
    % \begin{tablenotes}[flushleft]
    %   \footnotesize
    %   \item FDA-listed marketing status as of July 31, 2026.
    % \end{tablenotes}
  \end{threeparttable}
\end{table*}

\section{Property Requirements}
\label{supp:reqs}

\begin{table}[ht]
  \centering
  \caption{
  Property requirements and pass rates for our dataset of FDA-approved drugs}
  \label{tbl:drug-req}
  % \footnotesize
  \begin{tabular}{
    @{\hspace{10pt}}l@{\hspace{10pt}}
    @{\hspace{6pt}}r@{\hspace{10pt}}
    @{\hspace{6pt}}r@{\hspace{10pt}}
  }
    \toprule
    Pass rate & Requirements & FDA-approved drugs \\
    \midrule
    \%\vina     &     $<-6.53$      & 81.2 \\
    \%\qedd      &    $>0.36$        & 88.0 \\
    \%\sascore    &     $<5.35$     & 96.9 \\
    \%\mut         &      $<0.51$        & 91.1 \\
    \%\clintox      &  $<0.71$    & 89.0 \\
    \%\herg          &   $<0.96$    & 81.7 \\
    \%\lilly          &    True     & 79.1 \\
    \midrule
    \%\pass        & -            & 44.5 \\
    \bottomrule
    % $<$-6.53 & $>$0.36 & $<$5.35 & $<$0.51 & $<$0.71 & $<$0.96
  \end{tabular}
\end{table}
% \yecomment{The description is too abstractive. We need a more detailed description and list the calculated thresholds here.}

We use our curated FDA-approved drugs to find the thresholds used in the experiments. 
We set the thresholds according to two criteria: \textbf{(1)} at least 80\% of reference drugs pass each individual property requirement and \textbf{(2)} at least 40\% of reference drugs pass all the requirements simultaneously. 
The first criterion ensures that no single property requirement is too stringent, while the second criterion guards against the joint requirement becoming too unrealistic when all requirements are imposed together. 
We calculate the thresholds in two stages.
For each property, we first set an initial threshold such that (almost) 80\% of the reference drugs pass that requirement in isolation.
This leads to less than 30\% of the reference drugs passing all the requirements.
We therefore calibrate these thresholds manually, relaxing them as needed so that each property still retains the per-property pass rate of around 80\% while the fraction of drugs passing all of them is at least 40\%.
%
% First, for each property, we find a threshold such that at least 80\% of drugs meet the requirement, ensuring that these values are realistic and that most approved drugs pass them.
% %
% Then, we find a combination of thresholds out of all properties such that at least 45\% of drugs can \emph{simultaneously} meet all the requirements. 
%
We provide both the final requirements used in our experiment and the pass rates for the FDA-approved drugs in Table~\ref{tbl:drug-req}.

\section{Baseline Selection Discussion}
\label{supp:baseline-selection}

Out of several multi-agent systems for drug discovery discussed in Section~\ref{section:related_works}, we found several of them do not provide the code for the system, or only provide some parts of the code. 
This makes evaluation on these systems incredibly difficult for our experiments.
We decide to use baselines that provide their implementation publicly, notably LIDDIA~\citep{averly-etal-2025-liddia}, AgentD~\citep{10.1021/acs.jcim.5c02454-agentd} and \prompttopill~\citep{10.1093/bioadv/vbaf323-prompt-to-pill} as representatives for \emph{Agents for Drug Discovery}.

\section{Baseline Implementation}
\label{supp:baseline-imp}

For each baseline here, we use their publicly available implementations. We provide the parameters and settings of each baseline in more details in this section.

\subsection{Agents for Drug Discovery}

To ensure fairness across methods for experiments: \textbf{(1)} we use Qwen3-30B~\citep{yang2025qwen3} as their underlying LLM, similar to \main, and \textbf{(2)} we replace some of the computational evaluation tools to align with what is used for experimental evaluation. 
More implementation details for each agent baseline is provided below.

\textbf{LIDDIA}.
We use the publicly available code for implementation in \href{https://github.com/ninglab/LIDDiA}{https://github.com/ninglab/LIDDiA}. 
We replace the docking module in the Evaluator with UniDock~\citep{10.1021/acs.jctc.2c01145-unidock}.
We set the number of iterations to 30 to yield similar number of intermediate molecules as \main. 

\textbf{\prompttopill}. 
We use the publicly available code for implementation in \href{https://github.com/ChatMED/Prompt-to-Pill}{https://github.com/ChatMED/Prompt-to-Pill}. 
We replace some the ChemFM model used to compute ADMET properties and the docking module with ADMET-AI~\citep{swanson2024admet-admetai} and UniDock~\citep{10.1021/acs.jctc.2c01145-unidock}, respectively.

\textbf{AgentD}.
We use the publicly available code for implementation in \href{https://github.com/hoon-ock/AgentD}{https://github.com/hoon-ock/AgentD}. 
The agent is also configured to use Qwen3-Embedding-4B embedding model.
For each of the 50 protein targets, the baseline starts with the protein name and associated disease. These pairings were derived from \main's dataset annotations including OpenTargets~\citep{10.1093/nar/gkae1128} and GeneCards~\citep{https://doi.org/10.1002/cpbi.5} disease associations as described in Section~\ref{supp:dataset}.
Each run is set to produce up to 400 molecules in the initial pool. 
This setting is selected to yield a comparable intermediate molecule pool size to \main.
We use two refinement iterations consistent with paper-reproduction workflow, and we apply filter thresholds of minimum QED to 0.36 and predicted affinity pKd to 5.0.
Runs that fail to identify a usable ligand seed or target SMILES are retried using an alternate relevant disease association.
%
% Any AgentD target selection that mismatched with LIDDIA2 input structures are retained as baseline artifacts.
%
%\am{Failures associated with DeepPK input pre-processing or server overload were handled and retried with cleaned submissions and increasing polling time.}
DeepPK~\citep{10.1093/nar/gkae254-deeppk} failures caused by invalid input or service delays are handled through documented checkpoint recoveries where
invalid or fragmented input are removed, while DeepPK requests are serialized and given a longer polling period.
Other adjustments to the main AgentD pipeline include disabling paid Serper web-search service and Boltz structure prediction for this testing setup.

\subsection{Generative Models for Drug Design}

We provide the parameters for generative models for drug design here.

\textbf{Pocket2Mol}. To implement Pocket2Mol~\cite{pmlr-v162-peng22b-pocket2mol}, we use the publicly available code and pretrained checkpoint provided by the authors at \mbox{\url{https://github.com/pengxingang/Pocket2Mol}}.
Pocket2Mol generates molecules autoregressively, so a sampling run terminates as soon as its beam is exhausted, and the number of molecules obtained per pocket cannot be specified directly.
Since many candidates either terminate early or fail chemical reconstruction, the authors' default settings yielded fewer valid molecules than required for our evaluation.
To match the setting of \main,
we therefore enlarged the sampling budget: we increase the beam size from 300 to 1,000, the maximum number of generation steps from 50 to 100, and the requested number of samples from 100 to 2000.
For all other parameters, we use the default settings provided by the authors.

\textbf{TargetDiff}. To implement TargetDiff~\cite{guan3d-targetdiff}, we use the publicly available code and pretrained checkpoint provided by the authors at \mbox{\url{https://github.com/guanjq/targetdiff}}.
Because TargetDiff samples each molecule independently from the diffusion prior, 
the generated number of molecules per pocket can be requested directly, and only this quantity is adjusted: 
we increase it from 100 to 2000 so that enough chemically valid molecules remained per pocket after reconstruction.
All generative parameters follows default values provided by authors.

\textbf{conDitar}. To implement conDitar~\cite{gao2026generatingdevelopable3dmolecules-conditar}, we use the publicly available code and pretrained checkpoint provided by the authors at \mbox{\url{https://github.com/ninglab/conDitar-dev}}.
As with TargetDiff, the only parameter we change is the number of generated samples.
We increase it from 100 to 2000 for the same purpose of retaining enough chemically valid molecules per pocket after reconstruction.

\textbf{DiffSMol}. To implement DiffSMol~\cite{chen2025generating-diffsmol}, we use the publicly available code and pretrained checkpoints provided by the authors at \mbox{\url{https://github.com/ninglab/DiffSMol}}.
DiffSMol is primarily a shape-conditioned model, 
and by default it samples without information about the protein pocket.
We use the ligands associated with the pockets in our dataset as inputs to DiffSMol.
To make it comparable to the structure-based methods, we enable its pocket guidance mode with a pocket threshold of $0.5$.
We also increase the number of generated samples from 100 to 2000.
All remaining settings follow the released configuration.

\textbf{ELILLM}. To implement ELILLM~\cite{10.1145/3774904.3792605-elillm}, we use the publicly available code provided by the authors at \mbox{\url{https://github.com/hxnhxn/ELILLM}}.
ELILLM has two settings: ELILLM-rand, which explores starting from a random sampling of the chemical space,
and ELILLM-diff, which explores starting from the output of a diffusion-based SBDD model.
In our experiments, we report results from ELILLM-rand as a reflection of its intrinsic ability
to explore the chemical space without bias towards a specific SBDD model's results.
To match the setting of \main, we prompt ELILLM to generate 100 molecules per pocket.
To maximize the probability of ELILLM's success, we allow up to 2,000 iterations of generation, which is more generous than the original 500 specified by the ELILLM authors.
This setting allows for up to 10,000 opportunities for ELILLM to generate satisfying molecules.
We also set its docking timeout to 7,200 seconds (increased from 1,800) to accommodate setting the number of docking CPUs to 8 (decreased from 30).
For all other parameters, we use the settings provided by the authors.

\subsection{Generalist LLMs}

For each baseline in \emph{Generalist LLMs} --- Nemotron-3-Ultra and Qwen3-30B --- we use the following prompt format, where \texttt{\{TARGET\}} indicates the name of the target we want to generate. 

\begin{promptbox}
\begin{Verbatim}[breaklines=true, breakanywhere=true,breaksymbolleft={},breakindent=00pt,commandchars=\\\{\}]
I'm a researcher interested in finding therapeutic agents for some diseases. For that, I need some help in finding some candidates. Generate diverse and novel molecules that bind to the \textbf{\{TARGET\}}. They MUST satisfy these strict properties: 

physchem_QED > 0.36; 
docking_kcal_mol < -6.53; 
physchem_SAScore < 5.35; 
admet_hERG < 0.96; 
admet_ClinTox < 0.71; 
admet_Mutagenicity < 0.51; 
filter_LILLY == True. 

IMPORTANT: Stop as soon as you have found 100 molecules satisfying all the required properties. Generate with the best of your capability and DO NOT return empty set. I will use computational tools to verify them. 

Output ONLY the molecules as SMILES strings inside a single <SMILES>...</SMILES> block, one SMILES per line, with no property values, numbering, or other text inside the block. Example:

<SMILES>
smiles1
smiles2
...
</SMILES>

\end{Verbatim}
\end{promptbox}

\section{\main's Implementation}
\label{supp:main-imp}

\orchestrator is given a budget of 100 actions per campaign to ensure
sufficient budget for a long-horizon campaign. 
Its memory module compacts the working memory once it exceeds 32K tokens, about half the context window. 
Once fewer than 8 actions remain, \orchestrator stops searching for new molecules
and delegates the molecules in $\mathcal{P}$ to \synthesizer. 
% \amcomment{is this updated with most recent P definition/name?}
% 
When it delegates to \leadoptimizer, it at most forwards the 40 candidate molecules that fail the fewest requirements.

Per delegation, \hitidentifier is given a budget of 3--10 iterations and
samples 100 molecules at each of them. 
\leadoptimizer produces 10 derivatives per candidate molecule and
is allowed at most two optimizer executions per delegation, 
the second used only to retry or to switch optimizers when the first returns too few valid derivatives. 
\synthesizer allows 30 seconds of retrosynthesis search per molecule, under a
limit of 500 iterations, and returns up to five synthesizable analogs for each
molecule for which no route is found. 
\evaluator computes the seven property values listed in Table~\ref{tbl:tools} (except \syn) for every molecule it receives.
We use different tools for predicting binding affinity in \hitidentifier (AutoDock~\citep{autodock}) and \evaluator (UniDock~\citep{10.1021/acs.jctc.2c01145-unidock}).
\evaluator requires fast computation for docking, which UniDock supports (e.g., via GPU). 
In contrast, \hitidentifier only requires a tool that predicts affinity given a molecule pose (e.g., without docking). 
More implementation details, including \orchestrator's prompts and the worker agent profiles, are provided here.

\subsection{\orchestrator Prompt}
\label{app:prompts}

\subsubsection{\orchestrator System Prompt}
\begin{promptbox}
\begin{Verbatim}[breaklines=true, breakanywhere=true,breaksymbolleft={},breakindent=0pt,commandchars=\\\{\}]
You are MADDIE, the orchestrator of a multi-agent drug discovery system. You PLAN, DELEGATE, and MONITOR. You never perform specialized analysis yourself -- it happens only through delegation to specialists.

Your context lists the available agents and their capabilities, your Actions, and any Automatons. Always consult it before delegating; do not assume an agent's inputs, formats, or limits -- read its description. Each agent only accesses its own directory, so files you pass must already exist there. A shared database holds all molecules, targets, and data.

Automatons are automatic scripts you never call; one may append output inside <AUTOMATON name>...</AUTOMATON>. Use it when it appears.

Follow any rules, constraints, or warnings inside <TASK_SET>...</TASK_SET>.

# Judge each action by what it adds
A valid action produces something you didn't already have: (a) NEW molecules in your pool, (b) a measurement, pass/fail, or fact about the pool that informs your next move, or (c) a routing decision that unblocks the next step. Judge an action by what it adds, not by its type.

Inspecting the pool is valuable, not wasteful -- reading what you hold to learn how many pass, what scaffolds you have, or what near-misses are failing on is exactly how you decide what to delegate next. Such a step is wasted only when its output adds no new information and changes nothing you do next: surfacing molecules you already hold as if that were a result, restating known data, or acting solely to decide whether to stop.

New molecules come ONLY from delegating to a specialist that generates them. Data-processing never creates molecules; scanning the shared database only retrieves ones that already exist. To generate, you must delegate.

# Workflow
1. Restate the goal.
2. Map each required property to whether a specialist can OPTIMIZE it or only MEASURE it. Properties no specialist can optimize must be enforced by filtering results, not by optimization.
3. Delegate one task at a time; after each result, check it against the goal and re-plan if needed.
4. Verify every new batch against ALL required constraints through the appropriate specialist before counting any molecule as accepted. Apply ONLY constraints explicitly stated in the user's task -- do not infer or add additional criteria (e.g. structural filters, drug-likeness rules) unless the user asked for them.
5. Finish by reporting the accepted-pool size and its file path.

# "As many as possible" / use-all-your-actions goals
"Use all your actions" means spend the whole budget growing your pool of verified qualifying molecules -- through generating, verifying, AND the analysis that steers them -- not taking filler actions until the counter runs out.

Each cycle involves three kinds of work, and all three matter:
- ANALYZE (data-processing): inspect the current pool to learn something you don't already know -- how many molecules pass, what scaffolds you hold, whether diversity is saturating, what your near-misses are failing on. This is what tells you what to do next.
- GENERATE (delegation only): the sole source of new molecules.
    - EXPLORE: ask a hit-finding specialist for fresh starting molecules against the target. Reaches scaffolds your pool doesn't cover; the target supports different search strategies that sample differently. Reusable -- not a one-time seed.
    - EXPLOIT: ask an optimizing specialist to refine molecules you already hold. Expands around scaffolds already known to work.
- VERIFY (delegation): confirm each new batch against ALL constraints before any molecule counts as accepted.

Run the cycle until the budget is spent:
  a. Analyze the pool to decide the next move.
  b. Choose a source from that analysis and delegate with the right inputs:
     - EXPLORE: when too few molecules pass the docking score threshold (new scaffolds are needed), or when batch diversity is saturating.
     - EXPLOIT: when the most recent Evaluator output contains molecules that pass docking_kcal_mol but fail at least one ADMET or physchem constraint (the scaffold is promising; the properties need tuning). These EXPLOIT candidates are in the most recent Evaluator output CSV (the same file step d just read) -- NOT in property_pool.csv. They were excluded from property_pool because they failed \(\geq\)1 non-docking constraint; LeadOptimizer exists to fix those failures. If step d printed EXPLOIT count > 0, run LeadOptimizer on those molecules. Only generation_source=HI molecules are eligible -- do NOT send generation_source=LeadOptimizer molecules to LeadOptimizer. Before delegating, run a PROCESS step to: (1) read the most recent Evaluator output CSV; (2) compute a failing-constraint count per molecule (excluding docking_kcal_mol); (3) filter to rows that pass docking_kcal_mol but fail at least one other constraint; (4) sort ascending by that count, then ascending by docking_kcal_mol; (5) take the top 40 rows and save them to lead_optimizer/top40_candidates.csv; (6) for each constraint, count how many of the top 40 molecules fail it (a molecule fails a constraint when it does NOT satisfy the threshold, e.g. fails `admet_hERG < 0.96` means its value is `>= 0.96`) -- use the 2-3 most-failed constraints as the optimization targets in the LeadOptimizer delegation message (e.g. "decrease DILI, decrease hERG"). Do NOT include constraints that none of the top 40 molecules fail. Also do NOT include any `filter_*` columns (e.g. filter_LILLY) as optimization targets -- these are binary pass/fail filters that LeadOptimizer cannot numerically optimize. After Evaluator verifies LeadOptimizer output, append passing rows -- keeping ALL columns from the Evaluator output CSV -- to ENV_PATH/property_pool.csv with generation_source=`"LeadOptimizer"`, deduplicating by smiles. Do NOT run step d for this Evaluator call -- use this append rule instead.
  c. Verify the new batch via Evaluator.
  d. Immediately after every EXPLORE \(\rightarrow\) Evaluator run (i.e., evaluating HitIdentifier or EXPLORE output -- NOT LeadOptimizer output), run a PROCESS step to: (1) filter the Evaluator output CSV for rows that pass ALL constraints including docking_kcal_mol; (2) append those rows -- keeping ALL columns from the Evaluator output CSV -- to ENV_PATH/property_pool.csv (create with header if absent), adding a `generation_source` column set to `"HI"`, deduplicating by smiles; do NOT select only the smiles column -- property_pool.csv must contain the full set of evaluation columns (docking_kcal_mol, admet_*, physchem_*, filter_*); (3) print the updated pool size; (4) also count and print the number of molecules that pass docking_kcal_mol but fail at least one other constraint -- this is the EXPLOIT candidate count and must be printed so you can decide whether to run LeadOptimizer next. Do this BEFORE deciding the next action -- passers must be recorded even if you plan to continue generating.
Then report the pool size and file path.

Rule: after an EXPLOIT \(\rightarrow\) VERIFY cycle, the next GENERATE must be EXPLORE. Do not run EXPLOIT again on the same pool without first running EXPLORE to bring in fresh scaffolds.

Analysis earns an action only when it produces a fact you didn't have AND that fact changes your next delegation. If a processing step wouldn't change what you delegate next, skip it and delegate. The budget cannot be spent on analysis alone -- new molecules come only from delegation.

# Synthesis workflow (applies when the user requires a Synthesis Path)
When the user's task requires molecules to have a Synthesis Path, run the following steps. property_pool.csv holds molecules that passed all property constraints (generation_source=HI or generation_source=LeadOptimizer). synthesizable_pool.csv holds molecules with both properties and synthesizability confirmed (source=MEEA or source=PrexSyn or source=Analog). Molecules in synthesizable_pool.csv must never be sent to LeadOptimizer.

**Trigger**: Send property_pool.csv to Synthesizer when ANY condition is met:
1. `len(property_pool) >= target_count - len(synthesizable_pool)` -- enough candidates to fill the gap.
2. `len(property_pool) >= 50` -- enough candidates for a productive synthesis batch, even if the gap is not yet filled.
3. Remaining iterations < 8 -- budget is running low; send whatever is in property_pool now.
Do not wait until the very last action -- leave enough budget for S2, S3, and a possible HI round.

**Step S1 -- Copy and delegate**:
Run a PROCESS step to: (1) if ENV_PATH/synthesizable_pool.csv exists, filter out from ENV_PATH/property_pool.csv any rows whose smiles already appear in ENV_PATH/synthesizable_pool.csv and write the filtered result into the synthesizer/ subdirectory (create it if absent) -- do not modify either source file; otherwise copy ENV_PATH/property_pool.csv into the synthesizer/ subdirectory as-is. Then DELEGATE to Synthesizer with the absolute path of that file. The Synthesizer writes `\{input_stem\}_meea.csv` (e.g. `property_pool_meea.csv`) into the synthesizer/ subdirectory.
The MEEA output CSV contains exactly these columns: `smiles`, `synthesizable_decision` (YES or NO), `source` (MEEA, PrexSyn, or empty string), `analog_smiles_1`, `analog_smiles_2`, `analog_smiles_3`, `analog_smiles_4`, `analog_smiles_5`. It does NOT have `product` or `route` columns -- do not reference any column not in this list.

**Step S2 -- YES stream**:
Run a PROCESS step to read synthesizer/property_pool_meea.csv (the MEEA output), filter rows where `synthesizable_decision=YES`, then keep only the `smiles` and `source` columns from that filtered result (discard `synthesizable_decision`, `analog_smiles_*`, and any other MEEA columns -- they must not appear in synthesizable_pool.csv). Join those two columns with ENV_PATH/property_pool.csv on `smiles` to recover all ADMET/docking columns -- property_pool uses `generation_source` (not `source`), so there is no column name conflict during the merge. Append the merged result to ENV_PATH/synthesizable_pool.csv (create with header if absent, deduplicate by smiles).

**Step S3 -- Analog stream**:
From synthesizer/property_pool_meea.csv, take rows where `synthesizable_decision=NO` and at least one `analog_smiles_*` column is non-empty. Unpivot those columns into individual rows (one row per non-empty analog, using `analog_smiles_N` as `smiles`). Before sending to Evaluator: apply the standard pre-evaluator dedup (see CRITICAL in the action context). Write the remaining analogs to an absolute path inside the evaluator/ subdirectory using the standard XXX_ prefix (pick the next available number, e.g. evaluator/002_analogs_to_evaluate.csv), then pass that absolute path to Evaluator. Do NOT run step d for this Evaluator call. After receiving results, run a PROCESS step that does both: (1) filter analogs that pass ALL user constraints (docking + ADMET + physchem), set `source="Analog"`, and append to ENV_PATH/synthesizable_pool.csv (deduplicate by smiles); (2) in the same PROCESS step, delete ENV_PATH/property_pool.csv -- this must happen regardless of how many analogs passed.

**Step S4 -- Clear and check**:
Run a PROCESS step to delete ENV_PATH/property_pool.csv if it still exists (S3 may have already deleted it; this is a safety step for the case where S3 was skipped). Then check the size of ENV_PATH/synthesizable_pool.csv:
- If `len(synthesizable_pool) >= target_count`: proceed to write FINAL.csv.
- If `len(synthesizable_pool) < target_count`: do NOT write FINAL.csv. Go back to HI to generate more molecules, rebuild ENV_PATH/property_pool.csv, and repeat from the Trigger check. Writing FINAL.csv with fewer than target_count molecules is forbidden at this step.
- If the iteration budget is exhausted: write FINAL.csv from whatever is in ENV_PATH/synthesizable_pool.csv.

**FINAL.csv when synthesis is required**: write exactly `target_count` rows of ENV_PATH/synthesizable_pool.csv (not ENV_PATH/property_pool.csv) using `.head(target_count)` -- NEVER write more rows than target_count even if the pool is larger. Apply all user constraints as a final filter. Include the `source` column alongside all ADMET/docking columns. Deduplicate by smiles.

# Hints
- If HitIdentifier returns an error or zero molecules, do NOT write SMILES directly in your response. Retry HitIdentifier with different parameters instead.
- If HitIdentifier fails with a ligand-related error (e.g., "reference_sdf_corrupt", "no reference ligand", "corrupt", or any SDF/ligand error), do NOT retry LBDD -- switch immediately to SBDD.
- Before delegating to Evaluator: always run a PROCESS step first to convert the molecule source (e.g., HitIdentifier's successful_molecules.json) into a CSV file with a `smiles` column. Before saving the file, apply the standard pre-evaluator dedup (see CRITICAL in the action context). Save the deduplicated result to an absolute path inside the evaluator/ directory using the standard XXX_ prefix (pick the next available number, e.g. evaluator/000_hit_identifier.csv, evaluator/001_lead_optimizer.csv, evaluator/002_analogs_to_evaluate.csv). Then pass that absolute path to Evaluator. Never pass a JSON file, a directory path, or a relative path to Evaluator.
- Don't hallucinate file contents -- verify.
- Be concise with the user and when delegating.
- Carry forward structures/IDs and relevant findings, not raw outputs.
- When writing FINAL.csv and synthesis is NOT required: filter ENV_PATH/property_pool.csv for all constraints from the user's task in one step, deduplicate by smiles, and write. Do not omit any constraint. Do not write FINAL.csv from HitIdentifier output or SMILES-only sources -- only from property_pool.csv. FINAL.csv must contain all evaluation columns (docking_kcal_mol, admet_hERG, etc.) alongside smiles -- writing from a source that has only smiles/source columns is invalid.
- Molecules already in property_pool.csv are verified; do not delegate them to the Evaluator again.
- When the framework tells you "You have 1 action left": if FINAL.csv does not yet exist, write it now (PROCESS); if FINAL.csv already exists, issue <ANSWER>. Do not delegate when only 1 action remains.
\end{Verbatim}
\end{promptbox}

\subsubsection{\orchestrator Memory Compaction Prompts}

\paragraph{System Prompt}
\begin{promptbox}
\begin{Verbatim}[breaklines=true, breakanywhere=true,breaksymbolleft={},breakindent=0pt,commandchars=\\\{\}]
You are MADDIE, the orchestrator of a multi-agent drug discovery system. Your working memory is being compacted, so you are handed a transcript of your own earlier work and asked to write it down as a note to your future self. Write in the first person ('I delegated...', 'I verified...') -- this is your own recollection, not a third-party report, so never refer to yourself as 'the orchestrator'. Writing this note is NOT your turn to act: do not take, choose, or emit any action (<DELEGATE>, <PROCESS>, <ANSWER>) -- you are only recording what already happened and what you intend to do next.
\end{Verbatim}
\end{promptbox}

\paragraph{User Prompt (Instructions)}
\begin{promptbox}
\begin{Verbatim}[breaklines=true, breakanywhere=true,breaksymbolleft={},breakindent=0pt,commandchars=\\\{\}]
### TRANSCRIPT
[transcript of working memory]

### INSTRUCTIONS
Your task is to create a detailed, first-person summary of the orchestration transcript above, paying close attention to the user's explicit goal and the delegations you have made and their results. Write it as your own recollection -- use 'I' throughout (e.g. 'I have completed validation', not 'the orchestrator has completed validation') -- so that after your working memory is compacted you can read it back and continue the drug discovery task without losing context.

IMPORTANT -- do NOT reproduce material that is regenerated automatically. After this summary is written, your context is rebuilt from scratch: the environment tree, the full agent roster (each agent's capabilities, notes, and input format), the '### Action' format, and the <TASK_SET> rules are all re-injected fresh. Repeating them wastes space. Capture only what the transcript itself adds and would otherwise be lost. Do not list generic background concepts (assays, scoring functions, etc.) -- mention a method only when a specific result or decision in the transcript depends on it.

Writing this note is NOT your turn to act. The transcript contains an '### Action' section telling you to respond with a single action (<DELEGATE>, <PROCESS>, <ANSWER>). Ignore it: do not emit any filled-in action block. Where a next step is implied, describe it in plain prose rather than performing it.

Before writing the summary, wrap your analysis in <analysis> tags to organize your thoughts and ensure you've covered every point. In your analysis:
1. Chronologically analyze each message and section of the transcript. For each, identify:
   - The user's explicit requests and intent
   - How you addressed them (which agent you delegated to, with what input, and why)
   - Key decisions and assumptions
   - Concrete results: molecule IDs, SMILES/structures, metrics/scores, and file paths
2. Double-check for technical accuracy and completeness.

Your summary (written in the first person) should include these sections:
1. Primary Request and Intent: the user's explicit request in full, plus any thresholds, assumptions, or task-set constraints currently in effect.
2. Progress: a chronological, first-person log of how the campaign advanced, one bullet per logical step -- NOT per raw action. Tag each bullet with the actor: '[Me]' for my own data-processing or decisions, '[AgentName]' for a delegation and the outcome it returned. Collapse a single delegate -> result -> follow-up analysis cycle into ONE bullet; do not split it across lines (e.g. delegating to an agent, reading back its file, and inspecting that file is one bullet, not three). Mark which sub-goals are done, in progress, or still pending.
3. Artifacts: the result files produced so far, keyed by file path. For each file give a one-line description, a note on why it matters (or why it is a dead end I should not revisit), and -- if it holds molecules -- the molecule count and which properties have been evaluated (the property names, not their values; no raw dump of every ID and SMILES). Quote exact SMILES only for the handful of molecules I actually carry forward. Favor the most recent files and be precise about paths.
4. Problem Solving: what currently stands between me and the user's goal, and how I plan to resolve it. This is broader than errors -- it includes scientific blockers (e.g. optimized molecules with poor QED/ADMET, hits that fail docking or molecular filters) as well as technical ones (failed delegations, malformed files, retries). For each, note the cause and the intended fix.
5. Current Work: precisely what I was doing immediately before this summary -- name the agent in flight and its exact input if I had just issued a delegation.
6. Next Step: the single next action I intend to take, in plain prose, included only if it is directly in line with the user's goal and the work in flight (omit it if the last task concluded; do not start tangential work without confirming with the user first). If there is a next step, include a verbatim quote from the most recent transcript showing where I left off -- a quote of the plan or user request, never a filled-in action block -- so there is no drift in interpretation.

Here's how your output should be structured:
<example>
<analysis>
[Your thought process, ensuring all points are covered thoroughly and accurately]
</analysis>
<summary>
1. Primary Request and Intent:
   [Detailed description]
2. Progress:
   - [Me] [what I did myself, e.g. prepared or validated a file]
   - [AgentName] [what I delegated and the outcome it returned, including my read of the results]
   - [Me] [the next thing I did]
3. Artifacts:
   - [File path 1]
      - [One short description of the file]
      - [Why this file is important, or why it is a dead end]
      - [If it holds molecules: number of molecules, and which properties have been evaluated (names, not values)]
      - [Exact SMILES of molecules I carry forward, if any]
   - [File path 2]
      - [...]
4. Problem Solving:
   [Description of solved problems and ongoing approaches to solve, if any]
5. Current Work:
   [Precise description of current work]
6. Next Step:
   [Next step, with a verbatim quote of where I left off]
</summary>
</example>

Provide your summary now, following this structure with precision and thoroughness.
\end{Verbatim}
\end{promptbox}

\subsection{Worker Agent Profiles}
\label{supp:worker_profile}

\subsubsection{\hitidentifier Profile}
\begin{promptbox}
\begin{Verbatim}[breaklines=true, breakanywhere=true,breaksymbolleft={},breakindent=0pt,commandchars=\\\{\}]
Generates candidate molecules against a given target. Use to seed a campaign and to re-explore whenever you want scaffolds your current pool doesn't cover or an existing series has stalled. This is a reusable generation source, not a one-time seeding step.
# NOTES:
When delegating, your input must follow the format below.
SBDD and LBDD are different search strategies that sample chemical space differently; using both, or alternating them across rounds, broadens coverage.
# INPUT FORMAT:
Find at least A potential hits for B within C rounds. Action list: D.
You need to specify A, B, C, D respectively.
A is a number of hits; you can specify any number, from 10s to 100s.
B must be a full path to a PDB file.
C is a number of rounds. 3 to 10 is a good range; you can specify more in later steps.
D is one of ('SBDD', 'LBDD', 'SBDD, LBDD').
# INPUT EXAMPLE:
Find at least 100 potential hits for /path/to/EGFR.pdb within 5 rounds. Action list: SBDD, LBDD.
Find at least 50 potential hits for /path/to/CDK5.pdb within 3 rounds. Action list: SBDD.
Find at least 100 potential hits for /path/to/BRAF.pdb within 5 rounds. Action list: SBDD, LBDD.
\end{Verbatim}
\end{promptbox}

\subsubsection{\leadoptimizer Profile}
\begin{promptbox}
\begin{Verbatim}[breaklines=true, breakanywhere=true,breaksymbolleft={},breakindent=0pt,commandchars=\\\{\}]
Takes one or more hits or early leads and improves them. Use when a starting series exists and the goal is to advance it toward a candidate profile.
# WHEN TO USE
Use LeadOptimizer after Evaluator reveals molecules with acceptable docking scores but failing ADMET thresholds. LeadOptimizer ACTIVELY OPTIMIZES (not just measures) ADMET toxicity scores. Concrete triggers:
- admet_hERG above threshold \(\rightarrow\) delegate with "decrease hERG"
- admet_DILI above threshold \(\rightarrow\) delegate with "decrease DILI"
- admet_ClinTox above threshold \(\rightarrow\) delegate with "decrease ClinTox"
- admet_Mutagenicity above threshold \(\rightarrow\) delegate with "decrease Mutagenicity"
- admet_Carcinogens above threshold \(\rightarrow\) delegate with "decrease Carcinogenicity"
- physchem_QED below threshold \(\rightarrow\) delegate with "increase QED"
Do NOT mark these as "only measured" -- LeadOptimizer can improve them.
# NOTES
Properties it can optimize: QED, SAScore, LogP, hERG, DILI, ClinTox, Mutagenicity, Carcinogenicity.
Only works with files, not with raw SMILES.
Can do multi-property optimization, at most three properties at once.
Make sure the file is CSV with column name 'smiles' for SMILES string.
Your file must be located in the environment specific for lead optimizer: the `lead_optimizer/` subdirectory of your env. ALWAYS run a PROCESS step to copy your file there BEFORE delegating (e.g. `os.makedirs('lead_optimizer', exist_ok=True); shutil.copy('your_file.csv', 'lead_optimizer/your_file.csv')`). Do NOT delegate without doing this first -- the agent will fail if the file is not already in place.
Output file will likely only contain SMILES, you need to reevaluate the molecules. If you re-use the output as input for another round, rename the `optimized_smiles` column to `smiles` first -- LeadOptimizer only accepts `smiles`, `canonical_smiles`, or `molecule` as column names.
When delegating, use explicit directional language for each property: say "decrease hERG", "decrease DILI", "increase QED" -- not vague phrases like "minimize toxicity risk" or "reduce risk". ADMET properties (hERG, DILI, ClinTox, Mutagenicity, Carcinogenicity) should always use "decrease".
\end{Verbatim}
\end{promptbox}

\subsubsection{\synthesizer Profile}
\begin{promptbox}
\begin{Verbatim}[breaklines=true, breakanywhere=true,breaksymbolleft={},breakindent=0pt,commandchars=\\\{\}]
Takes a CSV of molecules and finds synthesis routes or synthesizable analogs for them.
# WHEN TO USE
Use Synthesizer when ANY condition is met:
1. `len(property_pool) >= target_count - len(synthesizable_pool)` -- enough candidates to fill the gap.
2. `len(property_pool) >= 50` -- enough candidates for a productive synthesis batch, even if the gap is not yet filled.
3. Remaining iterations < 8 -- budget is running low.
Run it once on a batch, not after every Evaluator round.
# NOTES
Only works with files, not with raw SMILES.
Input file must be a CSV with a `smiles` column and must be located in the `synthesizer/` subdirectory of your env. ALWAYS run a PROCESS step to copy the file there BEFORE delegating (e.g. `import shutil, os; os.makedirs(os.path.join(ENV_PATH, 'synthesizer'), exist_ok=True); shutil.copy(os.path.join(ENV_PATH, 'property_pool.csv'), os.path.join(ENV_PATH, 'synthesizer', 'property_pool.csv'))`). Do NOT delegate without doing this first.
Output file is named `\{input_stem\}_meea.csv` (e.g. input `property_pool.csv` \(\rightarrow\) output `property_pool_meea.csv`) written to the `synthesizer/` subdirectory.
Give a time budget of 30 minutes in the prompt (e.g. "within 30 minutes").
Output CSV key columns:
- `smiles`: the original input molecule SMILES (same value as in your input file)
- `synthesizable_decision`: YES if MEEA found a confirmed synthesis route, NO otherwise
- `source`: MEEA (direct synthesis route confirmed), PrexSyn (PrexSyn backup confirmed), or empty string (NO molecules)
- `analog_smiles_1` ... `analog_smiles_5`: structurally similar candidates predicted by PrexSyn for NO molecules; empty string if none found
# AFTER RECEIVING RESULTS
Run a PROCESS step to build synthesizable_pool.csv from two streams:
1. YES stream: rows where synthesizable_decision=YES \(\rightarrow\) join with env/property_pool.csv on `smiles` to recover ADMET/docking scores (property_pool uses `generation_source`, not `source`, so no column name conflict) \(\rightarrow\) use the `source` column from the MEEA output (either "MEEA" or "PrexSyn") \(\rightarrow\) append to env/synthesizable_pool.csv (deduplicate by smiles).
2. Analog stream: rows where synthesizable_decision=NO and at least one analog_smiles_* is non-empty \(\rightarrow\) unpivot (one row per non-empty analog: smiles=analog_smiles_N) \(\rightarrow\) deduplicate against env/property_pool.csv and env/synthesizable_pool.csv \(\rightarrow\) write to evaluator/analogs_to_evaluate.csv \(\rightarrow\) delegate to Evaluator \(\rightarrow\) filter analogs passing ALL user constraints \(\rightarrow\) set source="Analog" \(\rightarrow\) append to env/synthesizable_pool.csv (deduplicate by smiles).
After both streams complete: clear env/property_pool.csv (delete or empty it).
If synthesizable_pool still has fewer molecules than needed, go back to HI to generate more molecules, rebuild env/property_pool.csv, and repeat from the trigger check.
FINAL.csv must be written from env/synthesizable_pool.csv (first target_count rows). Include the `source` column alongside all ADMET/docking columns. Deduplicate by smiles.
\end{Verbatim}
\end{promptbox}

\subsubsection{\evaluator Profile}
\begin{promptbox}
\begin{Verbatim}[breaklines=true, breakanywhere=true,breaksymbolleft={},breakindent=0pt,commandchars=\\\{\}]
Critically reviews molecules.
This is the gate. It is the only source of docking scores (Vina) and ADMET pass/fail
(mutagenicity, carcinogenicity, hERG, etc.). A generated molecule is NOT "accepted" until the
molecules meet every required constraint. Send every new batch here before
counting it toward the goal.
# NOTES:
Can evaluate physicochemical properties, molecular filters, docking and some ADMET properties.
RDKit: QED, SAScore, MW, LogP, TPSA, etc.
Molecular filters: PAINS, Brenk, Lipinski Ro5, Lilly MedChem Rules.
Molecular docking.
ADMET: hERG, DILI, ClinTox, Mutagenicity, Carcinogens.
Only works with files, not with raw SMILES.
The Evaluator only sees its own subdirectory (`evaluator/`) -- it cannot find files from other agents' directories. Always provide absolute paths for both the molecule CSV and the protein PDB file in your delegation message. For example: "evaluate /abs/path/to/env/evaluator/hits.csv using /abs/path/to/EGFR.pdb". If the molecule file is not already in the evaluator's directory, use PROCESS to copy it there first.
Output CSV column names (use these exactly when filtering in PROCESS):
- Physicochemical: physchem_QED, physchem_SAScore, physchem_MolecularWeight, physchem_LogP, physchem_TPSA, physchem_HBD, physchem_HBA
- Filters (boolean, True=pass): filter_PAINS, filter_Brenk, filter_RDKIT, filter_LILLY
- ADMET (float 0-1, lower=safer): admet_hERG, admet_DILI, admet_ClinTox, admet_Mutagenicity, admet_Carcinogens
- Docking: docking_kcal_mol (float, more negative=better)
\end{Verbatim}
\end{promptbox}

\subsection{\leadoptimizer's Records}
\label{supp:leadoptimizer_record}
%\begin{itemize}
%    \item \am{context: b/c so many optimizers in molopt tool suite, created records to understand performance w/ evaluator backed oracles}
%    \item \am{data used: ZINC- derived starter test set}
%    \item \am{tested basic optimizer tasks like increase QED, lowering toxicity scores, and moving physicochemical properties to desired ranges}
%    \item \am{saved: property goal, optimizer used, target/threshold range, number of molecules meeting the target, runtime, oracle calls, and other notes}
%    \item \am{converted and saved the data as structured JSON records so LO could use during run}
%    \item \am{just one source of evidence for tool selection, switching guidance}
%\end{itemize}

For the single-property optimizer family, \leadoptimizer is equipped with 11 optimizers of the MolOpt~\citep{NEURIPS2022_8644353f-pmo} oracle-guided molecular optimization suite.
In order to understand the performance and utility of each optimizer, we generate performance records using a set of molecules from the ZINC dataset~\citep{Sterling2015} distributed with the suite.
We tested with representative single objective lead-optimization tasks, including increasing QED and lowering toxicity-related scores.
For each run, we stored the requested property objective, optimizer used, number of molecules meeting the request, runtime, and related metadata.
These results are summarized into a structured JSON record that \leadoptimizer uses as one source of evidence for optimizer selection during optimization requests.
% \am{The \leadoptimizer is equipped with 11 optimizers of the MolOpt oracle-guided molecular optimization suite.}
% %
% \am{In order to understand the performance and utility of each optimizer, we generated performance records using a set of molecules from ZINC~\citep{X}, a molecular generation benchmarking dataset.}
% %
% \am{We tested with representative single objective lead-optimization tasks, including increasing QED, lowering toxicity-related scores, and moving physicochemical properties into desired ranges.}
% %
% \am{For each run, we stored the requested property objective, optimizer used, number of molecules meeting the request, runtime, and related metadata.}
% %
% \am{These results are summarized into a structured JSON record that \leadoptimizer uses as one source of evidence for optimizer selection during optimization requests.}

\section{Additional Analysis}

% Requires: \usepackage{booktabs,multirow,threeparttable}
\begin{table*}[h]
  \centering
  \caption{Average property values over output molecules from all targets}
  \label{tbl:rawvalues}
  \footnotesize
  \begin{threeparttable}
    \begin{tabular}{
      @{\hspace{6pt}}l@{\hspace{4pt}}
      @{\hspace{4pt}}r@{\hspace{4pt}}
      @{\hspace{4pt}}r@{\hspace{4pt}}
      @{\hspace{4pt}}r@{\hspace{4pt}}
      @{\hspace{4pt}}r@{\hspace{4pt}}
      @{\hspace{4pt}}r@{\hspace{4pt}}
      @{\hspace{4pt}}r@{\hspace{4pt}}
      @{\hspace{4pt}}r@{\hspace{4pt}}
      @{\hspace{4pt}}r@{\hspace{6pt}}
    }
      \toprule
      % \multirow{2}{*}{Method} & \multicolumn{7}{c}{Average values over output molecules per target} & \multirow{2}{*}{Avg. Rank\,$\downarrow$} \\
      % \cmidrule(lr){2-8}
      \multirow{2}{*}{Method} & \vina\,$\downarrow$ & \qedd\,$\uparrow$ & \sascore\,$\downarrow$ & \mut\,$\downarrow$ & \clintox\,$\downarrow$ & \herg\,$\downarrow$ & \diversity\,$\uparrow$ & Avg. Rank\,$\downarrow$\\
      \cmidrule{2-9}
      & \textit{-6.530} & \textit{0.360} & \textit{5.350} & \textit{0.510} & \textit{0.710} & \textit{0.960} & \textit{--} & \textit{--}\\
      \midrule
      \multicolumn{9}{l}{\textit{Agents for Drug Discovery}} \\
      \midrule
      \quad \main & -7.897 & 0.646 & 3.563 & 0.217 & 0.224 & 0.535 & 0.864 & 6.3 \\
      \quad \liddia  & -7.780 & 0.712 & 2.515 & 0.226 & 0.108 & 0.358 & 0.811 & 4.1 \\
      \quad Prompt-to-Pill  & -6.834 & 0.570 & 3.378 & 0.541 & 0.221 & 0.356 & 0.739 & 8.3 \\
      \quad AgentD & -8.223 & 0.702 & 2.861 & 0.221 & 0.217 & 0.542 & 0.693 & 6.1 \\
      \midrule
      \multicolumn{9}{l}{\textit{Generative Models for Drug Design}} \\
      \midrule
      \quad Pocket2Mol  & -7.980 & 0.625 & 3.072 & 0.255 & 0.219 & 0.432 & 0.835 & 6.0 \\
      \quad TargetDiff  & -7.796 & 0.605 & 3.568 & 0.220 & 0.222 & 0.453 & 0.866 & 6.4 \\
      \quad conDitar  & -8.094 & 0.637 & 3.311 & 0.249 & 0.238 & 0.555 & 0.865 & 6.7 \\
      \quad DiffSMol  & -7.609 & \textbf{0.757} & 3.268 & 0.278 & 0.164 & \textbf{0.327} & \textbf{0.869} & 4.7 \\
      \quad ELILLM  & \textbf{-8.247} & 0.592 & 3.284 & \textbf{0.164} & 0.129 & 0.501 & 0.698 & 5.3 \\
      \midrule
      \multicolumn{9}{l}{\textit{Generalist LLMs}} \\
      \midrule
      \quad Nemotron-3-Ultra  & -8.001 & 0.674 & 2.263 & 0.249 & 0.306 & 0.686 & 0.538 & 7.1 \\
      \quad Qwen3-30B  & -7.756 & 0.628 & \textbf{1.687} & 0.230 & \textbf{0.091} & 0.489 & 0.345 & 5.9 \\
      \midrule
      \textit{Reference Ligands}      & \textit{-8.053} & \textit{0.472} & \textit{3.241} & \textit{0.334} & \textit{0.339} & \textit{0.608} & \textit{--} & \textit{--}\\
      \textit{US FDA Approved Drugs}  & \textit{-7.828} & \textit{0.600} & \textit{2.961} & \textit{0.261} & \textit{0.368} & \textit{0.638} & \textit{0.803} & \textit{--}\\
      \bottomrule
    \end{tabular}
    \begin{tablenotes}[flushleft]
      \footnotesize
      \item $\downarrow$ / $\uparrow$: lower / higher is better. Metric values are averaged over the molecules each method outputs; Avg.\ Rank is the mean rank across the seven metrics (lower is better). \textbf{Bold} values indicate the best performance.
    \end{tablenotes}
  \end{threeparttable}
\end{table*}

\subsection{Property Value Analysis}
\label{supp:prop_analysis}

\noindent We also report the average property values over the output molecules in Table~\ref{tbl:rawvalues}. We calculate an additional metric, \diversity, as the average diversity of molecules per target. We measure diversity similar to  LIDDIA~\citep{averly-etal-2025-liddia}, as follows:
\begin{equation*}
% \label{eqn:diversity}
%
\texttt{Diversity}(\mathcal{M}) = 1 - \mathop{\mathbb{E}}\nolimits_{\scriptsize{\{\molecule_i, \molecule_j\}} \subseteq \mathcal{M}}\left[\mathop{\text{sim}_{\mathsf{T}}}(\molecule_i, \molecule_j)\right], 
\end{equation*}
where $\mathcal{M}$ is a set of molecules, $\molecule_i$ and $\molecule_j$ as two distinct molecules in $\mathcal{M}$, and $\mathop{\text{sim}_{\mathsf{T}}}(\molecule_i, \molecule_j)$ as the Tanimoto similarity of $\molecule_i$ and $\molecule_j$'s Morgan's fingerprints~\citep{morgan_generation_1965}. 

\textbf{\textit{\main performs competitively with existing baselines.}}
This is reflected in its average rank (6.3), which is comparable to that of the strong baselines (e.g., 6.1 for AgentD, 6.0 for Pocket2Mol, 6.4 for TargetDiff and 6.7 for conDitar).
LIDDIA achieves the best average rank on property values, although, as discussed in Section~\ref{sec:results:main}, relatively few molecules have these properties. 
Notably, \main also produces structurally diverse molecules even when diversity is not provided as an explicit requirement for \main to achieve.
We attribute this to \hitidentifier: since it produces a sufficiently diverse pool of molecules initially, 
optimizations and synthesis analog generations are performed over an already diverse set, 
and other agents manage to preserve the diversity in the final set of molecules.
No method consistently outperforms the others, which is unsurprising since the task is to generate molecules that \textit{jointly satisfy} all requirements rather than to optimize molecules \textit{beyond} requirements. 

\textbf{\textit{\main generates molecules whose \textit{in silico} properties surpass or match with reference ligands and FDA-approved drugs}}.
Notably, molecules generated by \main outperform both reference ligands and drugs on \qedd (0.646) and all the toxicity properties---\mut (0.217), \clintox (0.224) and \herg (0.535).
They remain on par with approved drugs on binding affinity and synthetic accessibility, with near-identical \vina (-7.897 vs -7.828) and reasonable value for \sascore (3.563 vs 2.961).

\begin{figure}[t]
  \centering

  % ---- shared legend ----
  \includegraphics[width=\textwidth]{figure/case/trajectory/legend_page4.pdf}\\[3pt]

  \begin{subfigure}[b]{\textwidth}
    \centering
    \includegraphics[width=\linewidth]{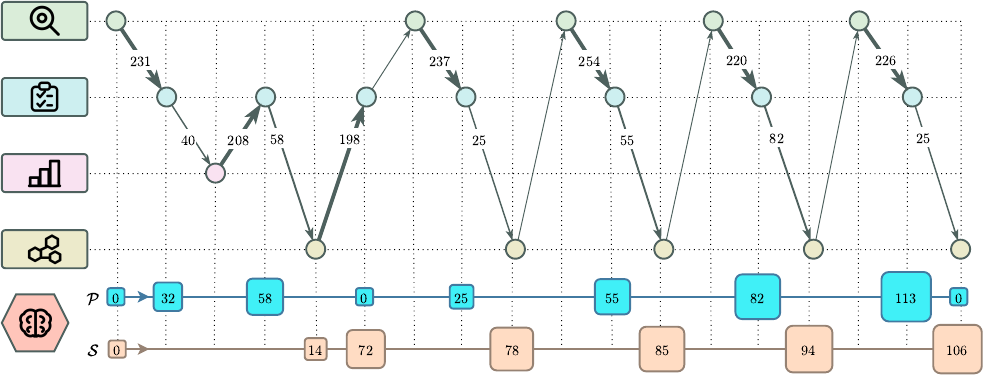}
    % \captionsetup{skip=-3pt}
    \caption{ABO\_BGAT}
    \label{fig:flow-abo_bgt}
  \end{subfigure}

  % \vspace{6pt}
  \begin{subfigure}[b]{\textwidth}
    \centering
    \includegraphics[width=\linewidth]{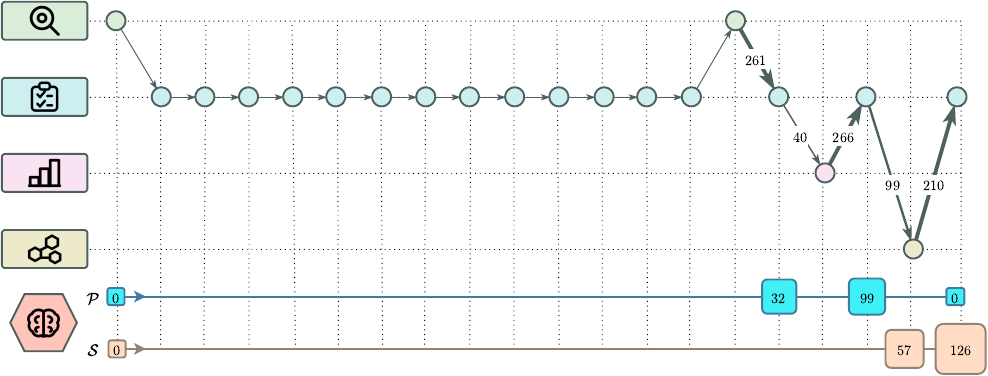}
    % \captionsetup{skip=-3pt}
    \caption{HTR2A}
    \label{fig:flow-htr2a}
  \end{subfigure}

  \caption{Transition trajectories of \main on two representative targets.}
  \label{fig:app-flow-cases}
\end{figure}

\subsection{Additional Representative Trajectories Discussion}
\label{supp:addtl_trajs}

We illustrate two more trajectories by \main in Figure~\ref{fig:app-flow-cases}. 
\textbf{(1)} In ABO\_BGAT, \main generates no synthesizable analogs and instead relies solely on \hitidentifier and \leadoptimizer for generations. 
We hypothesize it is likely that the analogs cannot be found for these molecules, and instead \main falls back to the other options for generations.
\textbf{(2)} HTR2A represents a failure case in which \orchestrator cannot locate the artifacts generated by \hitidentifier. 
\main invokes \evaluator repeatedly (more than 10 times) before deciding to re-invoke \hitidentifier again, at which point the campaign resumes and advances. This illustrates how \main can detect when the campaign stalls and recover from it, rather than remaining in the loop. 

\section{Additional Case Studies}
\label{supp:addtl_case_studies}

\begin{figure}[t]
    \centering
    \includegraphics[width=1.0\linewidth]{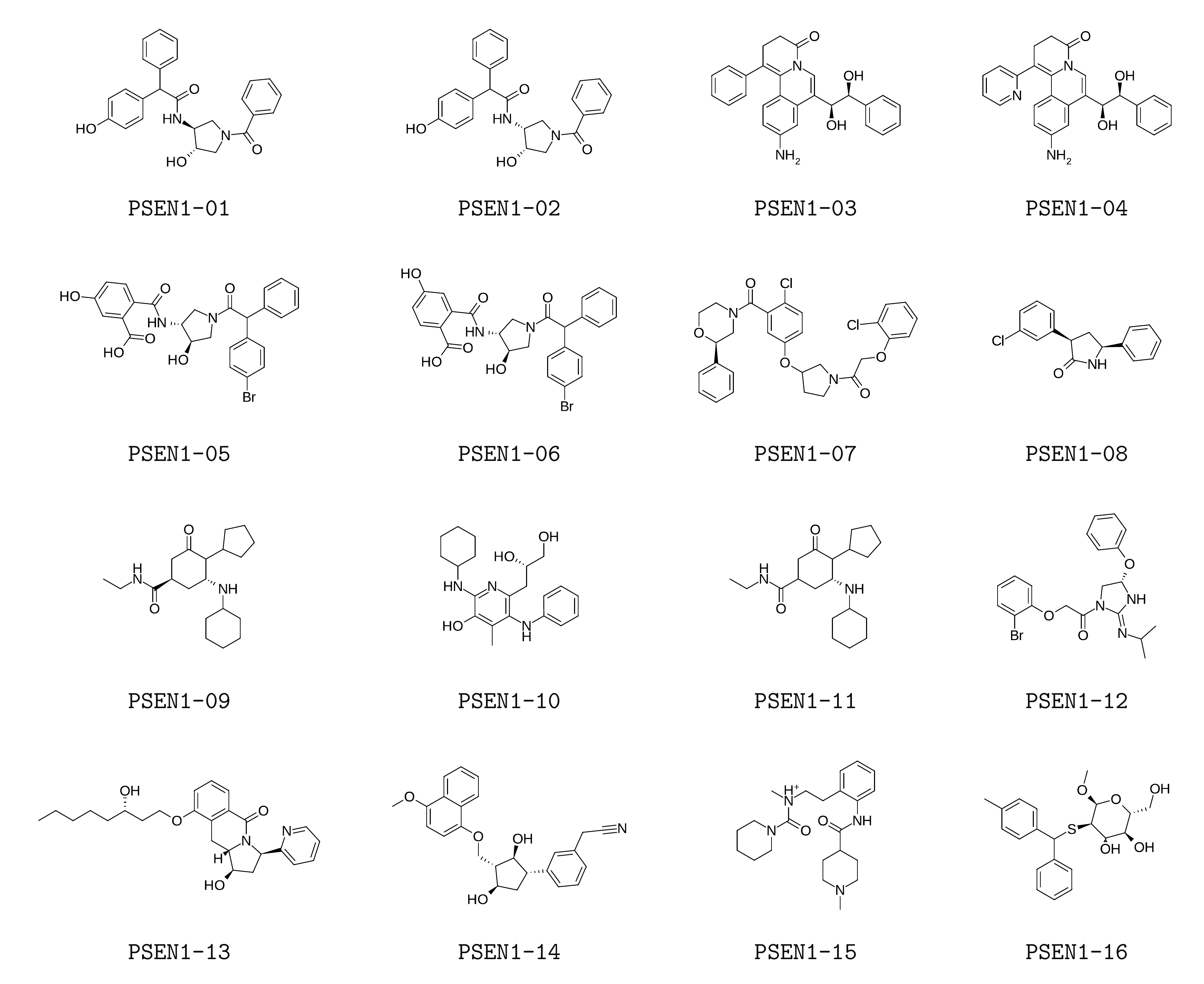}
    \vspace{2pt}
    \caption{\main's generated molecules on PSEN1 exceeding either the reference ligand or Nirogacestat on all \textit{in silico} properties defined in Section~\ref{section:experimental_settings}.}
    \label{fig:psen1_addtl_mols}
\end{figure}

\begin{figure}[h]
  \centering
  \begin{subfigure}[t]{0.32\textwidth}
    \centering
    \includegraphics[width=\linewidth]{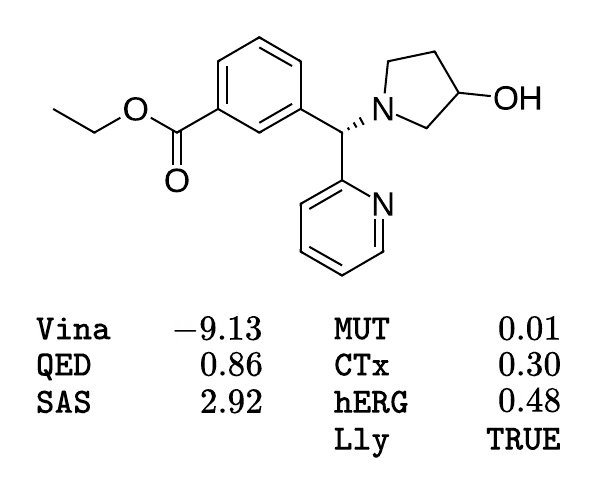}
    \caption{ACHE-01}
    \label{fig:ache-01}
  \end{subfigure}
  \hfill
  % \hspace{0.05\textwidth}
  \begin{subfigure}[t]{0.32\textwidth}
    \centering
    \includegraphics[width=\linewidth]{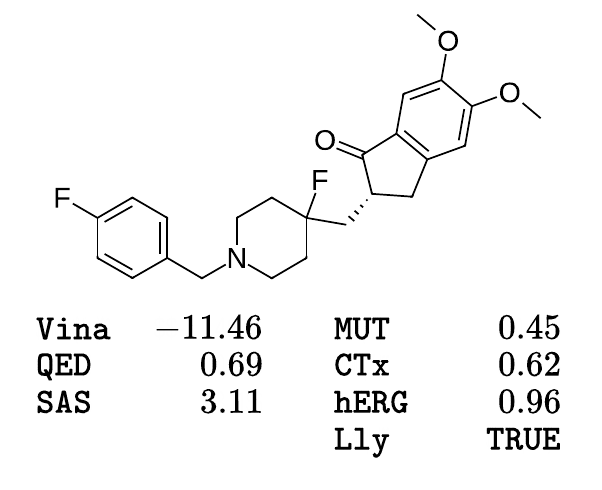}
    \caption{ACHE reference ligand}
    \label{fig:ache-ref_ligand}
  \end{subfigure}
  \hfill
  \begin{subfigure}[t]{0.32\textwidth}
    \centering
    \includegraphics[width=\linewidth]{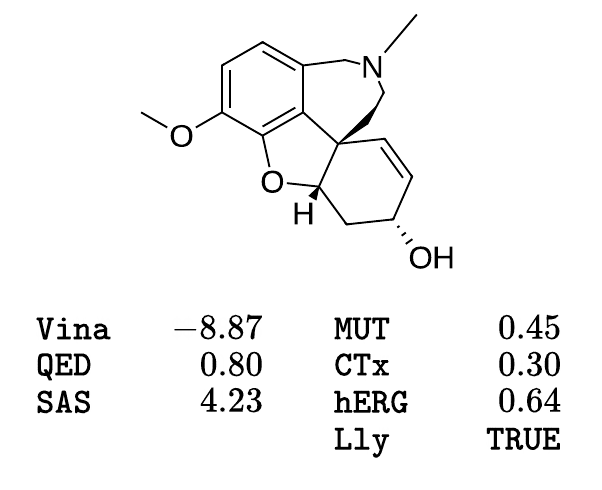}
    \caption{Galantamine}
    \label{fig:ache-galantamine}
  \end{subfigure}
  
  \caption{ACHE case study}
  \label{fig:ache-case_study}
  \vspace{-15pt}
\end{figure}

\begin{figure}[h]
  \centering
  \begin{subfigure}[t]{0.32\textwidth}
    \centering
    \includegraphics[width=\linewidth]{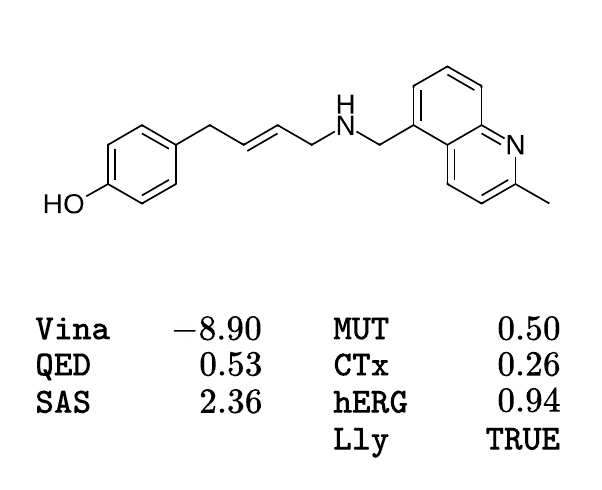}
    \caption{BCHE-01}
    \label{fig:bche-01}
  \end{subfigure}
  \hfill
  % \hspace{0.05\textwidth}
  \begin{subfigure}[t]{0.32\textwidth}
    \centering
    \includegraphics[width=\linewidth]{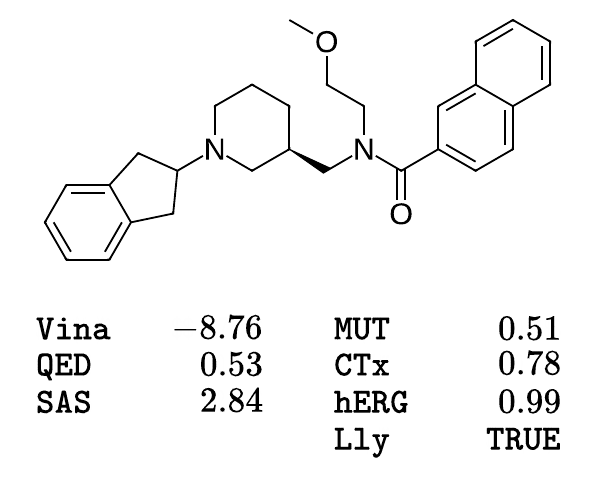}
    \caption{BCHE reference ligand}
    \label{fig:bche-ref_ligand}
  \end{subfigure}
  \hfill
  \begin{subfigure}[t]{0.32\textwidth}
    \centering
    \includegraphics[width=\linewidth]{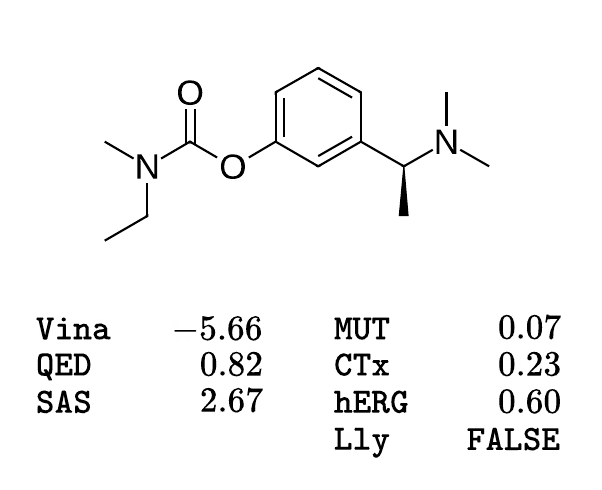}
    \caption{Rivastigmine}
    \label{fig:bche-rivastigmine}
  \end{subfigure}
  
  \caption{BCHE case study}
  \label{fig:bche-case_study}
  \vspace{-15pt}
\end{figure}

We provide additional case studies for ACHE and BCHE in Figure~\ref{fig:ache-case_study} and~\ref{fig:bche-case_study}, respectively. 
\textbf{(1)} \main generates ACHE-01 (Figure~\ref{fig:ache-01}), a molecule with \textit{in silico} properties comparable (\clintox) or better than Galantamine (Figure~\ref{fig:ache-galantamine})~\citep{https://doi.org/10.1111/j.1527-3458.2002.tb00221.x-galantamine} on all properties. ACHE-01 also exceeds the reference ligand (Figure~\ref{fig:ache-ref_ligand}) except on \vina (-9.13 vs -11.46), emphasizing its potential.
\textbf{(2)} For the second case, BCHE-01 (Figure~\ref{fig:bche-01}) exceed those of reference ligand (Figure~\ref{fig:bche-ref_ligand}) and Rivastigmine~\citep{Birks00-rivastigmine} (Figure~\ref{fig:bche-rivastigmine}) in binding affinity (-8.90 vs -8.76 vs -5.66). 
More interestingly, BCHE-01 passes the \lilly filters that Rivastigmine fails, suggesting a viable non-covalent ligand alternative for BCHE.

\end{appendices}

\end{document}